\documentclass[10pt,journal,compsoc]{IEEEtran}
\usepackage[switch]{lineno}

\ifCLASSOPTIONcompsoc
  \usepackage[nocompress]{cite}
\else
  \usepackage{cite}
\fi
\ifCLASSINFOpdf
   \usepackage[pdftex]{graphicx}
   \usepackage{wrapfig}
\else
\fi
\usepackage{amsmath}
\usepackage{algorithm}

\usepackage{algorithmic}

\usepackage{array}

 \usepackage[caption=false,font=footnotesize,labelfont=sf,textfont=sf]{subfig}
 \usepackage[caption=false,font=footnotesize]{subfig}
\usepackage{url}
\usepackage{hyperref}

\begin{document}

%
\title{Deep Video Prior for Video Consistency and Propagation}

%
%
%
%

\author{Chenyang~Lei,
        Yazhou~Xing,
        Hao~Ouyang,
        and~Qifeng~Chen,~\IEEEmembership{Member,~IEEE}

\IEEEcompsocitemizethanks{\IEEEcompsocthanksitem C. Lei, Y. Xing, H. Ouyang, and Q. Chen (corresponding author) are with the Department of Computer Science and Engineering, The Hong Kong University of Science and Technology, Clear Water Bay, Hong
Kong, China.\protect\\
Email: \{cleiaa, yxingag, houyangab\}@connect.ust.hk, cqf@ust.hk

}
}

%
%

\markboth{
}
{Shell \MakeLowercase{\textit{et al.}}: Bare Demo of IEEEtran.cls for Computer Society Journals}
\IEEEtitleabstractindextext{%
\begin{abstract}
Applying an image processing algorithm independently to each video frame often leads to temporal inconsistency in the resulting video. To address this issue, we present a novel and general approach for blind video temporal consistency. Our method is only trained on a pair of original and processed videos directly instead of a large dataset. Unlike most previous methods that enforce temporal consistency with optical flow, we show that temporal consistency can be achieved by training a convolutional neural network on a video with Deep Video Prior (DVP). Moreover, a carefully designed iteratively reweighted training strategy is proposed to address the challenging multimodal inconsistency problem. We demonstrate the effectiveness of our approach on 7 computer vision tasks on videos. Extensive quantitative and perceptual experiments show that our approach obtains superior performance than state-of-the-art methods on blind video temporal consistency. We further extend DVP to video propagation and demonstrate its effectiveness in propagating three different types of information (color, artistic style, and object segmentation). A progressive propagation strategy with pseudo labels is also proposed to enhance DVP's performance on video propagation. Our source codes are publicly available at \url{https://github.com/ChenyangLEI/deep-video-prior}.
\end{abstract}

}

\maketitle

\IEEEdisplaynontitleabstractindextext

%
\IEEEpeerreviewmaketitle

\IEEEraisesectionheading{\section{Introduction}\label{sec:introduction}}

%
%
%
%
\IEEEPARstart{N}{umerous} image processing algorithms have demonstrated great performance in single image processing tasks~\cite{li2016deep,yan2016automatic,isola2017image,zhang2016colorful,gharbi2017deep,Lei_2020_CVPR,Lei_2021_RFC}, but applying them directly to videos often results in undesirable temporal inconsistency (e.g., flickering). To encourage video temporal consistency, most researchers design specific methods for different video processing tasks~\cite{liu2018erase, nandoriya2017video, lang2012practical,xie2019video} such as video colorization~\cite{lei2019fully}, video denoising~\cite{liu2010high} and video super resolution~\cite{sajjadi2018frame}. 
Although task-specific video processing algorithms can improve temporal coherence, it is unclear or challenging to apply similar strategies to other tasks. Therefore, a generic framework that can turn an image processing algorithm into its video processing counterpart with strong temporal consistency is highly valuable. In this work, we study a novel approach to obtain a temporally consistent video from a processed video, which is a video after independently applying an image processing algorithm to each frame of an input video.

Prior work has studied general frameworks instead of task-specific solutions to improve temporal consistency~\cite{bonneel2015blind, yao2017occlusion, lai2018learning, eilertsen2019single}. Bonneel et al.~\cite{bonneel2015blind} present a general approach that is blind to image processing operators by minimizing the distance between the output and the processed video in the gradient domain and a warping error between two consecutive output frames. Based on this approach, Yao et al.~\cite{yao2017occlusion} further leverage more information from a key frame stack for occluded areas. However, these two methods assume that the output and processed videos are similar in the gradient domain, which may not hold in practice. To address this issue, Lai et al.~\cite{lai2018learning} maintain the perceptual similarity with processed videos by adopting a perceptual loss~\cite{johnson2016perceptual}. In addition to blind video temporal consistency methods, Eilertsen et al.~\cite{eilertsen2019single} propose a framework to finetune a convolutional network (CNN) by enforcing regularization on transform invariance if the pretrained CNN is available. Moreover, most approaches ~\cite{bonneel2015blind,yao2017occlusion,lai2018learning} enforce the regularization based on dense correspondences (e.g., optical flow or PatchMatch~\cite{barnes2009patchmatch}), and the long-term temporal consistency often degrades.

We propose a general and simple framework, utilizing Deep Video Prior by training a convolutional network on videos: the outputs of a CNN for corresponding patches in video frames should be consistent. This prior allows recovering most video information first before the flickering artifacts are eventually overfitted. 
Our framework does not enforce any handcrafted temporal regularization to improve temporal consistency, while previous methods are built upon enforcing feature similarity for correspondences among video frames~\cite{bonneel2015blind,yao2017occlusion,lai2018learning}.
Our idea is related to DIP (Deep Image Prior~\cite{ulyanov2018deep}), which observes that the structure of a generator network is sufficient to capture the low-level statistics of a natural image. DIP takes noise as input and trains the network to reconstruct an image. The network performs effectively to inverse problems such as image denoising, image inpainting, and super-resolution. For instance, the noise-free image will be reconstructed before the noise since it follows the prior represented by the network.
We conjecture that the flickering artifacts in a video are similar to the noise in the temporal domain, which can be corrected by deep video prior. 

Our method only requires training on the single test video, 
and no training dataset is needed. Training without large-scale data has been adopted commonly in internal learning~\cite{ulyanov2018deep,shocher2018ingan}. In addition to DIP~\cite{ulyanov2018deep}, various tasks~\cite{shocher2018zero,gandelsman2019double,shaham2019singan, zhang2019internal} show that great performance can be achieved by using only test data.

To address the challenging multimodal inconsistency problem, we further propose a carefully designed iteratively reweighted training (IRT) strategy.  Multimodal inconsistency may appear in a processed video. Our method selects one mode from multiple possible modes to ensure temporal consistency and preserve perceptual quality. We apply our method to diverse computer vision tasks. Results show that although our method and implementation are simple, our approach shows better temporal consistency and suffers less performance degradation compared with current state-of-the-art methods. 

As an extra contribution, we apply it to another video propagation task to demonstrate the effectiveness of DVP. Video propagation~\cite{jampani:cvpr:2017,Liu_2018_ECCV,DBLP:journals/tog/IizukaS19} is to propagating information (e.g., color) from reference frames in a video to all the other video frames. Video propagation can be quite useful in many areas~\cite{meyer2018deep,jampani:cvpr:2017}. Most video propagation methods propagate the information based on the similarity between video frames. Since our DVP can implicitly generate consistent features for corresponding patches in video frames, it can be applied directly to video propagation. Extensive quantitative and perceptual experiments show that applying DVP to video propagation obtains superior performance than state-of-the-art methods. 

Our contributions can be summarized as follows. In the preliminary version~\cite{DBLP:conf/nips/dvp} of this manuscript: first, we propose  Deep Video Prior (DVP) with its application to blind video temporal consistency; second, we propose an iteratively reweighted training (IRT) strategy to address multimodal inconsistency. Compared to the preliminary version, this manuscript is extended in the following three aspects:
\begin{itemize}
    \item We extend DVP to video propagation that can propagate user-provided information from several video frames to the whole video. 
    \item We propose progressive propagation with pseudo labels for video propagation.
    \item We conduct more experiments and analysis for deep video prior, including applying deep video prior to semi-supervised video object segmentation, analyzing failure cases, and presenting more quantitative results.
    \item We adopt two practical strategies to speed up the training process five times. Specifically, instead of training from scratch, we leverage a pretrained network that is early stopped. Besides, we further propose a coarse-to-fine training strategy to improve the training efficiency. We also propose a practical approach to decide when to stop the training by quantitatively measuring the training loss curve's smoothness.
\end{itemize}


\begin{figure*}[t]
\centering
\begin{tabular}{@{}c@{}}
\includegraphics[width=0.8\linewidth]{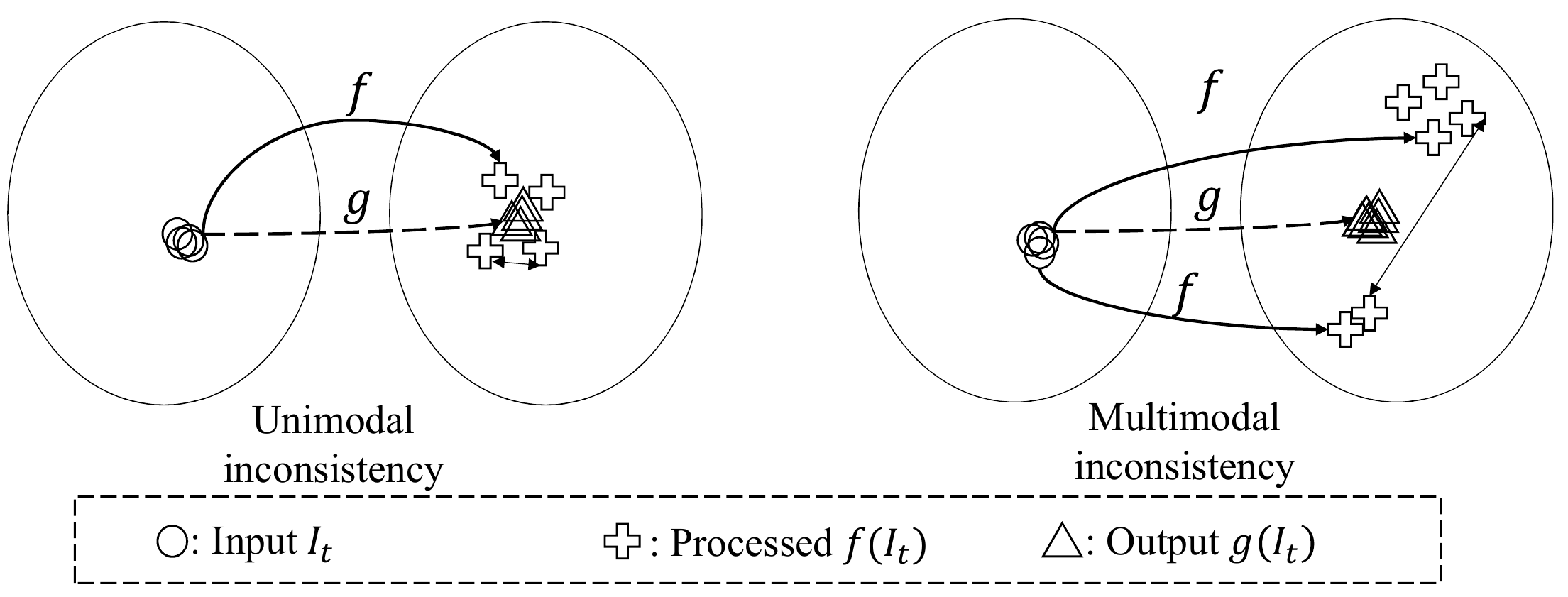}\\
\end{tabular}
\caption{Illustration of unimodal and multimodal inconsistency. In unimodal inconsistency, similar input video frames are mapped to moderately different processed frames within the same mode by $f$. In multimodal inconsistency, similar input video frames may be mapped to processed frames within two or more modes by $f$. A function $g$ is to improve the temporal consistency for these two cases. Note that $g$ is not our model.}

\label{fig:Illustration}
\end{figure*}

\section{Related Work}
\subsection{Blind video temporal consistency}
Blind video temporal consistency aims at designing a general approach to improving temporal consistency for many video processing tasks, such as colorization and dehazing. Lang et al.~\cite{lang2012practical} utilize edge-aware filters and propose a global optimization-based method for video temporal consistency problem. However, their method performs poorly at low-frequency video inconsistency, as shown in~\cite{bonneel2015blind}. Dong et al.~\cite{dong2015region} propose to segment each frame into regions and stabilize the video temporal inconsistency through adjusting the enhancement of the regions based on non-linear curves. Concurrent to~\cite{dong2015region}, Bonneel et al.\cite{bonneel2015blind} propose a general strategy through minimizing the distance between the output and the processed video in the gradient domain and a warping error between two consecutive output frames. To further improve this approach's performance, Yao et al.\cite{yao2017occlusion} further leverage more information from a keyframe stack for occluded areas. These two methods assume the similarity of the gradients between consecutive frames in processed videos, which may not hold in some tasks such as style transfer. To solve this problem, Lai et al.\cite{lai2018learning} propose a deep learning based approach to pursue the perceptual similarity between short-term and long-term frames in an unstable processed video. In addition to this, Eilertsen et al.~\cite{eilertsen2019single} propose to add transform invariance as temporal regularity to finetune the pretrained model to achieve temporal consistency. However, these methods require dense correspondence algorithms (e.g., optical flow or PatchMatch~\cite{barnes2009patchmatch}) as regularity and long-term temporal consistency often degrade. 

\subsection{Video propagation}
Propagating information across image pixels or pairs of images has been studied extensively by researchers. A classic algorithm~\cite{Levin2004} propagates color from color scribbles by the user to the whole gray image. In cooperation with optical flow estimation algorithms, it~\cite{Levin2004} could be used for video color propagation. Filtering techniques~\cite{he2012guided,rick2015propagated,kopf2007joint} could also be applied to propagate information in a faster way. However, these classical methods would hardly achieve good performance in general video propagation.

Deep learning largely expedites the research in video propagation. Jampani et al.~\cite{jampani:cvpr:2017} propose a deep learning based video propagation network and demonstrate the superiority on diverse tasks, including colorization and segmentation. They use a Bilateral Network to utilize similarities between frames based on spatial, temporal, and color information. Instead of manually defining a similarity measure, Liu et al.~\cite{Liu_2018_ECCV} utilize a neural network to learn the pixel affinity, which explicitly describes the task-specific temporal frame translations. Their method~\cite{Liu_2018_ECCV} outperforms Jampani et al.~\cite{jampani:cvpr:2017} in both performance and efficiency. In addition to the general video propagation network, there are also task-specific propagation methods such as video color propagation. Meyer et al.~\cite{meyer2018deep} propose a method for color propagation based on local matching and global matching. Vondrick et al.~\cite{vondrick2018tracking} propose a method to leverage the natural temporal coherency of color to colorize gray-scale videos by copying colors from a reference frame. With this temporal coherency, they can track an object in a video. Most mentioned work~\cite{Liu_2018_ECCV,vondrick2018tracking} can only propagate the color from a single frame to another one, which is not flexible. Iizuka et al.~\cite{DBLP:journals/tog/IizukaS19} propose a source-reference attention module to catch the similarity between arbitrary reference color frames and gray frames for video colorization. All these deep learning based video propagation methods require large-scale video datasets for training, which is quite time-consuming.

\subsection{Internal learning}
Ulyanov et al.~\cite{ulyanov2018deep} propose an internal learning approach based on Deep Image Prior (DIP) for image restoration. The input to the DIP network is random noise, and it is trained to reconstruct a single image. In the process of reconstructing an image, DIP tends to learn the natural features of an image first. Take image denoising as an example. DIP learns the noise-free image first and was shown to be quite powerful for solving inverse problems, including denoising, super-resolution, and inpainting. Gandelsman et al.~\cite{gandelsman2019double} observe that multiple DIPs tend to ``split" the image in the process of reconstructing an image, where the output image of each DIP is different from other images in low-level features. 

Shocher et al.~\cite{shocher2018zero} propose a method for super-resolution based on learning a model from a single image, which outperforms other methods trained on a large-scale dataset. They train a small CNN at test time. The data for training is extracted solely from the input image itself. Shocher et al.~\cite{shocher2018ingan} propose InGAN, which was the first internal GAN-based model combined with single image models. InGAN illustrates its ability to capture and remap the texture of an image. By doing so, they are able to draw random samples from an input image. SinGAN~\cite{shaham2019singan} is an unconditional generative model learned from a single image, which also shows excellent performance and general application. SinGAN can capture the internal distribution of a single image and generate similar image patches. More importantly, with the unconditional design, SinGAN applies to several image manipulations tasks. However, all these methods focus on the applications on a single image. Instead, we attempt to extend this idea to videos. Our proposed Deep Video Prior can reduce the training time largely while most previous deep learning based works for video processing require training on a large scale of videos~\cite{jampani:cvpr:2017,vondrick2018tracking,lei2019fully,Reda_2019_ICCV}.

\section{Deep Video Prior}
\label{sec:DVP}

\textbf{What is DVP?} The Deep Video Prior (DVP) is used to implicitly enforce video consistency in video processing tasks. This prior is based on an observation and a fact: the outputs of a CNN on two similar patches are expected to be similar (i.e., temporally consistent) at the early stage of training; the same object in different video frames has similar appearances. That is to say, the temporal consistency is implicitly achieved by DVP: the outputs of a CNN for corresponding patches in video frames should be consistent. 

DVP can be applied to the tasks that need to enforce the temporal consistency of generated videos. It can be directly applied to blind video temporal consistency, a task that improves the temporal consistency of processed video. Different from previous work, we do not need to enforce explicit regularization to make sure the features of correspondence are consistent, as analyzed in Sec.~\ref{sec:BTC}. In Sec.~\ref{sec:video_prop}, we show that DVP can also be applied to video propagation to propagate the information from reference frames to the other frames.

\noindent \textbf{Comparison to other regularization.}
In previous video processing task, a explicit correspondence-based regularization~\cite{lei2019fully,lai2018learning} is usually adopted to ensure the temporal consistency of generated videos. Let $I_t$ be the input video frame at time step $t$ and a output video $\{O_t\}_{t=1}^T$ can be obtained by applying the video processing algorithm $g$ to the input video $\{I_t\}_{t=1}^T$. The correspondences can obtained by optical flow~\cite{sun2018pwc} or PatchMatch~\cite{barnes2009patchmatch} in the output frames. Take optical-flow based regularization for an example, a regularization loss $L_{reg}$ is defined to minimize the distance between two frames in $\{O_t\}_{t=1}^T$:
\begin{align}
\label{eq:reg_loss}
    L_{reg} &= \sum_{t=2}^{T} {\|O_t - W(O_{t-1}, F_{t\xrightarrow{}t-1})\|},
\end{align}
    where $F_{t\xrightarrow{}t-1}$ is the optical flow from $I_t$ to $I_{t-1}$ and $W$ is the warping function. 

Note that our approach does not need to use the explicit correspondence-based regularization $L_{reg}$ in Eq.~\ref{eq:reg_loss}, and our generated output is still temporally consistent. This $L_{reg}$ is based on a video prior: correspondences in two frames should share similar features (e.g., color information). 

Since the video prior adopted correspondences-based regularization $L_{reg}$ can be implicitly achieved by our DVP, DVP does not have limitations of $L_{reg}$. $L_{reg}$ needs correspondences, and the correspondences obtained by optical-flow estimation algorithms~\cite{sun2018pwc} are not entirely correct, especially when the motion between two frames is large. Hence, the long-term consistency of videos with $L_{reg}$ is usually not satisfying.

\begin{figure*}
\centering
\begin{tabular}{@{}c@{}}
\includegraphics[width=0.9\linewidth]{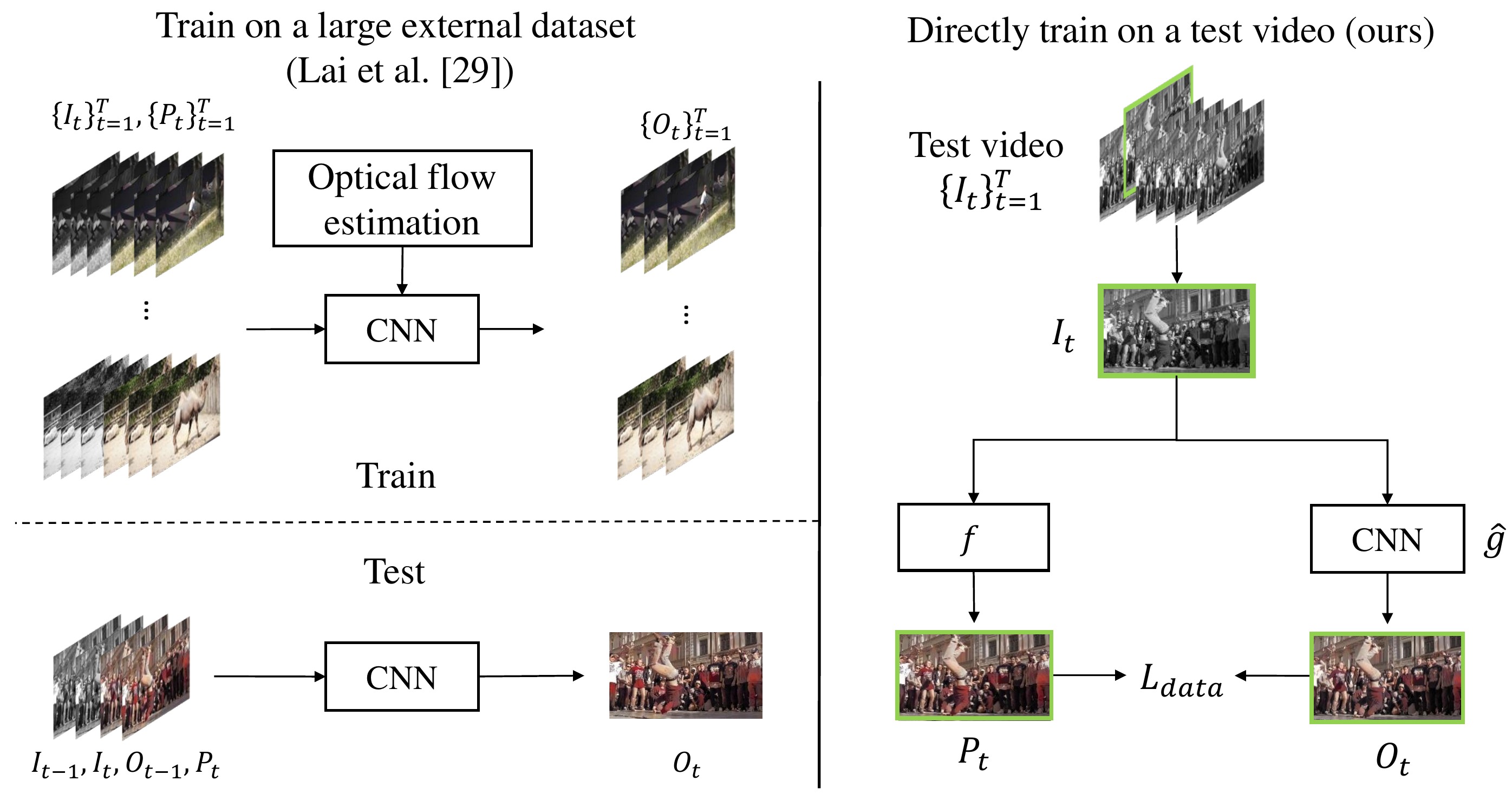}\\
\end{tabular}
\caption{The overview of our framework for blind video temporal consistency. The previous learning-based method~\cite{lai2018learning} requires a large-scale dataset which consists of input frames $\{I_t\}_{t=1}^T$ and processed frames $\{P_t\}_{t=1}^T$ pairs to train the network. Different from this, our pipeline directly trains on a test video. The network $\hat g$ is trained to mimic the image operator $f$. Note that network $\hat g$ is not a specific type of network so that our pipeline can adapt to different tasks. For each iteration, only one frame is used for training.}
\label{fig:Framework.}
\end{figure*}

\label{subsec:training}


\section{Blind Video Temporal Consistency}
\label{sec:BTC}
\subsection{Preliminaries}
Let $I_t$ be the input video frame at time step $t$, and the corresponding processed frame $P_t = f(I_t)$ can be obtained by applying the image processing algorithm $f$. For instance, $f$ can be image colorization, image dehazing, or any other algorithm. 

\noindent \textbf{Temporal inconsistency.} 
Temporal inconsistency appears when the same object has inconsistent visual content in $\{P_t\}_{t=1}^T$. For example, in Fig.~\ref{fig:LargeInconsistency_Cmp}, some corresponding local patches of two input frames vary a lot in processed frames. There are mainly two types of temporal inconsistency in a processed video: unimodal inconsistency and multimodal inconsistency. The left schematic diagram in Fig.~\ref{fig:Illustration} illustrates unimodal inconsistency, in which the $\{I_t\}_{t=1}^T$ are consistent and $\{P_t\}_{t=1}^T$ are not that consistent (e.g., flickering artifacts). For some tasks, multiple possible solutions exist for a single input (e.g., for colorization, a car might be colorized to red or blue). As a result, the temporal inconsistency in $\{P_t\}_{t=1}^T$ is visually more obvious, as shown in the right schematic diagram in Fig.~\ref{fig:Illustration}.

\begin{figure*}
\centering
\begin{tabular}{@{}c@{\hspace{1mm}}c@{\hspace{1mm}}c@{\hspace{1mm}}c@{\hspace{1mm}}c@{}}
\includegraphics[width=0.19\linewidth]{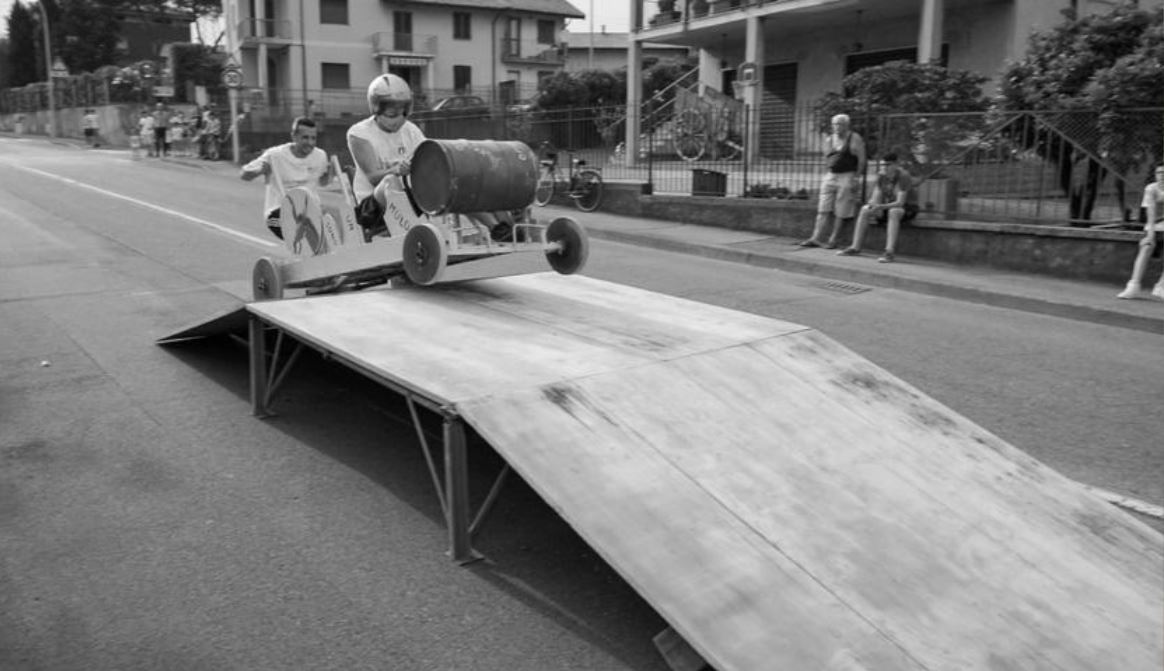}&
\includegraphics[width=0.19\linewidth]{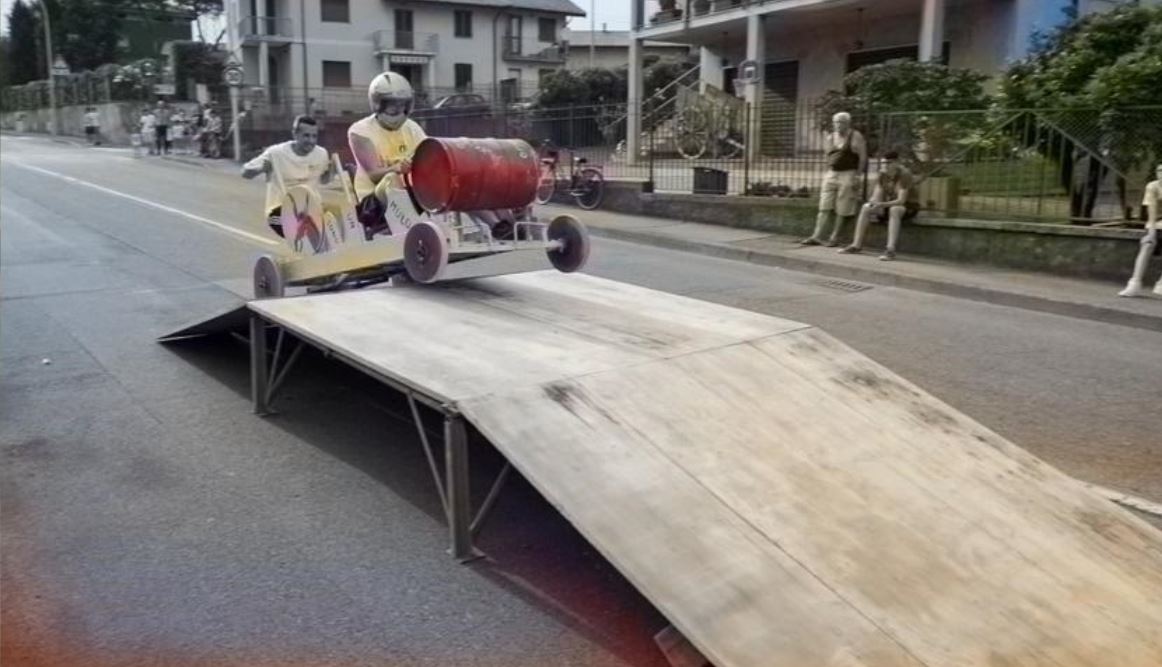}&
\includegraphics[width=0.19\linewidth]{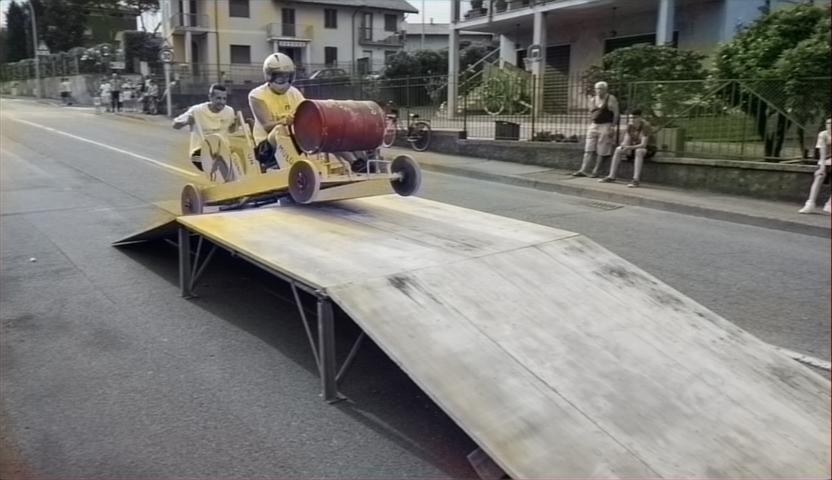}&
\includegraphics[width=0.19\linewidth]{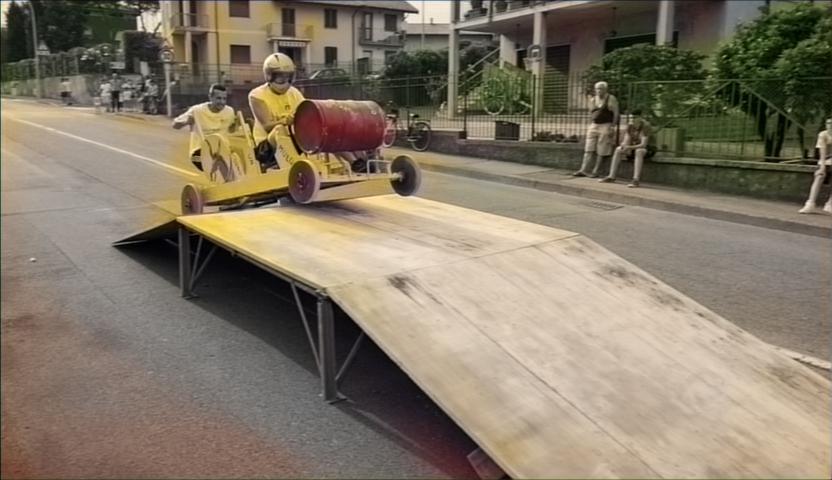}&
\includegraphics[width=0.19\linewidth]{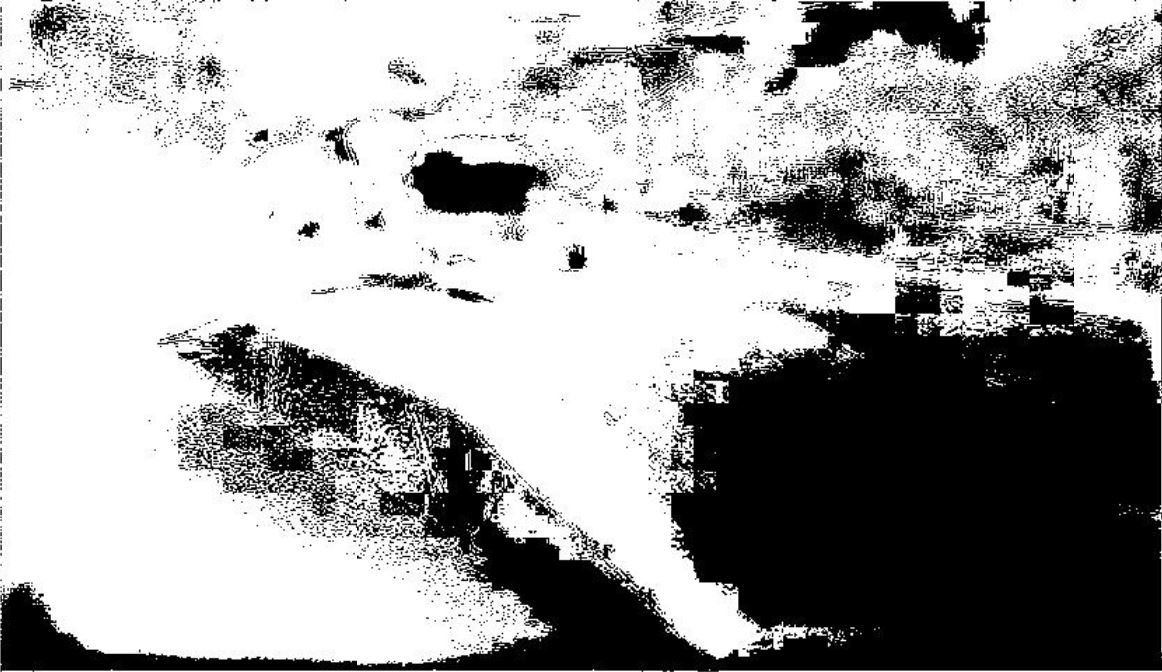}\\

Input $I_t$ & Processed $P_t$ & Output $O_{t}^{main}$ & Output $O_{t}^{minor}$ & Confidence $C_{t}$  \\
\end{tabular}
\caption{A confidence map can be calculated to exclude outliers.}
\label{fig:Confidence_map.}

\label{fig:Different_inconsistency}
\end{figure*}

\begin{figure*}
\centering
\begin{tabular}{@{}c@{}c@{}}
\rotatebox{90}{\small \hspace{12mm}(a) DVP}&
\includegraphics[width=0.8\linewidth]{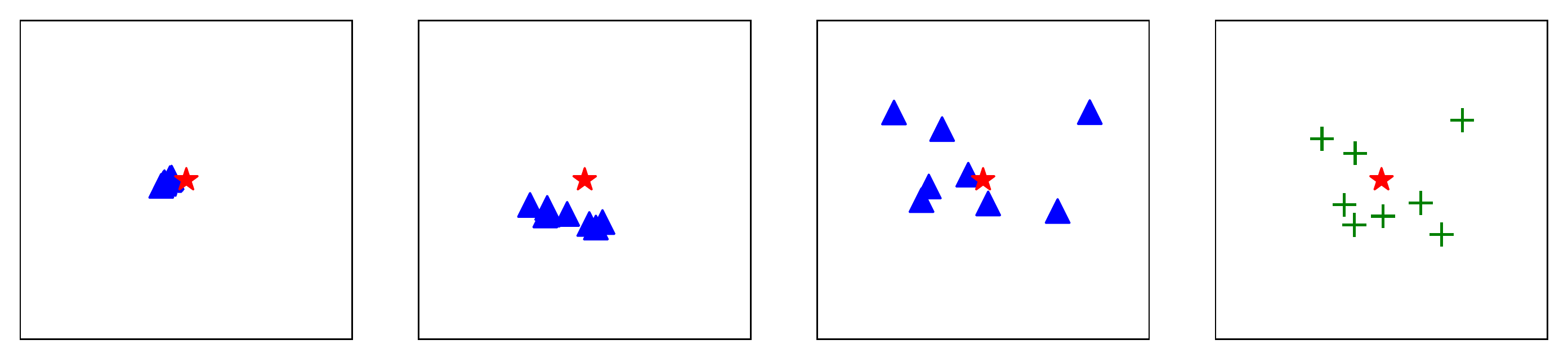}
\\
\rotatebox{90}{\small \hspace{12mm}(b) DVP} &
\includegraphics[width=0.8\linewidth]{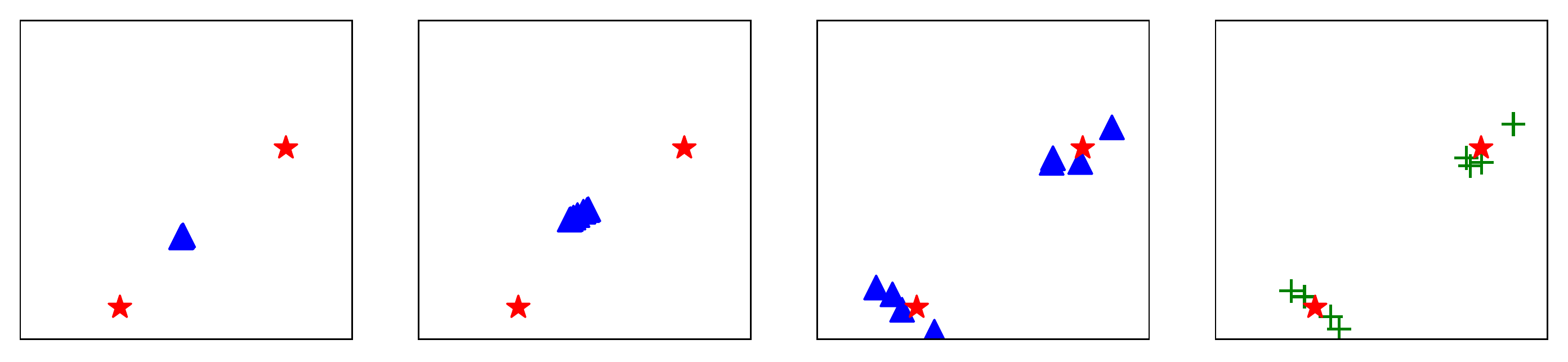}
\\
\rotatebox{90}{\small \hspace{12mm} \small{(c)DVP+IRT}}&
\includegraphics[width=0.8\linewidth]{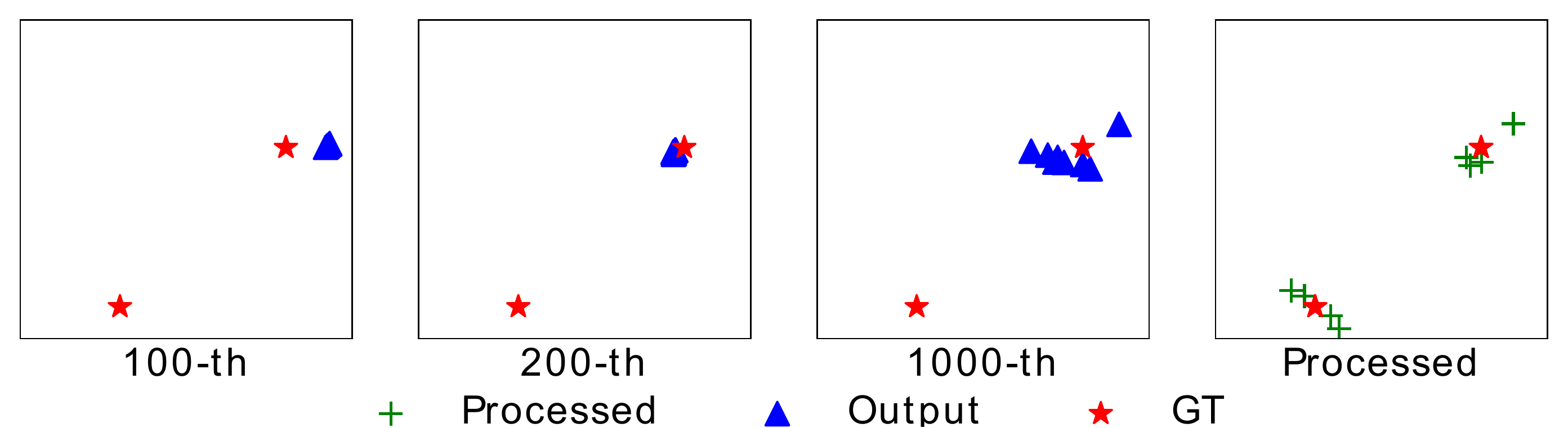}
\\

\end{tabular}
\caption{A toy example that demonstrates deep video prior. The green dots are the processed frames; the blue triangles are the output of the network; the red stars are the center of processed frames. (a): Training by DVP on $\{P_t\}_{t=1}^8$ with unimodal inconsistency. (b): Training by DVP on $\{P_t\}_{t=1}^8$ with multimodal inconsistency. (c): Training by DVP and IRT on $\{P_t\}_{t=1}^8$ with multimodal inconsistency. 
For all the cases, in the beginning, the outputs $\{O_t\}_{t=1}^8$ tend to be consistent with each other. After 200 iterations, outputs $\{O_t\}_{t=1}^8$ tend to separate from each other and completely overfit the processed frames $\{P_t\}_{t=1}^8$ after 1000 iterations. In (b), we notice that the output at the 200th iteration is not close to any ground truth. That is to say, our output has performance degradation performance if we only use the DVP for multimodal inconsistency. In (c), in cooperation with DVP and IRT strategy, we can handle the challenging multimodal inconsistency.}
\label{fig:DVP_toy}
\end{figure*}

\noindent \textbf{Blind video temporal consistency} Blind video temporal consistency~\cite{bonneel2015blind,lai2018learning} aims to design a function $g$ to generate a temporally consistent video $\{O_t\}_{t=1}^T$ that preserve the processed performance of $\{P_t\}_{t=1}^T$. Previous work~\cite{bonneel2015blind,lai2018learning} usually uses a regularization loss $L_{reg}$ to minimize the distance between correspondences in the output frames $\{O_t\}_{t=1}^T$, as introduced in Eq.~\ref{eq:reg_loss}. Also, a reconstruction loss $L_{data}$ is used to minimize distance between $\{O_t\}_{t=1}^T$ and $\{P_t\}_{t=1}^T$. Therefore, a loss function~\cite{lai2018learning} or objective function~\cite{bonneel2015blind} $L$ is commonly adopted for blind video temporal consistency:

\begin{align}
    L &= L_{data} + L_{reg}.
\end{align}

\subsection{Blind video temporal consistency via DVP}
Deep Video Prior (DVP) allows recovering most video information while eliminating flickering before eventually overfitting all information, including inconsistency artifacts. It is because valid video information is usually temporally consistent, which follows the DVP; while the temporal inconsistency artifacts do not follow the DVP. 

As shown in Fig.~\ref{fig:Framework.}, we propose to use a fully convolutional network $\hat g(\cdot;\theta)$ to mimic the original image operator $f$ while preserving temporal consistency. Different from Lai et al.~\cite{lai2018learning}, only a single video is used for training $\hat g$, and only a single frame is used in each iteration. Since we observe the $L_{reg}$ can be implicitly achieved by Deep Video Prior (DVP), our model does not need the $L_{reg}$ and thus avoids optical flow estimation. 
We initialize $\hat g$ randomly, and then the parameters of network $\theta$ can be updated in each iteration with a single data term without any explicit regularization:
\begin{align}
     \mathop{\arg\min}_{\theta} \ \ L_{data}(\hat g(I_t;\theta), P_t),
\end{align}
where $L_{data}$ measures the distance (e.g., $L_1$ distance) between $\hat g(I_t;\theta)$ and $P_t$. We stop training when $\{O_t\}_{t=1}^T$ is close to $\{P_t\}_{t=1}^T$ and before artifacts (e.g., flickering) are overfitted. Through this basic architecture, unimodal inconsistency can be alleviated. A neural network $\hat g$ and a data term $L_{data}$ are yet to design in our framework. 

In practice, we adopt u-net~\cite{ronneberger2015u} and perceptual loss~\cite{johnson2016perceptual,ChenK17} for all evaluated tasks. Note that $\hat g$ is not restricted to u-net, other appropriate CNN architectures (e.g., FCN~\cite{long2015fully} or original architecture of $f$) are also applicable. To further accelerate the training process, we introduce a better network initialization as well as a coarse-to-fine training strategy, as discussed in Section~\ref{sec:speedup}. 

\begin{figure*}[t]
\centering
\begin{tabular}{@{}c@{\hspace{1mm}}c@{\hspace{1mm}}c@{\hspace{1mm}}c@{\hspace{1mm}}c@{}}
\rotatebox{90}{\small \hspace{8mm} Input }&
\includegraphics[width=0.237\linewidth]{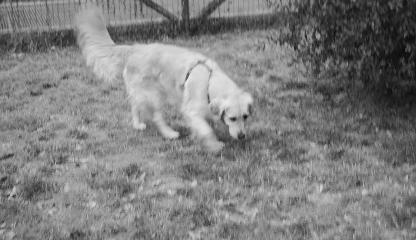}&
\includegraphics[width=0.237\linewidth]{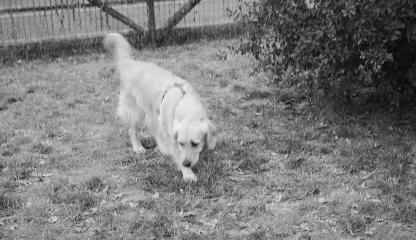}&
\includegraphics[width=0.237\linewidth]{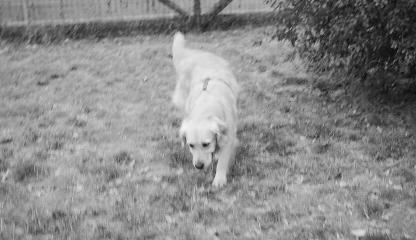}&
\includegraphics[width=0.237\linewidth]{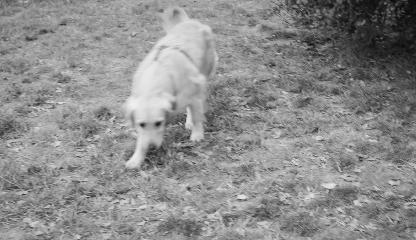}\\

\rotatebox{90}{\small \hspace{8mm} Ours }&
\includegraphics[width=0.237\linewidth]{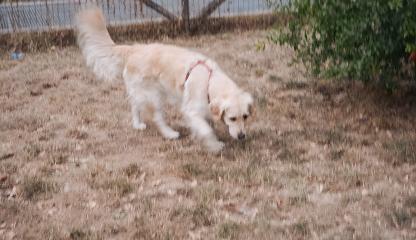}&
\includegraphics[width=0.237\linewidth]{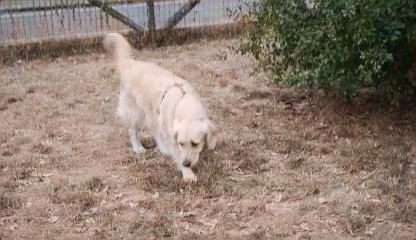}&
\includegraphics[width=0.237\linewidth]{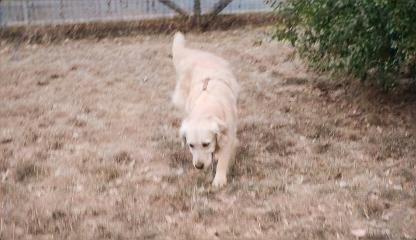}&
\includegraphics[width=0.237\linewidth]{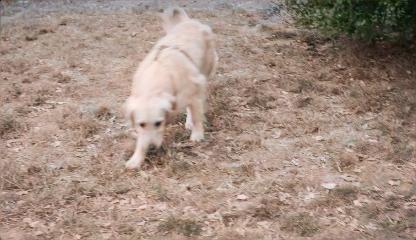}\\
& Reference & Ours: $t=10$  & Ours: $t=20$ & Ours: $t=30$\\

\end{tabular}
\vspace{1mm}
\caption{An example of video propagation via DVP. Only the first input frame has color and our method is able to propagate the color information to all the other gray frames.}
\label{fig:vpn_Teaser}
\end{figure*}

\noindent \textbf{Analysis.}
We analyze a toy example in Fig.~\ref{fig:DVP_toy} to study deep video prior. Consider 8 consecutive video frames as input $\{I_t\}_{t=1}^8$, and the processed frames $\{P_t\}_{t=1}^8$ are obtained by adding noise on ground truth. 
Two kinds of inconsistency are synthesized. The first row is illustrated for unimodal inconsistency: $\{P_t\}_{t=1}^8$ are close to each other but have small distance; the second and third rows are illustrated for multimodal inconsistency:  $\{P_t\}_{t=1}^8$ are separated to two clusters where the distance between two clusters is large. 
In Fig.~\ref{fig:DVP_toy}(a), in the beginning, i.e., 100th iteration, the outputs for consecutive frames are highly overlapped, which means $\{O_t\}_{t=1}^8$ are consistent with each other. At 200th iteration, they begin to separate from each other and the distance between $\{O_t\}_{t=1}^8$ becomes larger. After 1000 iterations, $\{O_t\}_{t=1}^8$ are not consistent anymore and are quite similar with $\{P_t\}_{t=1}^8$. 
For unimodal inconsistency at 100th iteration, $\{O_t\}_{t=1}^8$ are both close to ground truth and consistent with each other. In practice, we adopt this phenomenon to solve the unimodal inconsistency problem. 
However, for multimodal inconsistency shown in Fig.~\ref{fig:DVP_toy}(b), although $\{O_t\}_{t=1}^8$ is consistent at 100th iteration, they are not close to either ground truth. To address this problem, we propose an IRT strategy in Section~\ref{sec:IRT}. 
In Fig.~\ref{fig:DVP_toy}(c), after applying our IRT, we can obtain the results which are both consistent and close to one ground truth at a stage (100th and 200th iteration).

\subsection{Iteratively Reweighted Training (IRT)} 
\label{sec:IRT}
We propose an iteratively reweighted training (IRT) strategy for multimodal inconsistency because it cannot be solved by our basic architecture easily. Since the difference between multiple modes can be quite large, averaging different modes can result in a poor performance, which is far from any ground-truth mode, as shown in Fig.~\ref{fig:Illustration} and Fig.~\ref{fig:DVP_toy}. 
As a result, previous methods fail to generate consistent results~\cite{lai2018learning} or tend to degrade the original performance largely~\cite{bonneel2015blind}.

In IRT, a confidence map $C_t \in \{0,1\}^{\rm H\times W \times 1}$ is designed to choose one main mode for each pixel from multiple modes and ignore the outliers (one minor mode or multiple modes). We calculate the pixel-wise confidence map $C_{t,i}$ based on the network $\hat g(I;\theta^i)$ trained at $i$-th iteration. We increase the number of channels in the network output (e.g., six channels for two RGB images) to obtain two outputs: a main frame $O_{t,i}^{main}$ and an outlier frame $O_{t,i}^{minor}$. The confidence map $C_{t,i}$ (i.e., 
confidence map for the main mode) can then be calculated by:
\begin{align}
C_{t,i}(x)=\left\{
\begin{aligned}
1, &   \ \ d(O_{t,i}^{main}(x),P_{t}(x))<\\
& \ \ \  \ {\rm max}\{d(O_{t,i}^{minor}(x),P_{t}(x)), \delta \},\\
0, &   \ \ otherwise&
\end{aligned}
\right.
\end{align} 
where $d$ is the function to measure the distance between pixels and $\delta$ is a threshold. We use $L_1$ distance as $d$ and set $\delta$ to 0.02. For a pixel $x$, if the $P_{t}(x)$ is closer to $O_{t,i}^{main}$ compared with the other modes $O_{t,i}^{minor}$ at $i$-th iteration, the confidence will be 1. For some pixels, only a single mode exists. Hence, the distance to both $O_{t,i}^{main}$ and $O_{t,i}^{minor}$ can be small and these pixels should be selected for training. As shown in Fig.~\ref{fig:Confidence_map.}, for pixels 
that are: (1) close to both $O_{t,i}^{main}$ and $O_{t,i}^{minor}$; (2) closer to $O_{t,i}^{main}$, the confidence is high. 
The function of confidence maps is similar to cluster assignment in K-Means~\cite{macqueen1967some} when K=2 and the pixels in minor mode are similar to the outliers in iteratively reweighted least squares (IRLS)~\cite{holland1977robust}. At last, in the $(i+1)$-th iteration, the training loss function can be updated by the confidence map: 

\begin{align}
    \theta ^{i+1} = \mathop{\arg\min}_{\theta} \ \ \ 
    &L_{data}(C_{t,i} \odot O_{t,i}^{main},C_{t,i} \odot P_t)+ \nonumber \\
    &L_{data}( (1-C_{t,i}) \odot O_{t,i}^{minor},(1-C_{t,i}) \odot P_t) .
\end{align}

\noindent \textbf{Main mode selection.} In practice, we can use a specific frame (e.g., the first frame) to train the network for the main mode at the beginning of training. By doing so, we can make sure that the main outputs are close to the specific mode. 


 

\section{Video Propagation}
\label{sec:video_prop}
\subsection{Preliminaries}
The task of video propagation is for propagating information (e.g., color) from reference frames in a video to all the other video frames. Specifically, given an input frame sequence $\{I_t\}_{t=1}^T$, a subset of this sequence $\{I_{r_j}\}_{j=1}^{R}$ has corresponding processed frames $\{P_{r_j}\}_{j=1}^{R}$ where $r_j \in \{1,2,...,T\}$ and $R$ is the number of reference images. The goal of video propagation is to propagate the processed effect of reference frames $\{P_{r_j}\}_{j=1}^{R}$ to a output sequence of frames $\{O_t\}_{t=1}^T$. The processed effect can be obtained by a image processing algorithms or a natural information (e.g., color). Considering an example for video color propagation as shown in Fig.~\ref{fig:vpn_Teaser}, a sequence of gray frames $\{I_t\}_{t=1}^{30}$ and a reference frame of corresponding color frame $\{P_{1}\}$ (i.e., $r_0 = 1$ and $R=1$) are provided. Here ${I_1}$ and ${P_1}$ are the reference frame and the corresponding reference information, our goal is to obtain a colorful sequence $\{O_t\}_{t=1}^{30}$.

\begin{figure*}[t!]
\centering
\includegraphics[width=1.0\linewidth]{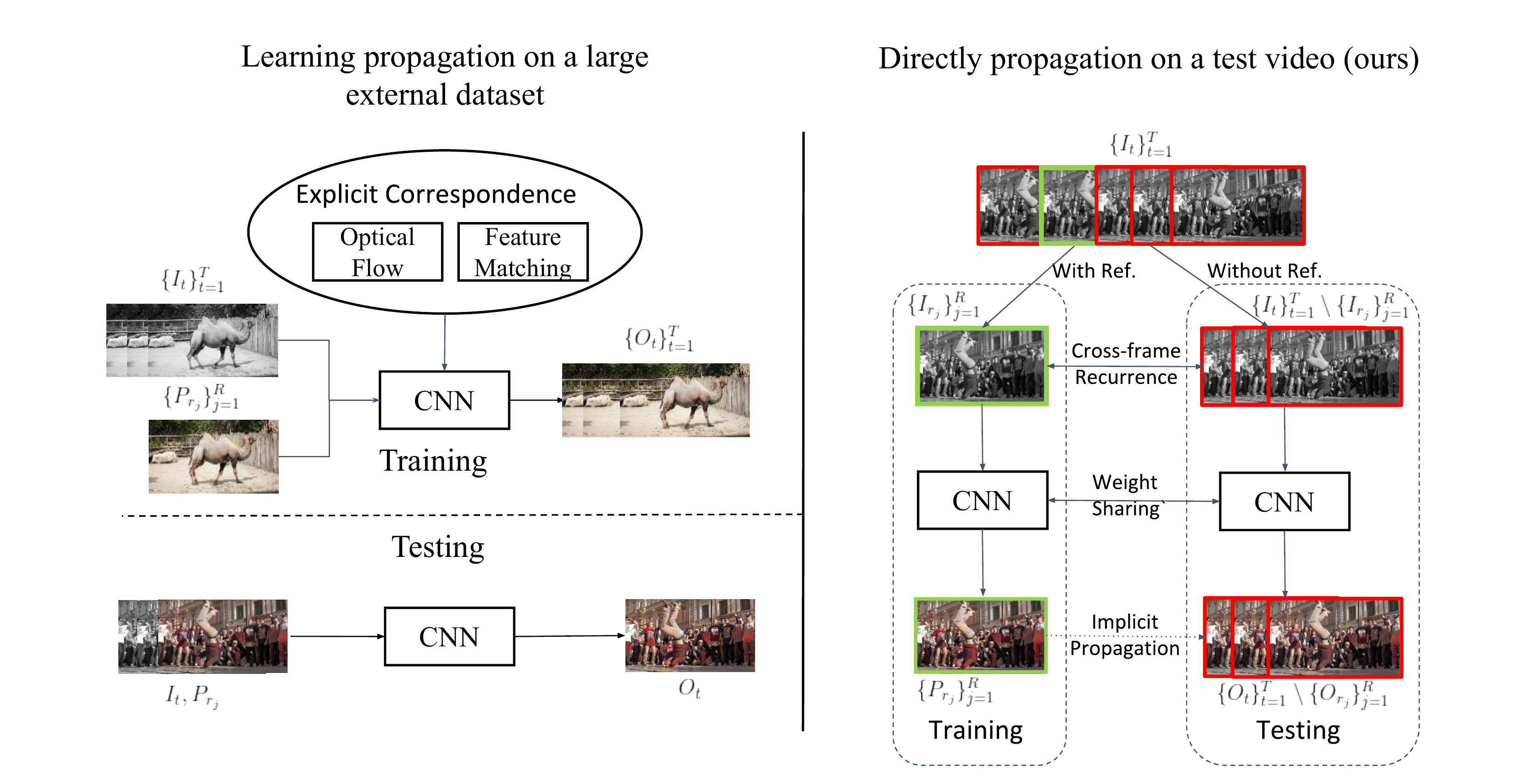}
\caption{The framework of video propagation via DVP. We only train on a few key frames and apply the trained network to infer other input images.}
\label{fig:prop_framework}
\end{figure*}

As shown in Fig.~\ref{fig:prop_framework}, previous work~\cite{jampani:cvpr:2017,DBLP:journals/tog/IizukaS19} in video propagation usually propagates the processed effect based on the similarity between reference frames and target frames. The similarity can be evaluated through the distance in the bilateral space~\cite{jampani:cvpr:2017} or  source-reference attention~\cite{DBLP:journals/tog/IizukaS19}. Different from them, we do not find any explicit correspondence by using the deep video prior.

\subsection{Video propagation via DVP}
Deep Video Prior can be further utilized in video propagation. Due to the fact that there are many similar visual contents and features in consecutive frames of a video, we argue that the propagation effect in previous work can be implicitly learned from DVP. As introduced in Sec.~\ref{sec:DVP}, we observe that similar input tends to generate similar output by a CNN. Using this property, we can obtain a temporally consistent video training on the whole sequence in Sec.~\ref{sec:BTC}. In video propagation, although the number of paired frames is much less than that in blind video temporal consistency, we find it is still possible to accurately learn the mapping function between input frames and processed frames. In other words, if we can learn the correct mapping function from reference images (e.g., contents and features) to reference information, it should be useful for the other frames.


As shown in Fig.~\ref{fig:prop_framework}, we propose to use a CNN $h(\cdot;\theta)$ to learn the a mapping function $f$ from pairs of reference input frames $\{I_{r_j}\}_{j=1}^{R}$ and reference processed frames $\{P_{r_j}\}_{j=1}^{R}$. This CNN $h(\cdot;\theta)$ is only trained on reference frames instead of a large dataset. Also, only a single frame is used in each iteration. We initialize $h$ randomly, and then the parameters of network $\theta_h$ can be updated in each iteration with a single data term without any explicit regularization:
\begin{align}
\label{eq:video_prop}
     \mathop{\arg\min}_{\theta} \ \ L_{data}(h(I_{r_j};\theta_h), P_{r_j}),
\end{align}
where $L_{data}$ measures the distance (e.g., $L_1$ distance) between $h(I_{r_j};\theta_h)$ and $P_{r_j}$. 

Our framework for video propagation is flexible and simple. Our model can adopt arbitrary numbers (i.e., one frame or multiple frames) of referenced frames to propagate the processed information. As a comparison, previous work~\cite{jampani:cvpr:2017,meyer2018deep} usually has a limitation: it is hard for them to use multiple reference frames at the same time. 

\begin{algorithm}[t]
  \renewcommand{\algorithmicrequire}{\textbf{Input:}}
  \renewcommand{\algorithmicensure}{\textbf{Output:}}
  \caption{Progressive propagation with pseudo labels}
  \label{alg1}
  \begin{algorithmic}[1]
    \REQUIRE $\{I_t\}_{t=1}^T$,$\{P_1\}$, $h(\cdot;\theta_h)$, $K$
    \ENSURE $\{O_t\}_{t=1}^T, \{Q_t\}_{t=1}^T$
    \STATE $Q \gets \{(I_1,P_1)\}$ 
    \FOR{$next:=2$ \TO $T$}
    \STATE Train $h$ with $Q$ for $K$ iterations
    \STATE Push $(I_{next}, h(I_{next};\theta_h))$ to $Q$
    \ENDFOR
  \end{algorithmic}  
\end{algorithm}

\subsection{Progressive Propagation via Pseudo Labels}
\begin{figure}[t]
\centering
\begin{tabular}{@{}c@{\hspace{1mm}}c@{}}
&
\includegraphics[width=0.96\linewidth]{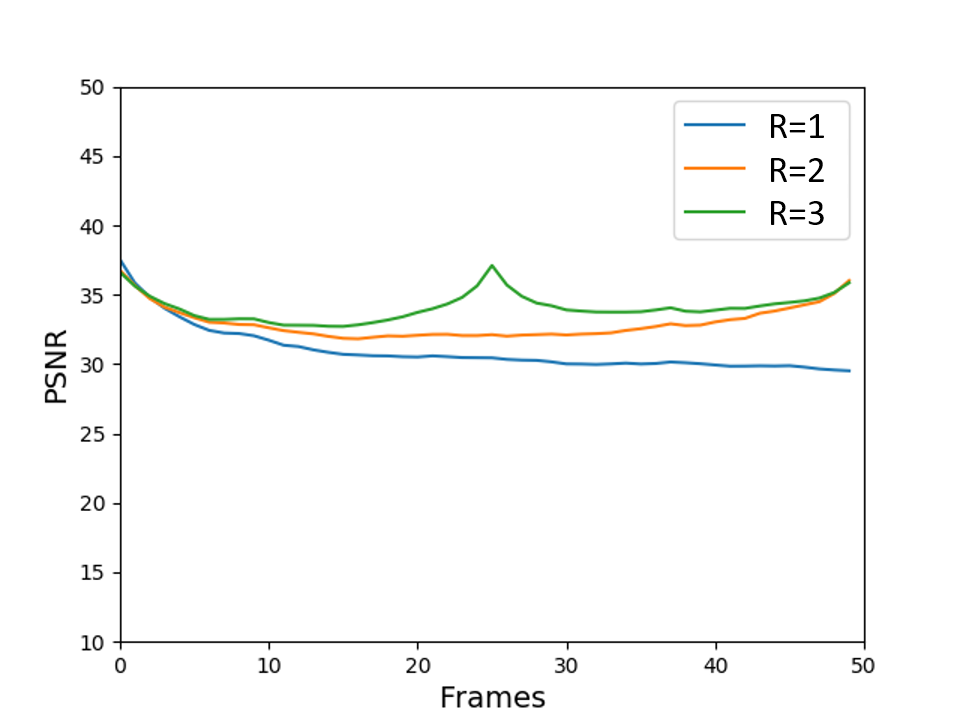}\\

\end{tabular}
\caption{The PSNR on target images decreases with the distance to the reference frame in video color propagation. The PSNR is calculated in the RGB space.}
\label{fig:Analysis_MetricsFrames}
\end{figure}

The performance of video propagation is affected by the similarity between the target images and the reference image. We conduct a controlled experiment to analyze this phenomenon. Specifically, we choose 50 frames for color propagation (i.e., $T=50$). In the controlled experiment, we set $R$ equal to 1, 2, or 3. When $R=1$, we use the first frame as the reference image with the ground-truth color information. When $R=2$, the first and last frames are provided. When $R=3$, the 1st, 25th, 50th frames are provided. After training on the reference images, we test the model on all the frames, and we use PSNR to evaluate the propagation performance. As shown in Fig.~\ref{fig:Analysis_MetricsFrames}, in all three cases, the PSNR decreases as the distance between the target and reference frames increases.

Training on reference images can only utilize the similarity between reference images and target images, leading to severe performance degradation in Fig.~\ref{fig:Analysis_MetricsFrames}. To solve this problem, we propose a progressive propagation strategy to utilize the similarity between all video frames. Specifically, we maintain a memory queue $Q$ and train the network with data in the queue. Let us take the case when there is only one reference image as an example. As shown in Algorithm~\ref{alg1}, the queue is initialized with the reference image $Q=\{(I_1,P_1)\}$ where $I_1$ is the first frame and $P_1$ is the reference processed image for $I_1$. In each step, we first train the network with all the images in the queue for $K$ iterations and then push the next video frame $I_{next}$ and the network output on this image $(I_{next}, h(I_{next}; \theta_h))$ into the queue where $h(I_{next};\theta_h)$ serves as the pseudo label for future training. The whole training process stops until all video frames are pushed into the queue.

\subsection{Implementation}
The designs of the network architecture and the loss function affect the performance of DVP. A specific architecture may be better for a specific task. In practice, we choose u-net in our experiments. Similar to the choice of network architecture, a task-specific loss might be better. For example, the perceptual loss~\cite{johnson2016perceptual} is better for style transfer compared with $L_1$ loss. 



In practice, we notice that implementing data augmentation is vital to achieving satisfying propagation performance because the number of training data is extremely limited (e.g., only a reference frame). In our experiments, we adopt rotation, flipping, and random cropping for data augmentation for video propagation.

\section{Experiments}
We first evaluate our framework through the seven tasks in our experiments.

\noindent \textbf{Colorization.}
A gray image can be colorized through the single image colorization algorithm~\cite{iizuka2016let,zhang2016colorful}. Multimodal inconsistency appears when a gray video is processed since there are multiple possible colors for one gray input. 

\noindent \textbf{Dehazing.}
Single image dehazing aims at removing haze to recover the clear underlying scenes in an image. However, directly applying an image dehazing algorithm (e.g., He et al.~\cite{he2010single}) to videos might lead to high-frequency inconsistency artifacts. 

\noindent \textbf{Image enhancement.}
We utilize DBL~\cite{gharbi2017deep} as the image enhancement algorithm. Their results reveal high-frequency flickering artifacts in videos. 

\noindent \textbf{Style transfer.}
Though image style transfer (e.g., WCT~\cite{li2017universal}) has achieved excellent performance, applying style transfer to videos is quite challenging because new content is generated in each frame. 

\noindent \textbf{Image-to-image translation.}
When applying the pretrained CycleGAN~\cite{CycleGAN2017} model for image-to-image translation on videos, inconsistent artifacts appear due to newly generated textures. 

\noindent \textbf{Intrinsic decomposition.}
Intrinsic decomposition~\cite{bell2014intrinsic} aims at decomposing each image $I$ into reflectance $R$ and shading $S$, which satisfies $I=R\times S$. Following Lai et al.~\cite{lai2018learning}, $R$ and $S$ are processed, respectively. The inconsistency in this task is relatively large.

\noindent \textbf{Spatial white balancing.}
White balance is a crucial task to eliminate color casts due to differing illuminations. When applying a single image white balance algorithm~\cite{hsu2008light} to videos, we notice that multimodal inconsistency appears.

\subsection{Experimental setup}




\begin{table*}[t]
\small
\centering
\renewcommand{\arraystretch}{1.2}
\caption{Our method achieves comparable numerical performance compared with Bonneel et al.~\cite{bonneel2015blind} and Lai et al.~\cite{lai2018learning}. Note that these metrics do not completely reflect the visual quality of output videos.}
\label{table:MainComparison}
\begin{tabular}{l|cccc|ccc}
\hline
Task &\multicolumn{4}{c}{$E_{warp} \downarrow$} & \multicolumn{3}{c}{$F_{data} \uparrow$}\\
              & {Processed} & {~\cite{bonneel2015blind}} & {~\cite{lai2018learning}}&  {Ours} & \small{\cite{bonneel2015blind}} &  {\small ~\cite{lai2018learning}}&  {Ours} \\ \hline

Dehazing~\cite{he2010single}  & 
\underline{0.139}&  0.150&  0.149&  \textbf{0.127}&
24.67&\underline{24.79} & \textbf{30.44}\\
Enhancement~\cite{gharbi2017deep}/ expertA &  
0.197 &\textbf{0.176}&  0.183&  \underline{0.179}&
23.98&  \textbf{27.77} &  \underline{25.77}\\

Enhancement~\cite{gharbi2017deep}/ expertB & 
0.188&  \underline{0.177}&  0.180&  \textbf{0.175}&
25.23&  \underline{28.46}&  \textbf{28.81}\\

Spatial White Balancing~\cite{hsu2008light}  &
{0.158}&{0.149}&  \underline{0.147}&  \textbf{0.139}&
21.80&  \underline{24.70}&  \textbf{27.80}\\

Colorization/ Iizuka et al.~\cite{iizuka2016let}& 
{0.181}&  \textbf{0.170}& \underline{0.173}&  {0.174}&

26.50&  \textbf{30.65}&\underline{30.19}  \\

Colorization/ Zhang et al.~\cite{zhang2016colorful} & 
{0.187} & \textbf{0.172} &  {0.180} & \underline{0.175} & 
{24.31} &  \textbf{29.90} & \underline{28.20}\\

CycleGAN~\cite{CycleGAN2017}/ ukiyoe & 
0.224&  \underline{0.161}&  0.164&  \textbf{0.159}&
22.04&  \textbf{26.90}& \underline{25.09}\\

CycleGAN~\cite{CycleGAN2017}/ vangogh & 
0.215&  \textbf{0.192}& 0.202&  \underline{0.193}&

21.17&  \textbf{26.53}& \underline{25.80}\\

Intrinsic~\cite{bell2014intrinsic}/ reflectance &
{0.211} & \textbf{0.160} &  {0.176} & \underline{0.164} & 
{21.75} &  \underline{24.37} & \textbf{24.97}\\

Intrinsic~\cite{bell2014intrinsic}/ shading &
{0.204} & \underline{0.158} &  {0.175} & \textbf{0.152} & 
{21.75} &  \underline{23.48} & \textbf{24.61}\\

Style Transfer~\cite{li2017universal}/ antimo. & 
{0.280} & \underline{0.242} &  {0.253} & \textbf{0.235} & 
{15.89} &  \underline{24.14} & \textbf{24.43}\\

Style Transfer~\cite{li2017universal}/ candy & 
{0.277} & \underline{0.242} &  {0.251} & \textbf{0.234} & 
{14.93} &  \textbf{23.53} & \underline{22.47}\\

\hline
Average Score
& {0.2051} & \underline{0.1791} &  {0.1860} & \textbf{0.1755} & 
{22.00} &  \underline{26.27} & \textbf{26.55}\\

\hline

\end{tabular}
\vspace{1mm}

\end{table*}

\begin{table}[t]
\renewcommand{\arraystretch}{1.2}
\centering
\caption{Comparisons among our method and other blind temporal consistency methods: Bonnel et al.~\cite{bonneel2015blind}, Lai et al.~\cite{lai2018learning}. The temporal consistency and data fidelity are based on evaluation results in Section~\ref{sec:results}.} 
\label{table:BaselinesComparison}
\begin{tabular*}{0.45\textwidth}{c@{\hspace{4mm}}c@{\hspace{4mm}}c@{\hspace{4mm}}c@{\hspace{4mm}}c@{\hspace{4mm}}}
\hline 
& \small{Training}  &  \small{Optical}&  \small{Temporal} & \small{Data} \\ 
& \small{dataset}  &  \small{flow} &   \small{consistency} & \small{fidelity} \\ 
\hline

{~\cite{bonneel2015blind}}   & \textbf{No} & {Yes} &  \underline{Very good} & {Good}\\ 

{~\cite{lai2018learning}}   & {Yes} & {Yes} &  {Good} & \underline{Very good}\\ 

{Ours}    & \textbf{No} & \textbf{No} & \textbf{Best} & \textbf{Best}\\ 
\hline
\end{tabular*}
\vspace{1mm}
\end{table}

\begin{table*}[h]
\centering
\caption{On average, our results are significantly preferred in the user study on blind temporal consistency.}
\label{table:UserStudy}
\renewcommand{\arraystretch}{1.2}
\begin{tabular*}{\textwidth}{c@{\hspace{4mm}}c@{\hspace{4mm}}c@{\hspace{4mm}}c@{\hspace{4mm}}c@{\hspace{4mm}}c@{\hspace{4mm}}c@{\hspace{4mm}}c@{\hspace{4mm}}c}
\hline
& \small{Colorization}  & \small{CycleGAN}& \small{Enhancement} & \small{Intrinsic} & \small{WhiteBalance} & \small{StyleTransfer}& \small{Dehazing}& \small{Average} \\ 
& {~\cite{zhang2016colorful,iizuka2016let}}  & {~\cite{CycleGAN2017}}& {~\cite{he2010single}} & {~\cite{gharbi2017deep}} & {~\cite{hsu2008light}}& {~\cite{li2017universal}}& {~\cite{bell2014intrinsic}} &  \\ 
\hline
{Processed}   & {7\%} & {6\%} & 
{12\%} & {3\%} & {0\%}& {1.5\%}& {5\%}& 
{5\%} \\ 
{~\cite{bonneel2015blind}}   & {36\%} & {24\%} & 
{19.5\%} & {14\%} & {16\%}& {23.5\%}& {25\%}& 
{23\%} \\ 
{~\cite{lai2018learning}}   & {16.5\%} & {10.5\%} & 
{34\%} & {18.5\%} & {6\%}& \textbf{40.5\%}& {5\%}& 
{18.71\%} \\ 
{Ours}    & \textbf{40.5\%} & \textbf{59.5\%} & 
\textbf{34.5\%} & \textbf{64.5\%} & \textbf{78\%}& 34.5\%& \textbf{65\%}& 
\textbf{53.79\%} \\ 
\hline
\end{tabular*}
\vspace{1mm}
\end{table*}

\textbf{Baselines.} 
We mainly choose two state-of-the-art methods~\cite{lai2018learning,bonneel2015blind} whose source codes or test results are available. Some characteristics of baselines are listed in Table~\ref{table:BaselinesComparison}. Compared with previous approaches~\cite{bonneel2015blind,lai2018learning}, we do not need training datasets or estimating optical flow. In addition to the simplicity, temporal consistency and data fidelity of our method are also quite competitive.
\begin{figure*}[t]
\centering
\begin{tabular}{@{}c@{\hspace{1mm}}c@{\hspace{1mm}}c@{\hspace{1mm}}c@{\hspace{1mm}}c@{\hspace{1mm}}c@{}}



\rotatebox{90}{\small \hspace{5mm} $t=3$ }&
\includegraphics[width=0.187\linewidth]{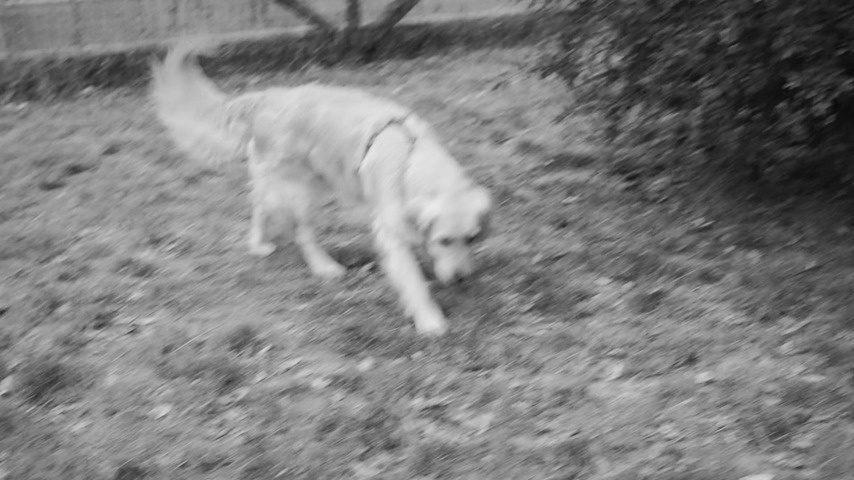}&
\includegraphics[width=0.187\linewidth]{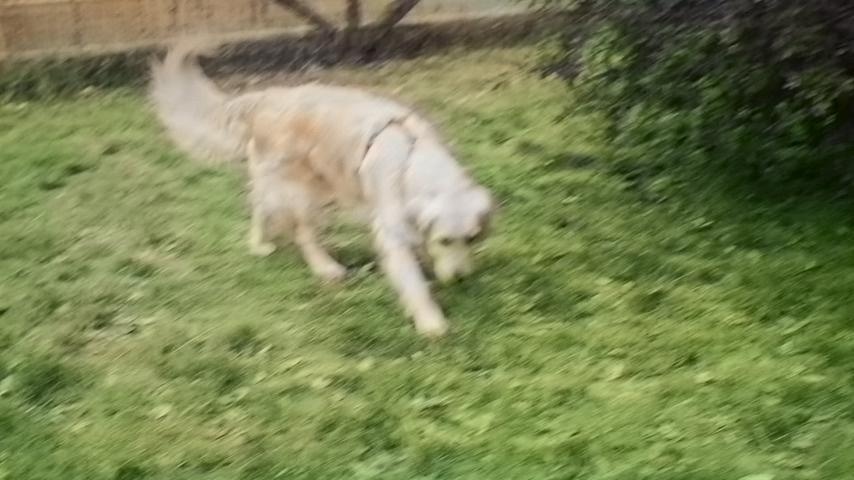}&
\includegraphics[width=0.187\linewidth]{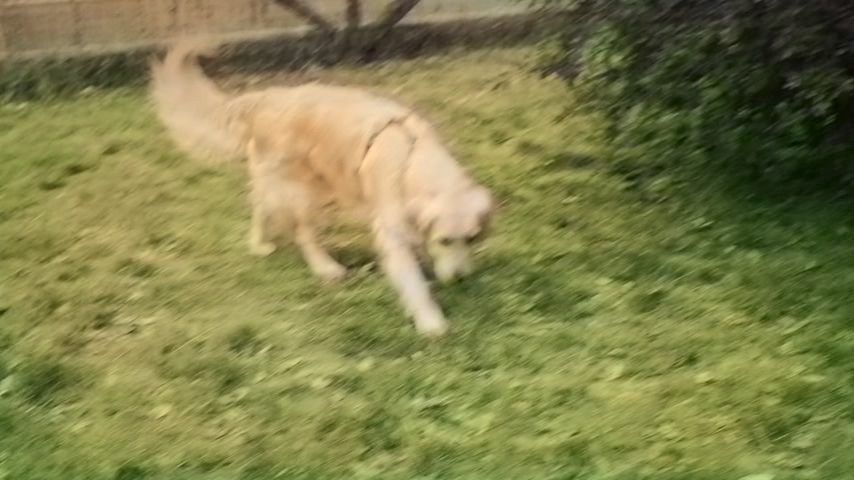}&
\includegraphics[width=0.187\linewidth]{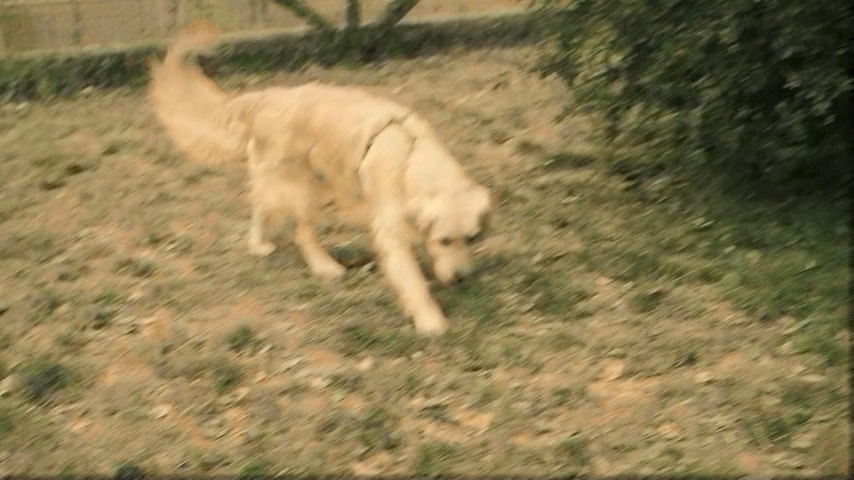}&
\includegraphics[width=0.184\linewidth]{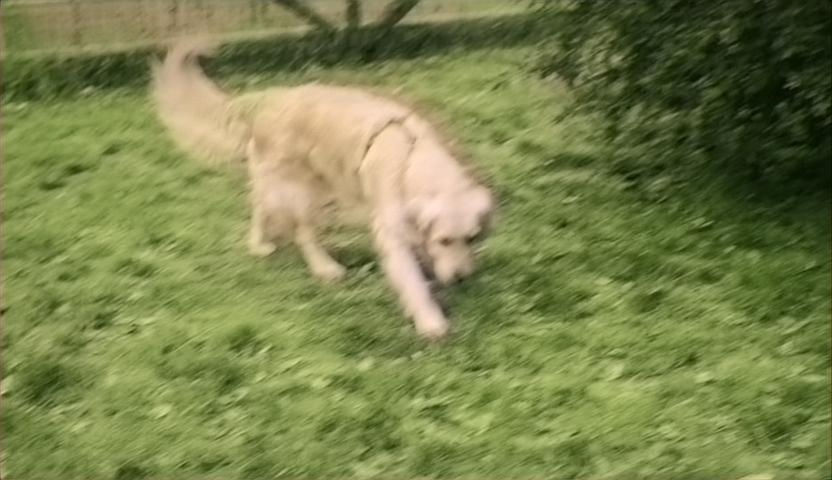}\\

\rotatebox{90}{\small \hspace{5mm} $t=24$ }&
\includegraphics[width=0.187\linewidth]{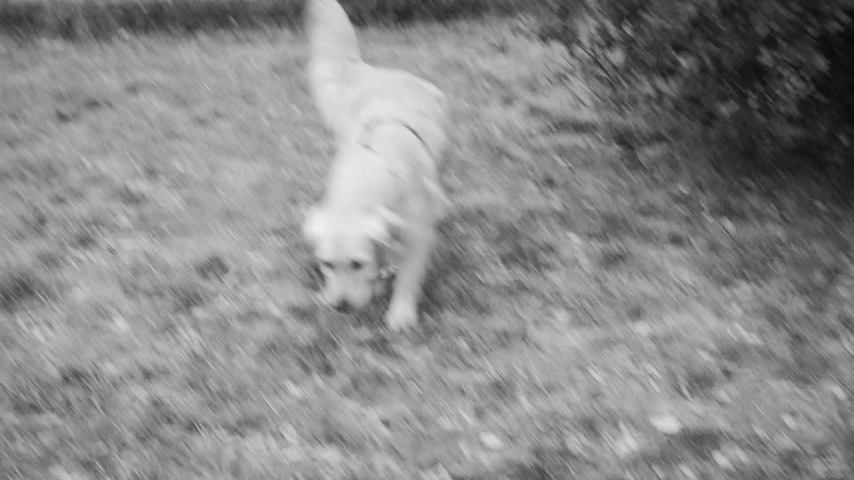}&
\includegraphics[width=0.187\linewidth]{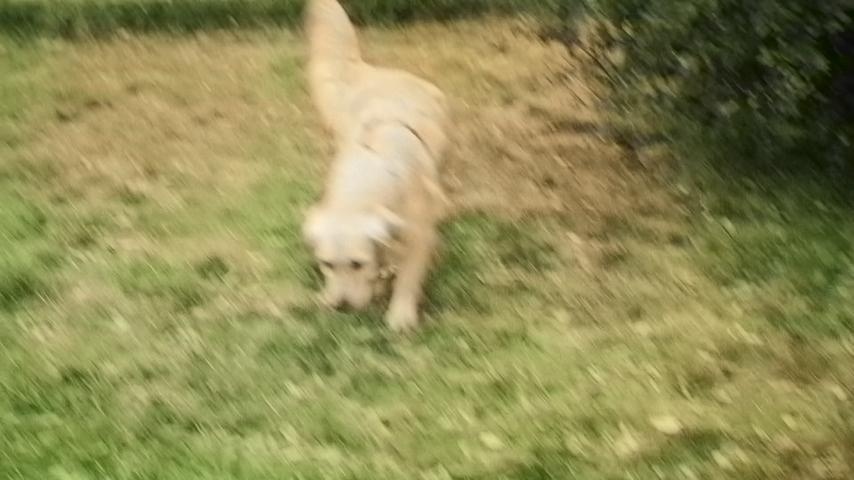}&
\includegraphics[width=0.187\linewidth]{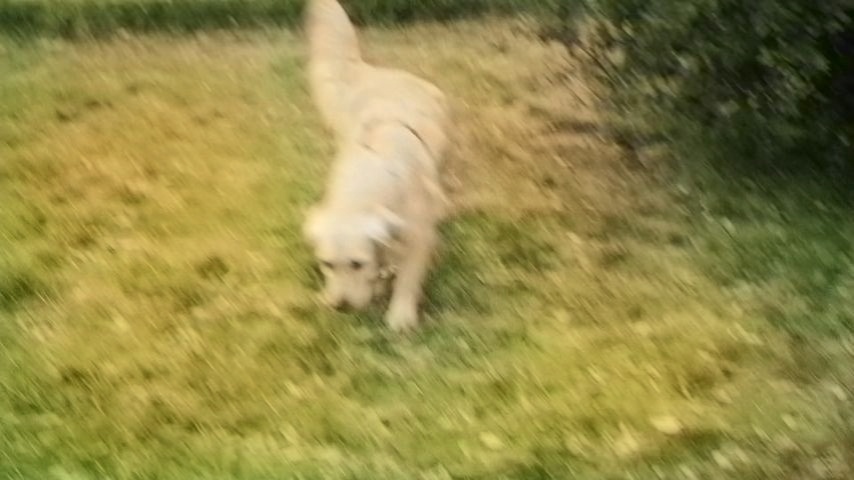}&
\includegraphics[width=0.187\linewidth]{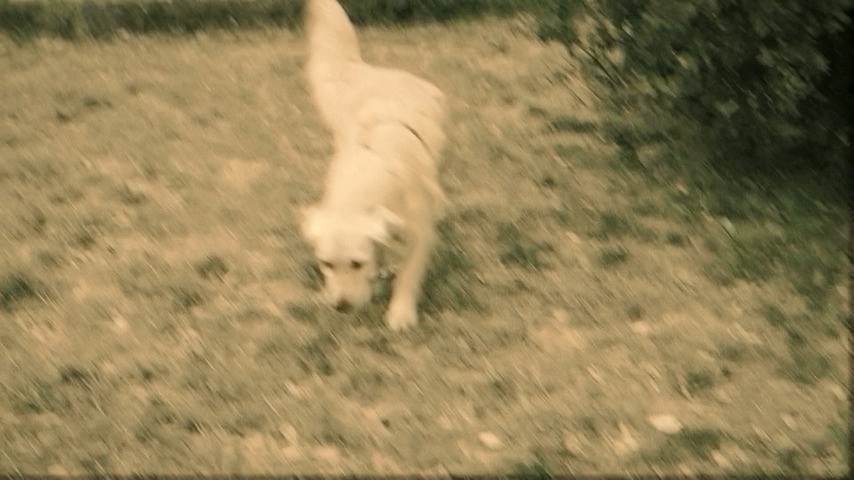}&
\includegraphics[width=0.184\linewidth]{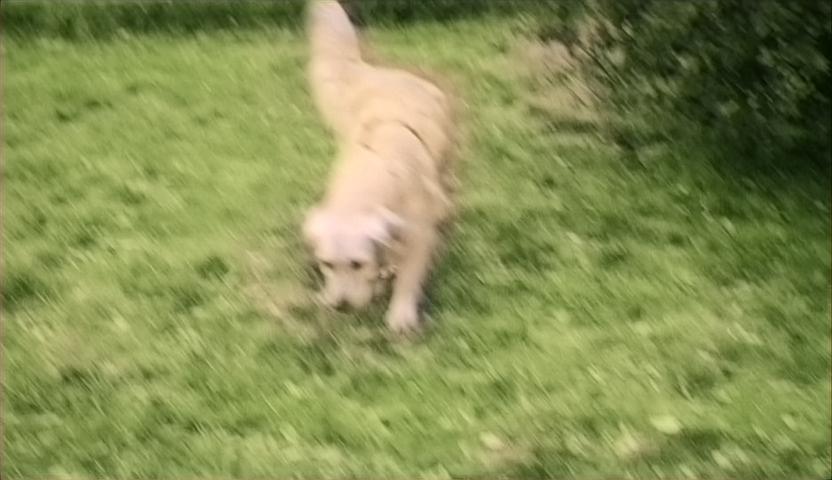}\\

\rotatebox{90}{\small \hspace{5mm} $t=0$ }&
\includegraphics[width=0.187\linewidth]{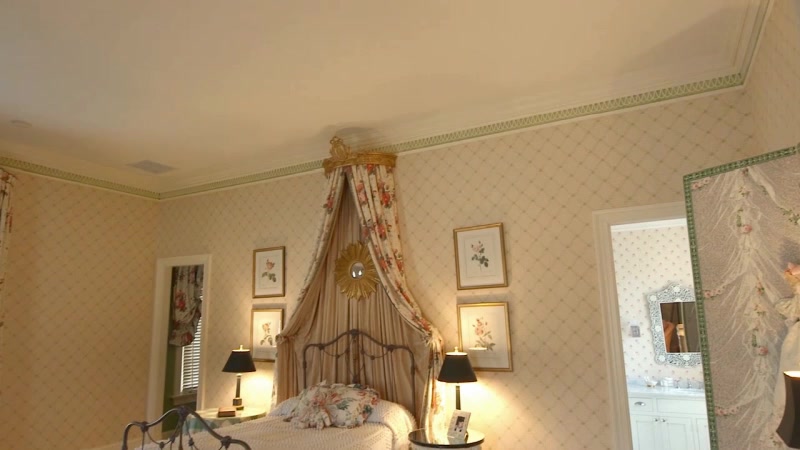}&
\includegraphics[width=0.187\linewidth]{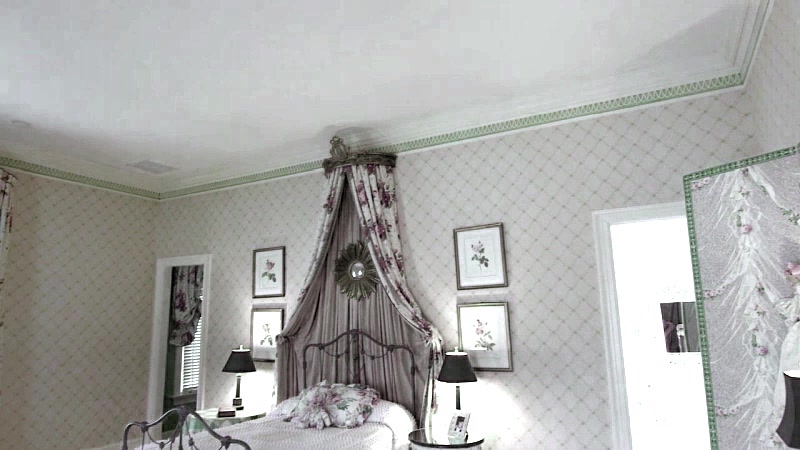}&
\includegraphics[width=0.187\linewidth]{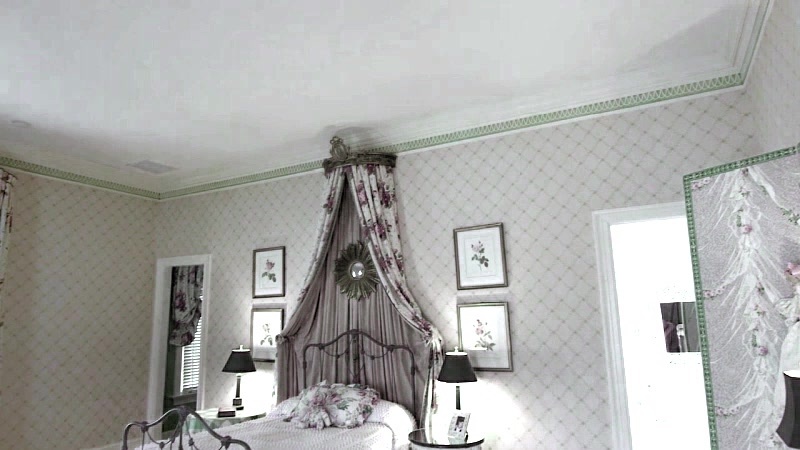}&
\includegraphics[width=0.187\linewidth]{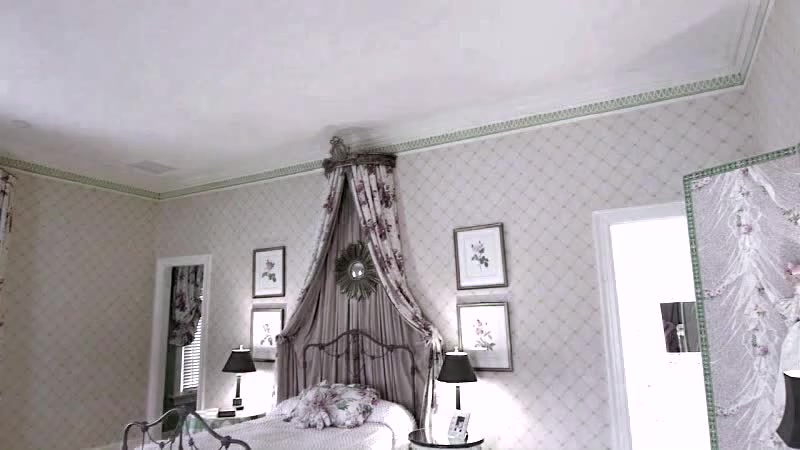}&
\includegraphics[width=0.187\linewidth]{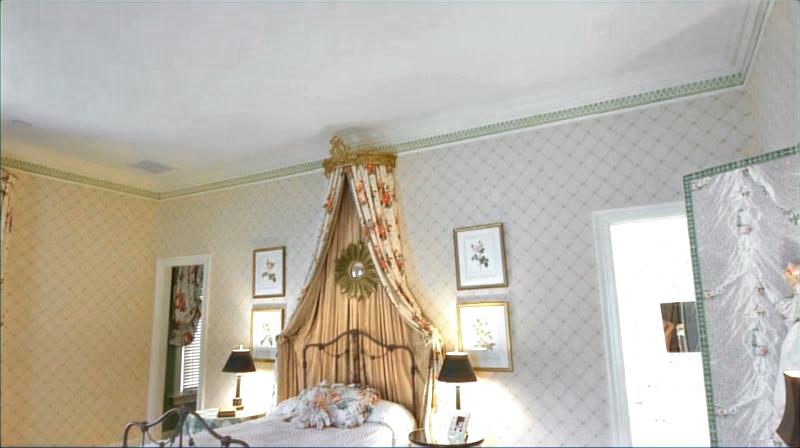}\\
\rotatebox{90}{\small \hspace{5mm} $t=80$ }&
\includegraphics[width=0.187\linewidth]{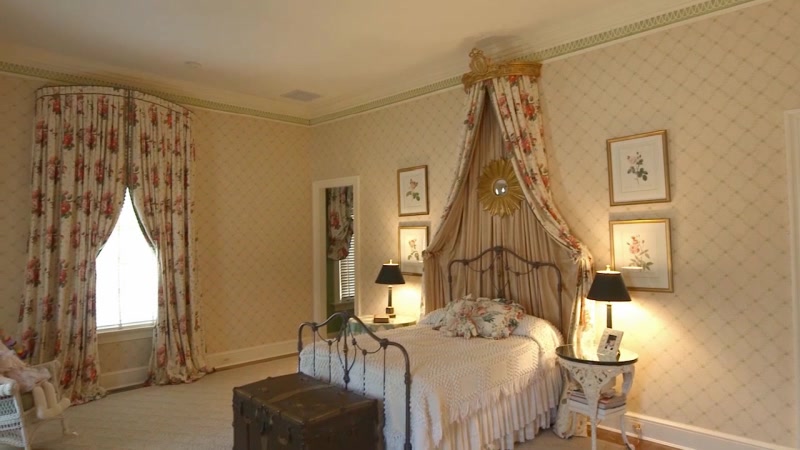}&
\includegraphics[width=0.187\linewidth]{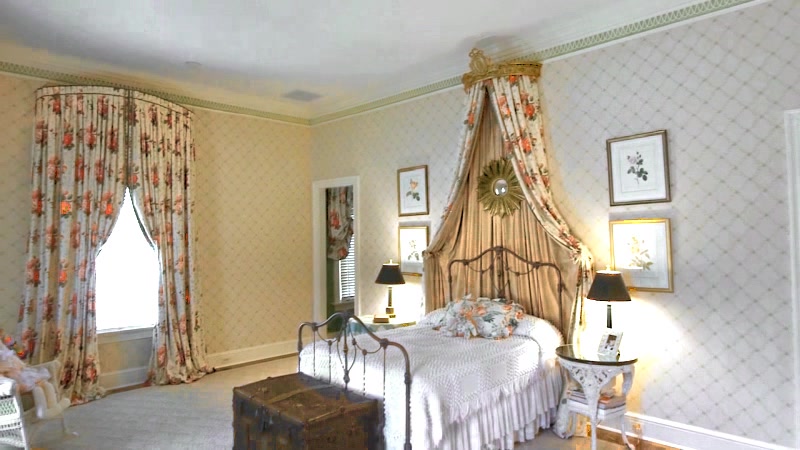}&
\includegraphics[width=0.187\linewidth]{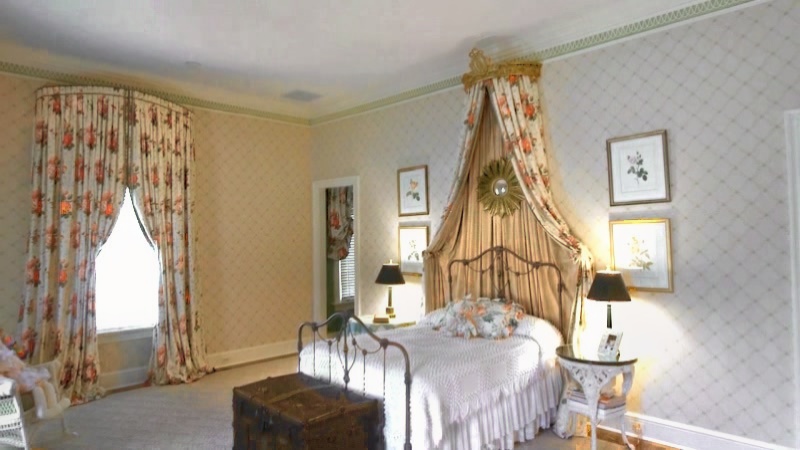}&
\includegraphics[width=0.187\linewidth]{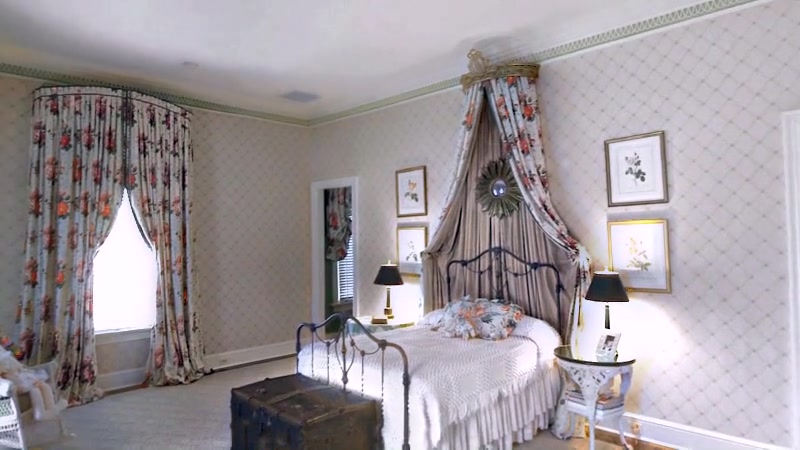}&
\includegraphics[width=0.187\linewidth]{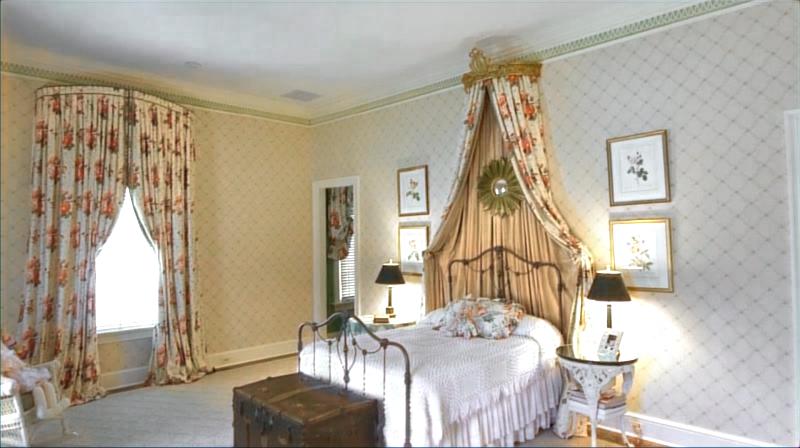}\\

&Input & Processed & \small{Lai et al.~\cite{lai2018learning}}  & \small{Bonneel et al.~\cite{bonneel2015blind}} & \small{Ours} \\

\end{tabular}
\\

\caption{The input frames are processed by colorization~\cite{zhang2016colorful} and white balancing~\cite{hsu2008light} respectively. As shown in these two examples, the results of Lai et al.~\cite{lai2018learning} look similar to the processed video but fail to preserve long-term consistency. Also, the results by Bonneel et al.~\cite{bonneel2015blind} have obvious performance decay: the color is changed in an undesirable way. Our method solves the multimodal inconsistency with IRT by producing a video with long-term consistency for the main mode.}
\label{fig:LargeInconsistency_Cmp}
\end{figure*}

\begin{figure}[]
\centering
\includegraphics[width=.8\linewidth]{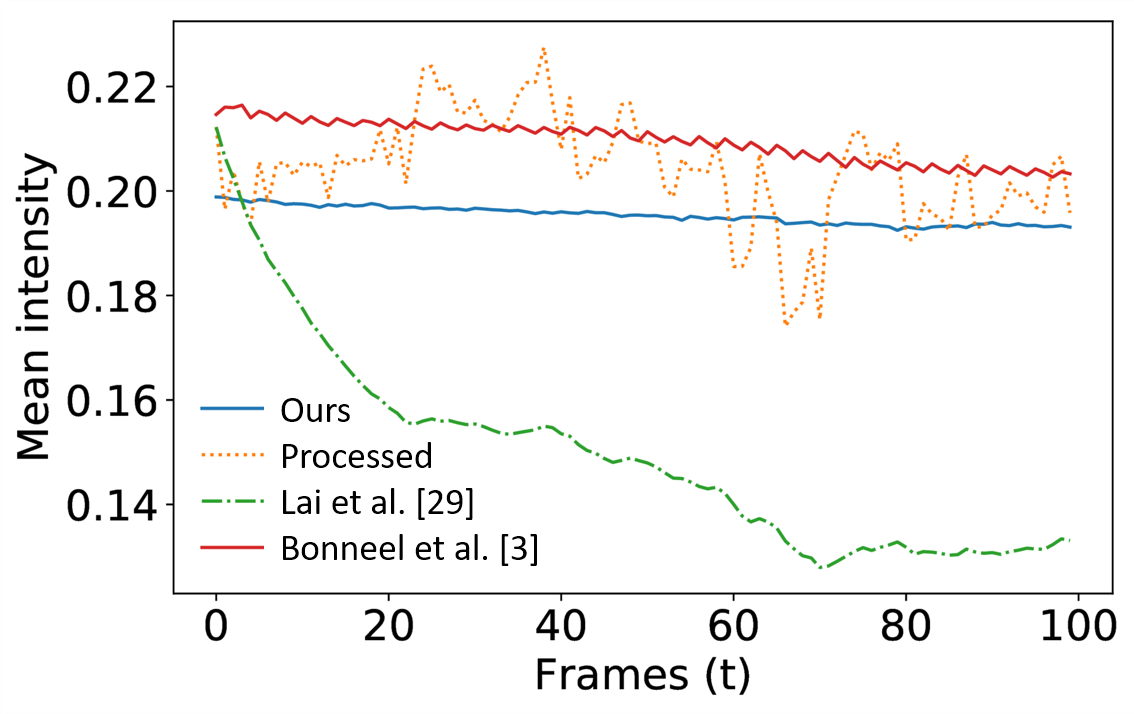}
\caption{Mean intensity of frames on dehazing~\cite{he2010single}. The video obtained by DVP is temporally consistent.}
\label{fig:Dehazing}
\end{figure}

\noindent \textbf{Datasets.} 
Following the previous work~\cite{lai2018learning,bonneel2015blind}, we adopt the DAVIS dataset~\cite{Perazzi2016} and the test set collected by Bonneel et al.~\cite{bonneel2015blind} for evaluation. The test set collected by Bonneel et al.~\cite{bonneel2015blind} is adopted for dehazing and spatial white balancing. The DAVIS dataset~\cite{Perazzi2016}, which contains 30 videos, is used for the rest applications. Since our model is trained on test data directly, we do not need to collect the training set. 

\noindent \textbf{Implementation details.} 
We use the Adam optimizer~\cite{kingma2014adam} and set the learning rate to 0.0001 for all the tasks. The batch size is 1. Dehazing, spatial white balancing, and image enhancement are trained for 25 epochs. Intrinsic decomposition, colorization, style transfer, and CycleGAN are trained for 50 epochs.





\subsection{Evaluation metrics}

\textbf{Temporal inconsistency.} The warping error $E_{warp}$ is used to measure the temporal inconsistency. For each frame $O_t$, we calculate the warping error with frame $O_{t-1}$~\cite{lai2018learning,chen2017coherent, huang2017real} and the first frame $O_{1}$~\cite{kim2019deep} for considering both short-term and long-term consistency. The final $E_{warp}$ is calculated by:
\begin{align}
    E_{pair}(O_t,O_s) &=  \frac{\sum_{i=1}^N  M_{t,s}(I_i) ||O_t(I_i) - W(O_s)(I_i)||_1}{\sum_{i=1}^N{ M_{t,s}(I_i)}} ,\\
    E_{warp}(\{O_t\}_{t=1}^T) &=   \frac{\sum_{t=2}^T\{ E_{pair}(O_t,O_1)
    + E_{pair}(O_t,O_{t-1})\}}{T-1},
\end{align}
where $M_{t,s}$ is the occlusion map~\cite{ruder2016artistic} for a pair of images $O_t$ and $O_s$, $N$ is the number of pixels, and $W$ is backward warping with optical flow~\cite{sun2018pwc}.

\noindent \textbf{Performance degradation.} 
Avoiding performance degradation is critical to blind temporal consistency. Since we do not have the ground truth videos for most evaluated tasks, we use data fidelity $F_{data}$ between $\{P_t\}_{t=1}^T$ and $\{O_t\}_{t=1}^T$ as a reference to evaluate the performance degradation. 
Note that data fidelity does not mean the perceptual performance of a video. For example, for $\{P_t\}_{t=1}^T$ with multimodal inconsistency, high-quality results $\{O_t\}_{t=1}^T$ have only a single mode, i.e., the distance can be quite large for some frames, as shown in Fig.~\ref{fig:LargeInconsistency_Cmp}. 
Since the first frame is used as a reference in baselines~\cite{lai2018learning,bonneel2015blind}, the first frame is excluded for fair comparison:
\begin{align}
\label{eq:f_data}
    F_{data}(\{P_t\}_{t=1}^T, \{O_t\}_{t=1}^T) = \frac{1}{T-1}\sum_{t=2}^T PSNR(P_t,O_t).
\end{align}

\subsection{Results} 
\label{sec:results}

\textbf{Quantitative results.} 
We show the quantitative comparison results in Table~\ref{table:MainComparison}.
We obtain the best average scores at both temporal consistency and data fidelity metrics. 
Although Bonneel et al.~\cite{bonneel2015blind} also achieve similar temporal consistency, they suffer from severe performance decay problems in some tasks, as shown in Table~\ref{table:MainComparison}. 
Lai et al.~\cite{lai2018learning} obtain comparable results for data fidelity metrics, but our temporal consistency performance is much better. We also observe that they cannot enforce long-term consistency for multimodal cases while their data fidelity performs well, and an example is illustrated in Fig.~\ref{fig:LargeInconsistency_Cmp}.

\noindent \textbf{Qualitative results.} 
More perceptual results will be presented in supplementary material due to limited space. Readers are recommended to check videos for better perceptual evaluation. 
Our performance is much better than baselines to solve the multimodal inconsistency problem. As shown in Fig.~\ref{fig:LargeInconsistency_Cmp}, the processed frames have two completely different results. 
Our framework produces temporally consistent results and also maintains the performance in one mode. 
Lai et al.~\cite{lai2018learning} fail to impose reasonable temporal regularization on both cases. The results of Bonneel et al.~\cite{bonneel2015blind} suffer from the serious performance decay problem. These qualitative results are consistent with our quantitative results. 
In Fig.~\ref{fig:Dehazing}, we compute the mean intensity of an image to evaluate the temporal consistency. 
The flickering artifact of processed frames is quite apparent. For Lai et al.~\cite{lai2018learning}, the flickering artifacts are handled well in the short term, but the difference between the first and last frame is too large. Also, although the amplitude of flickering is decreased by Bonneel et al.~\cite{bonneel2015blind}, flickering still exists. Compared with baselines, our result is consistent in both the short term and long term. 

\noindent \textbf{User study.} 
We conduct a user study to compare the perceptual preference of various methods. In total, 107 videos from all tasks are randomly selected. 
20 subjects are asked to select the best videos with both temporal consistency and performance similarity. 
As shown in Table~\ref{table:UserStudy}, our method outperforms baselines~\cite{lai2018learning,bonneel2015blind} and processed videos in most tasks. For tasks with simple unimodal inconsistency like enhancement~\cite{gharbi2017deep}, the difference between our method and Lai et al.~\cite{lai2018learning} is minor. However, for tasks with multimodal inconsistency, our method can outperform theirs to a large extent. 


\begin{figure}[]
\centering
\begin{tabular}{@{}c@{\hspace{1mm}}c@{}}
\includegraphics[width=0.475\linewidth]{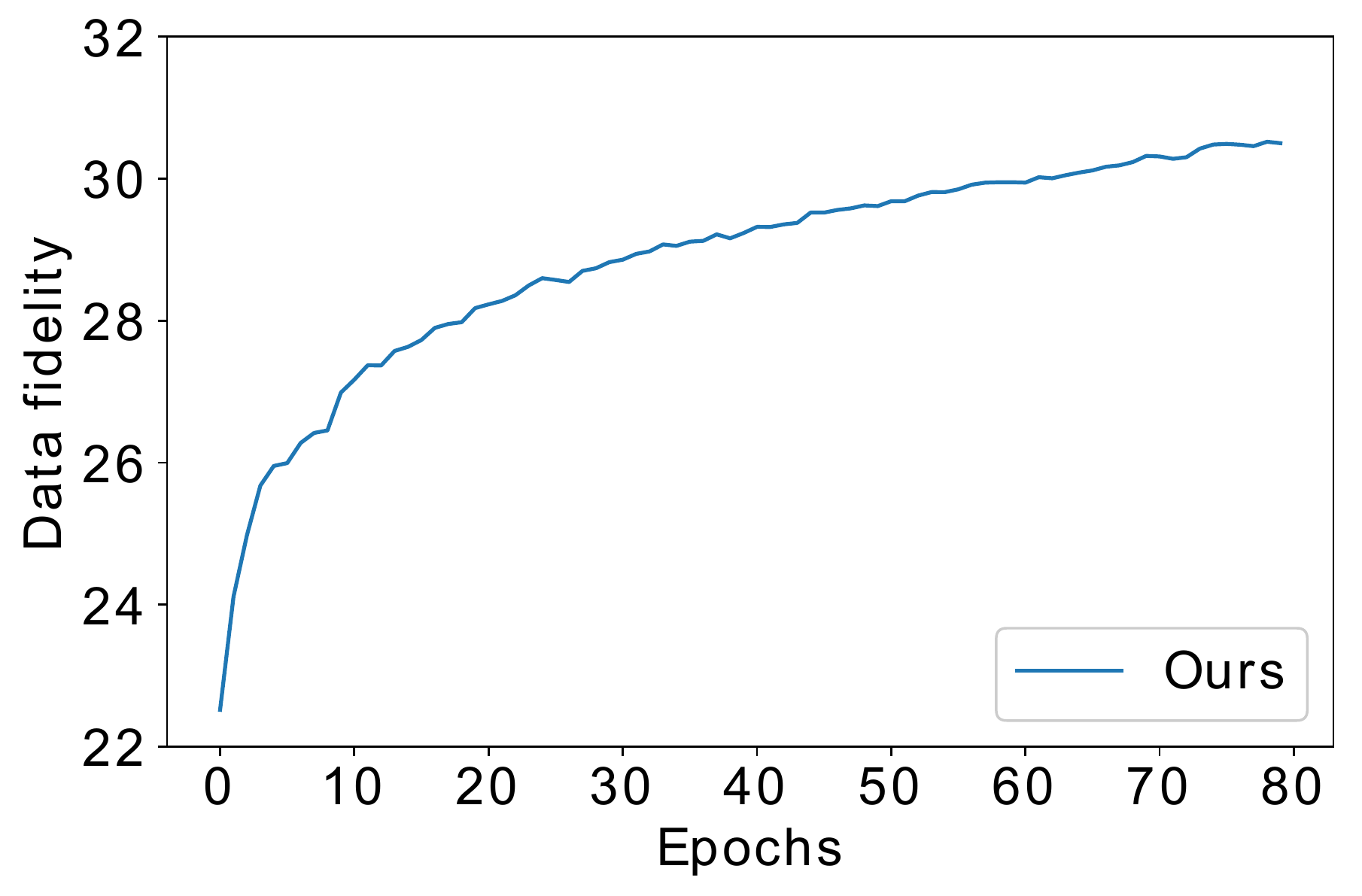}&
\includegraphics[width=0.5\linewidth]{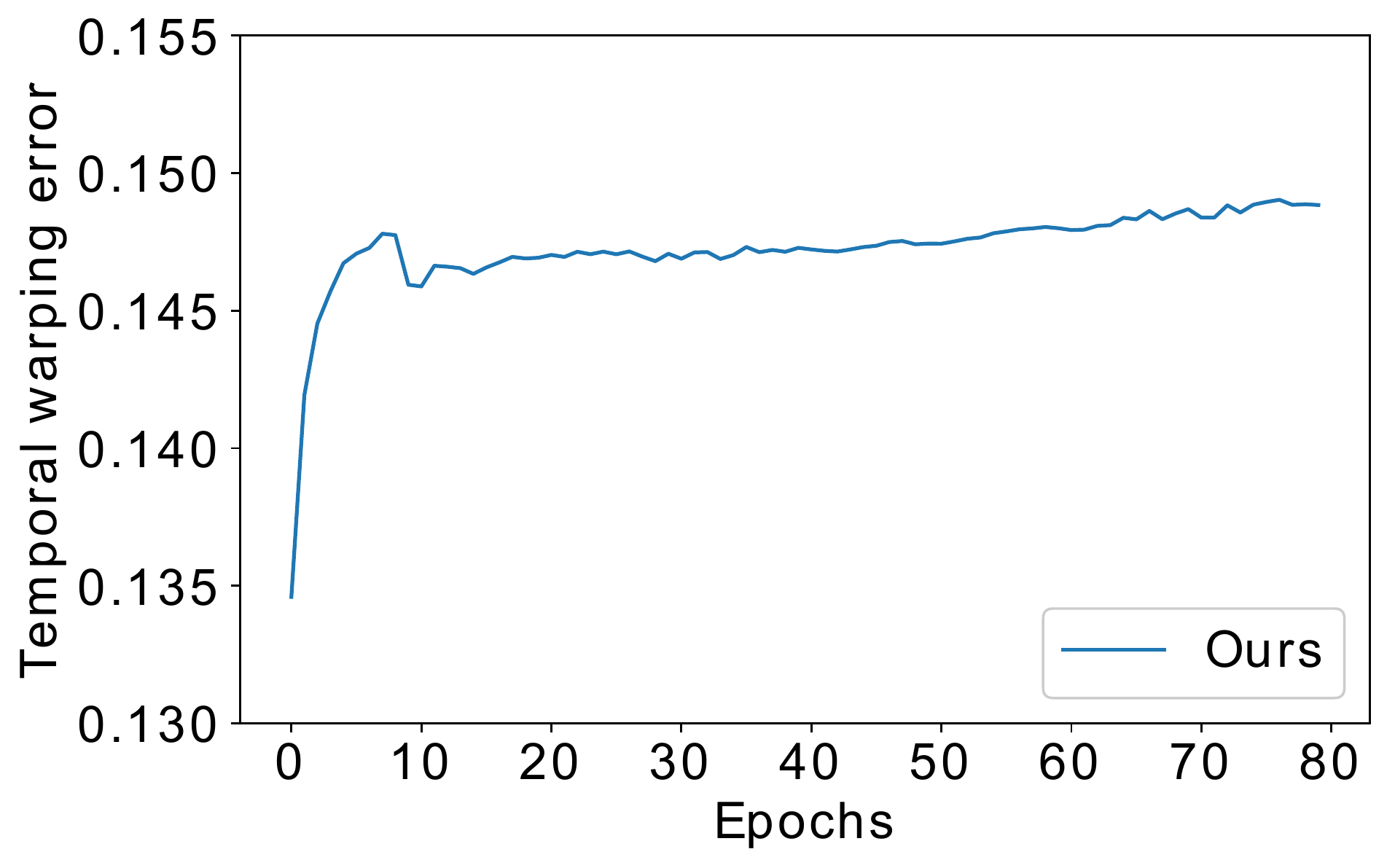}\\
(a) Data fidelity & (b) Temporal inconsistency \\
\vspace{1mm} \\
\end{tabular}

\begin{tabular}{@{}c@{\hspace{1mm}}c@{\hspace{1mm}}c@{\hspace{1mm}}c@{}}

\rotatebox{90}{\small \hspace{1mm} Processed~\cite{zhang2016colorful} }&
\includegraphics[width=0.45\linewidth]{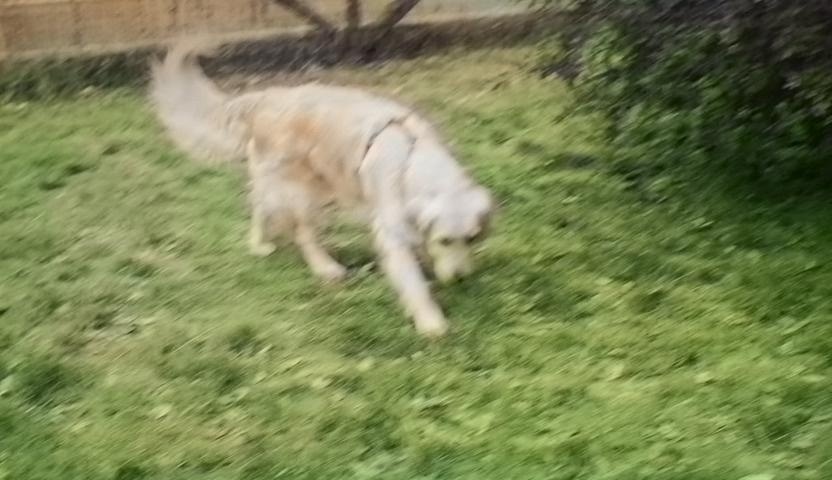}&
\includegraphics[width=0.45\linewidth]{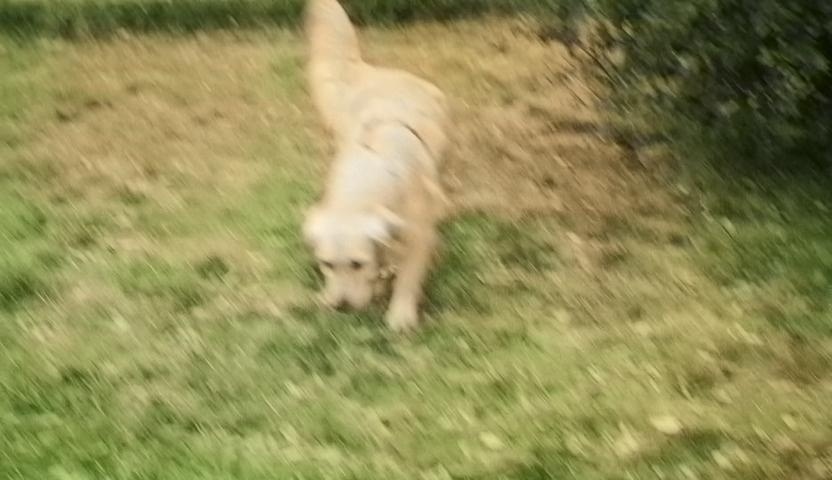}\\
\rotatebox{90}{\small \hspace{5mm} 1 epoch }&
\includegraphics[width=0.45\linewidth]{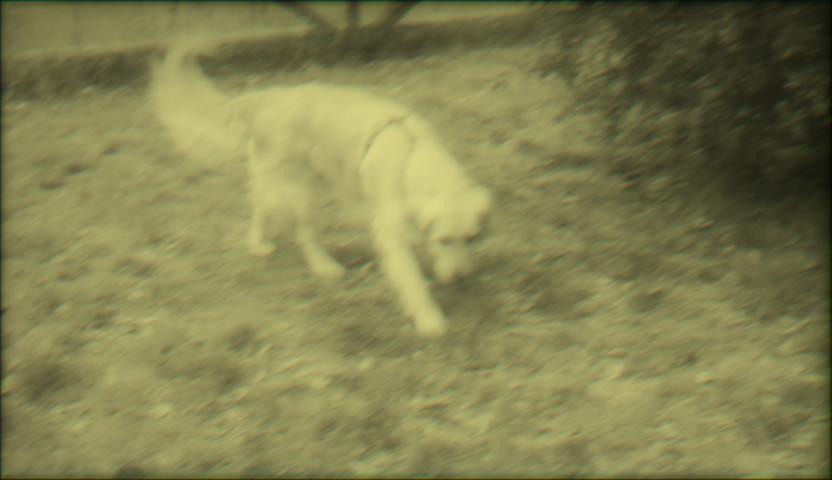}&
\includegraphics[width=0.45\linewidth]{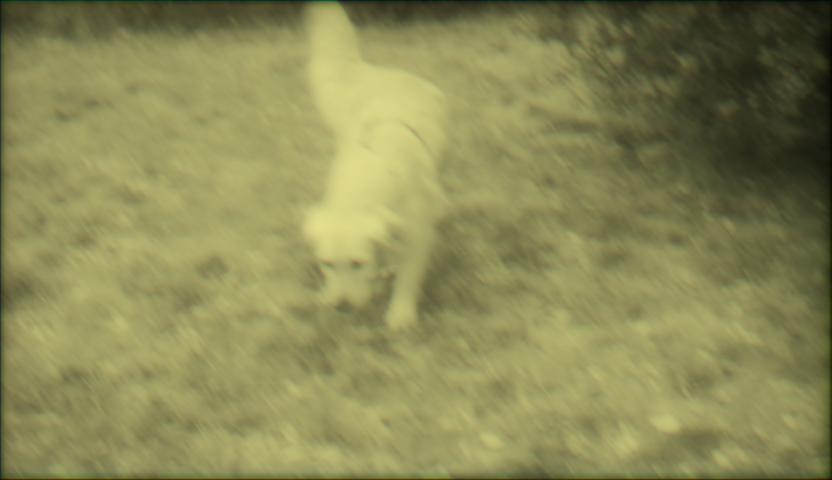}\\
\rotatebox{90}{\small \hspace{3mm} 5o epochs }&
\includegraphics[width=0.45\linewidth]{Figure/Experiments/Colorization/dog/ours/00003.jpg}&
\includegraphics[width=0.45\linewidth]{Figure/Experiments/Colorization/dog/ours/00024.jpg}\\
\rotatebox{90}{\small \hspace{2mm} 500 epochs }&
\includegraphics[width=0.45\linewidth]{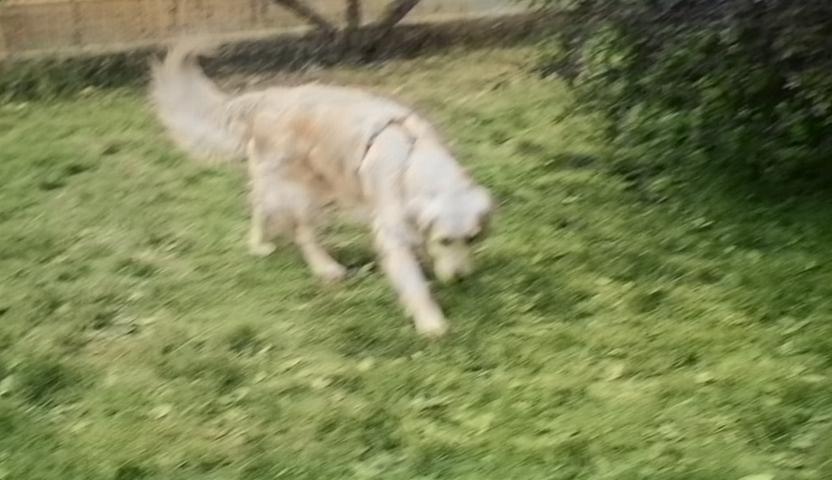}&
\includegraphics[width=0.45\linewidth]{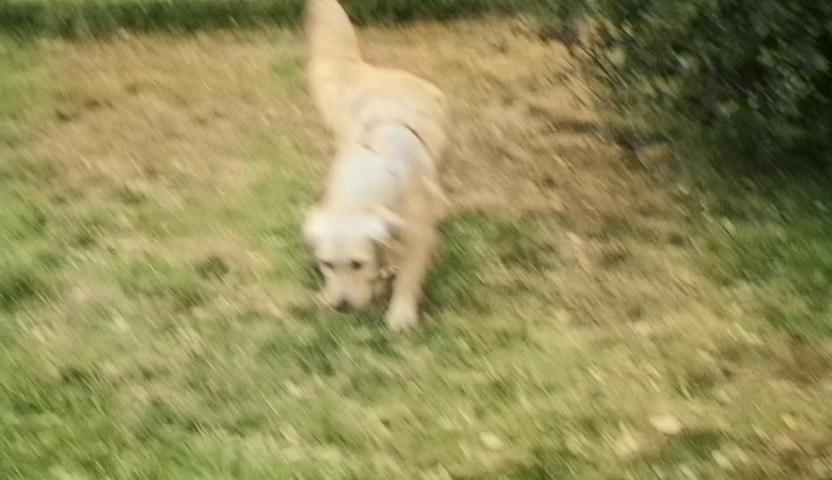}\\
& \multicolumn{2}{c}{(c) Our results with different numbers of epochs} \\

\end{tabular}





\caption{Analysis of blind temporal consistency via DVP. Both data fidelity $F_{data}$ (a) and temporal inconsistency $E_{warp}$ (b) are increasing in the process of training. (c) shows the results of our model after training for 1, 50, or 500 epochs. Our model can maintain a good balance between data fidelity and temporal consistency after training for 50 epochs. After 500 epochs, our model overfits the processed frames.}
\label{fig:Trade off}
\end{figure}

\begin{table}[t]
\small
\centering
\caption{Quantitative results of different acceleration strategies for blind temporal consistency via DVP. }
\label{table:speedup}
\renewcommand{\arraystretch}{1.2}
\begin{tabular}{l|c@{\hspace{2mm}}c@{\hspace{2mm}}c@{\hspace{2mm}}}
\hline
Task & Speed-up factor& $E_{warp} \downarrow $& $F_{data} \uparrow$  \\
\hline
{Processed~\cite{CycleGAN2017}} & - & 0.230 & -\\
\hline
{No acceleration} & $\times$ 1 & 0.218 &26.25\\
{Better initialization}&$\times$ 3& {0.217}&{26.33}\\
{Coarse-to-fine training} &$\times$  1.6&{0.217}&26.18\\
{Combined}&{$\times$ 4.8}  & {0.215}&{26.32}\\

\hline

\end{tabular}
\vspace{1mm}
\end{table}
\subsection{Training strategy analysis}
\label{sec:speedup} For an $800 \times 480$ video frame, our approach costs 80 ms for each iteration during training on Nvidia RTX 2080 Ti. For a video with 50 frames, 25 epochs (1,250 iterations) cost about 100 seconds for training and inference. In this case, the average time cost for each frame is 2 seconds.

\subsubsection{Efficient training}

We propose two practical strategies to accelerate the training process. Firstly, instead of training the network from scratch, we notice that the randomly initialized network can be replaced by a pre-trained network on specific tasks (e.g., CycleGAN/ukiyoe). This strategy can accelerate the training process significantly without sacrificing the performance obviously. Specifically, we assume the network successively goes through scratch-consistent-overfitting stages during the whole training process. In practice, the transition between the ``scratch'' to ``consistent'' stage can be largely shortened if we start the network training from an intermediate stage between these two processes. i.e., pretrained on specific tasks.

Secondly, we adopt a coarse-to-fine training strategy. We first train the network using two times downsampled pairs and then train the network with full-resolution pairs. This strategy is based on the fact that the DVP holds for different resolution video frames. We train on the two resolutions for both 50\% epochs, and it can expedite the training process to about 1.6 times faster without sacrificing the performance.


We conduct experiments to analyze the effectiveness of two strategies. We re-trained the CycleGAN model on the ukiyoe style transfer for 100 epochs to obtain the processed frames $\{P_t\}_{t=1}^T$. We select the model at the 5th epoch to replace the randomly initialized model for better initialization. We train our ``No acceleration'', ``Better initialization'' and ``Coarse-to-fine'' baseline for 15 epochs, 5 epochs, and 15 epochs respectively. Table~\ref{table:speedup} shows the results. If combining the better initialization and coarse-to-fine training strategy, we can achieve nearly five times speedup, i.e., the training time for each 800 $\times$ 400 video frame cost about 0.24 seconds in total. 

\subsubsection{When to stop training?} In our method, there is a trade-off between temporal consistency and data fidelity. This phenomenon is reported in all previous methods~\cite{bonneel2015blind,yao2017occlusion,lai2018learning,eilertsen2019single}. While Lai et al.~\cite{lai2018learning} and Eilertsen et al.~\cite{eilertsen2019single} must balance the data fidelity and temporal consistency in the process of training on a large dataset, it is much easier for us to achieve this goal since our method is trained on a single video. Fig.~\ref{fig:Trade off} shows that data fidelity and temporal inconsistency are both increasing in the process of training. Fig.~\ref{fig:Trade off}(c) shows some visual results of our model with different numbers of epochs for training.

In principle, the training epochs should be different for videos with different duration. For example, a video with 200 frames requires fewer epochs than a video with 50 frames. For two videos with the same length, the video with smaller motion requires fewer epochs. In our experiment, we observe that temporal consistency is stable in many epochs. 
For example, $E_{warp}$ is only increased by around 0.002 from 25th to 80th epoch in Fig.~\ref{fig:Trade off}. Therefore, we simply select the same epoch (25 or 50 epochs) for all the videos with different lengths (30 to 200 frames) in a task based on a small validation set up to 5 videos for all our evaluations. We do not need to carefully select the epoch because reconstructing the flickering artifacts takes much more time compared with common video content. 

To further automate our approach, we propose a simple epoch selection strategy based on the statistics of the training loss curve~\cite{ulyanov2018deep}. For each epoch during training, we calculate the average loss on all video frames. We notice that this average loss is relatively smooth over many epochs. Thus, we can record the losses of previous $k$ epochs (including the current one). Then we calculate the normalized variance: we normalize the $k$ losses with the largest one and then calculate the variance. We can stop the training when the normalized variance is smaller than an empirical threshold (1e-8 for perceptual loss). We quantitatively evaluate the performance of our epoch selection method on the CycleGAN-ukiyoe task. Table~\ref{table:epoch_selection} shows that our automatically selected epochs are quite close to the performance of manually selected results. 


%

\begin{table}[t]
\small
\centering
\caption{Quantitative results with different stopping epoch selection methods. }
\label{table:epoch_selection}
\renewcommand{\arraystretch}{1.2}
\begin{tabular}{l|ccc}
\hline
Selection Method &$F_{data} \uparrow$&    $E_{warp} \downarrow $\\
\hline
Processed~\cite{CycleGAN2017}  & - &0.224 \\
\hline
Manual  &  25.64 & 0.194\\
Automatic & 25.52 & 0.193 \\

\hline

\end{tabular}
\vspace{1mm}
\end{table}

\begin{figure*}[t]
\centering
\begin{tabular}{@{}c@{\hspace{1mm}}c@{\hspace{1mm}}c@{\hspace{1mm}}c@{\hspace{1mm}}c@{\hspace{1mm}}c@{}}

\rotatebox{90}{\small \hspace{5mm} $t=0$ }&
\includegraphics[width=0.190\linewidth]{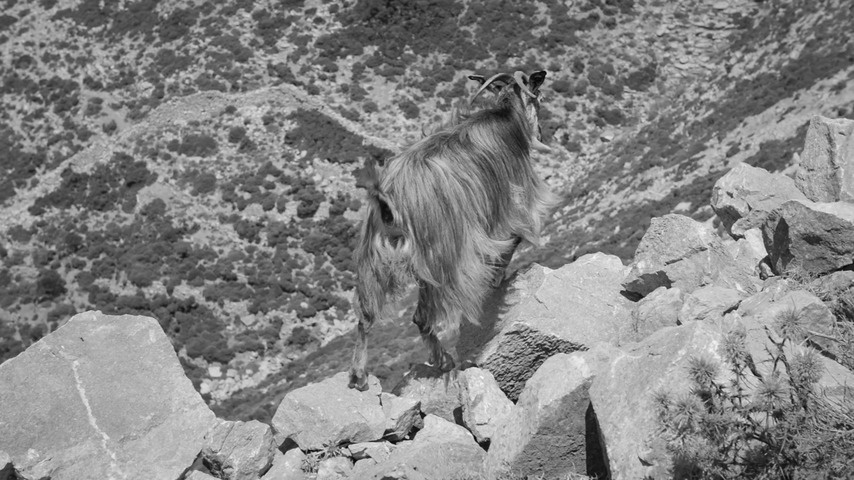}&
\includegraphics[width=0.190\linewidth]{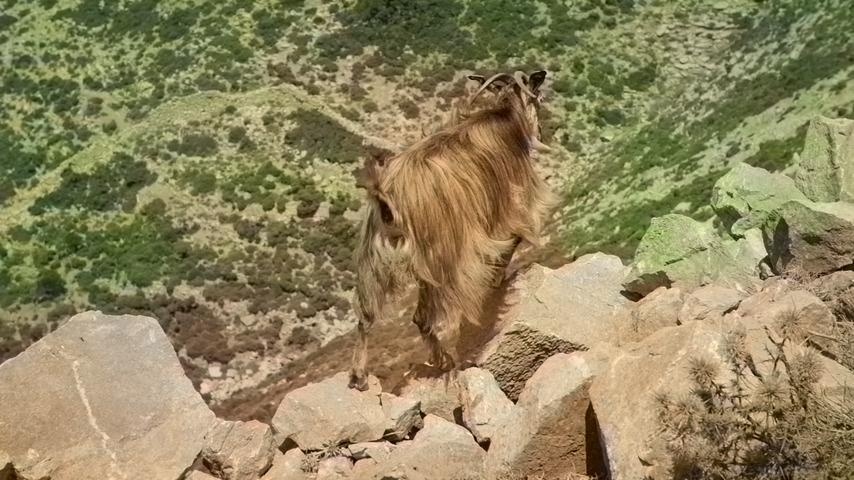}&
\includegraphics[width=0.186\linewidth]{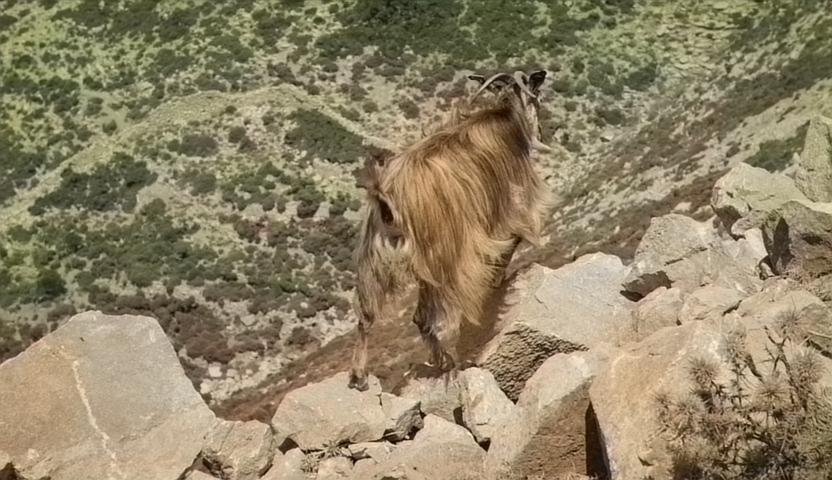}&
\includegraphics[width=0.186\linewidth]{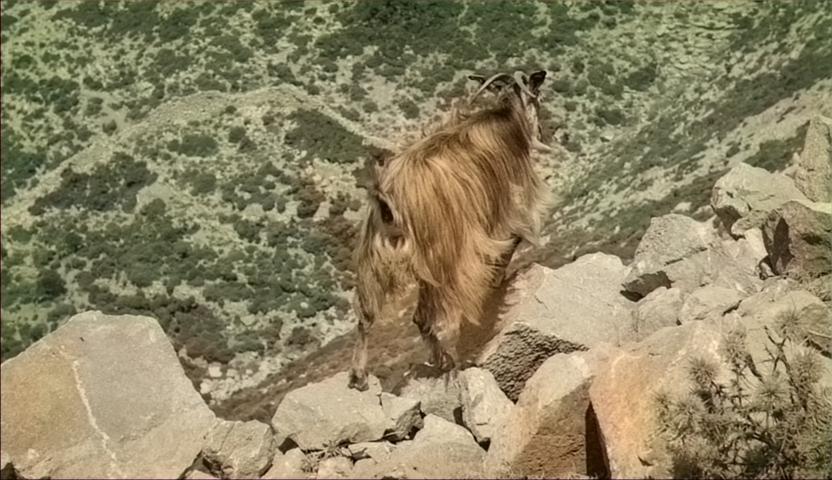}&
\includegraphics[width=0.186\linewidth]{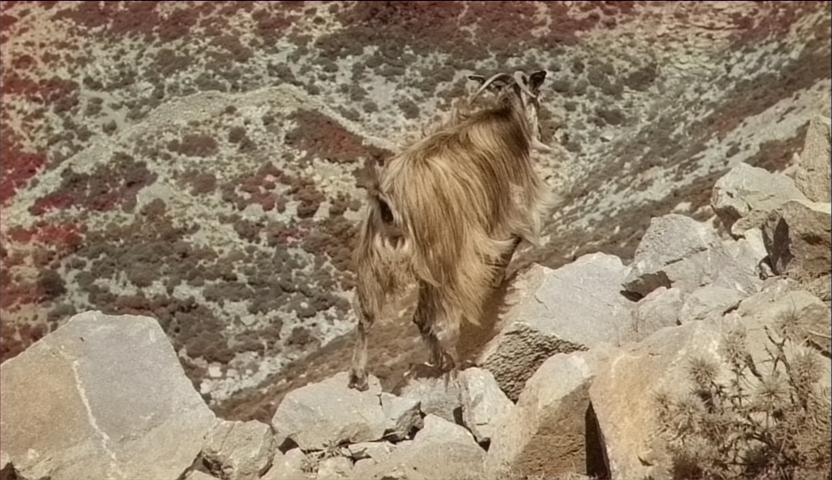}\\

\rotatebox{90}{\small \hspace{5mm} $t=10$ }&
\includegraphics[width=0.190\linewidth]{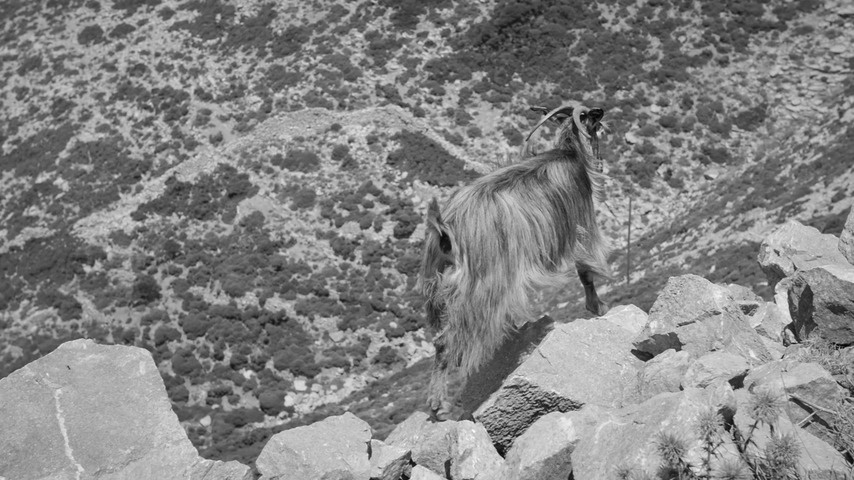}&
\includegraphics[width=0.190\linewidth]{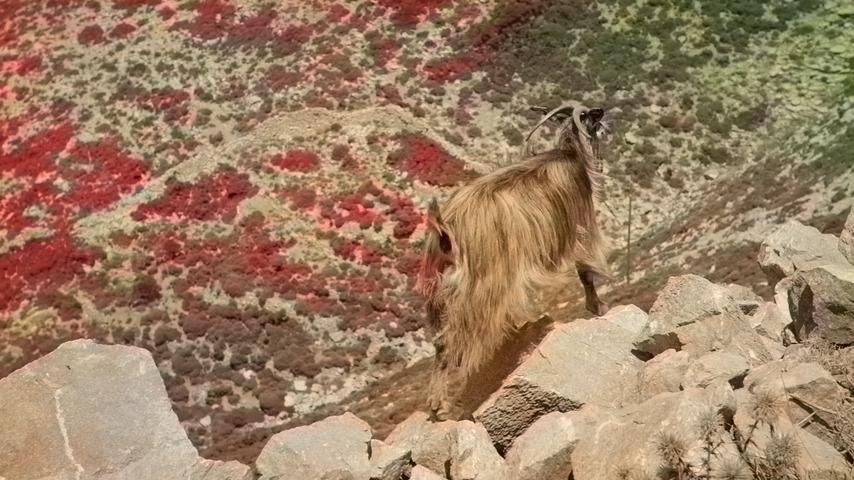}&
\includegraphics[width=0.186\linewidth]{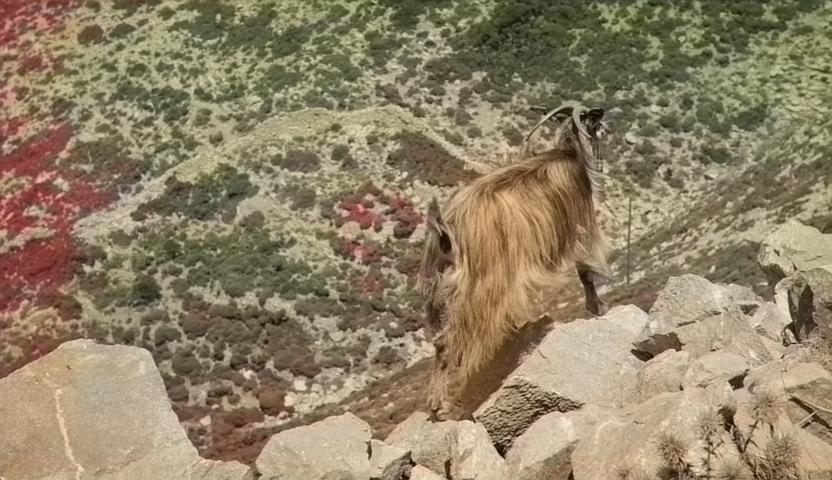}&
\includegraphics[width=0.186\linewidth]{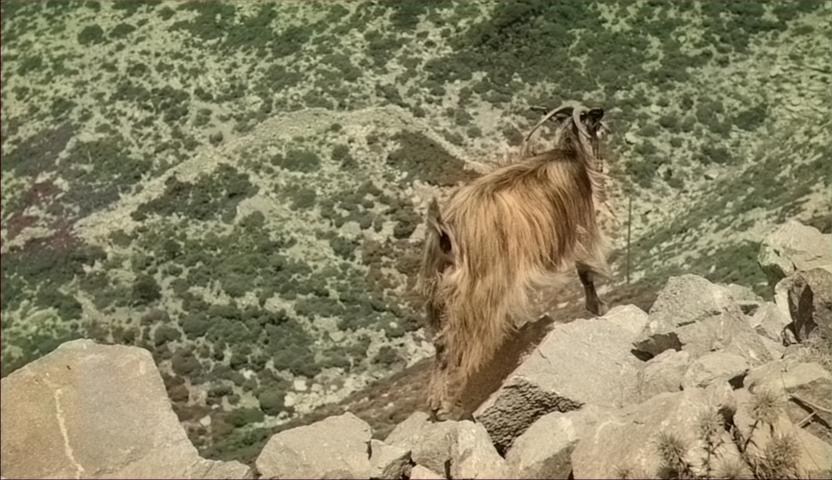}&
\includegraphics[width=0.186\linewidth]{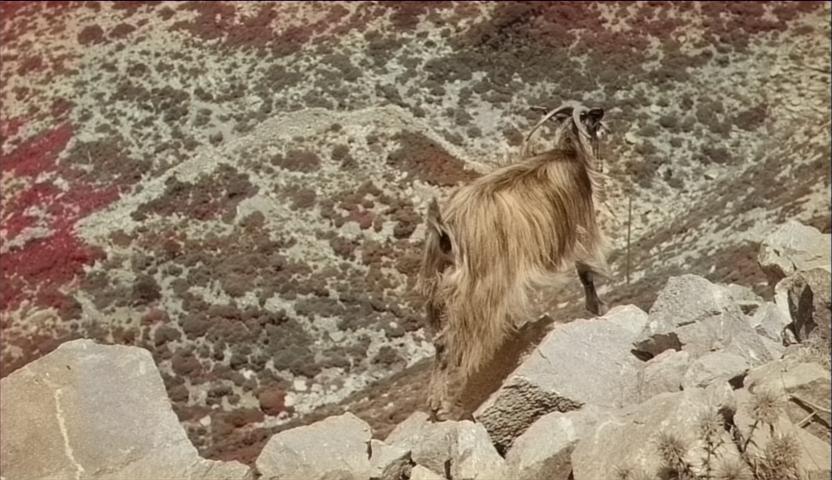}\\
& Input & Processed~\cite{zhang2016colorful}  & Without IRT: $O_t$   & With IRT: $O_t^{main}$ & With IRT: $O_t^{minor}$ \\
\end{tabular}
\\
\caption{With IRT, our method can improve video temporal consistency for the multimodal case, compared with the basic training architecture.}
\label{fig:AblationStudy}
\end{figure*}
\begin{figure*}[t]
\centering
\begin{tabular}{@{}c@{\hspace{1mm}}c@{\hspace{1mm}}c@{\hspace{1mm}}c@{\hspace{1mm}}c@{\hspace{1mm}}c@{\hspace{1mm}}c@{\hspace{1mm}}c@{\hspace{1mm}}c@{\hspace{1mm}}c@{}}

\includegraphics[width=0.119\linewidth]{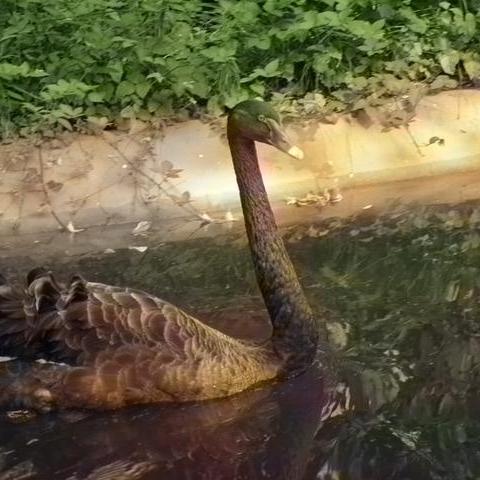}&
\includegraphics[width=0.119\linewidth]{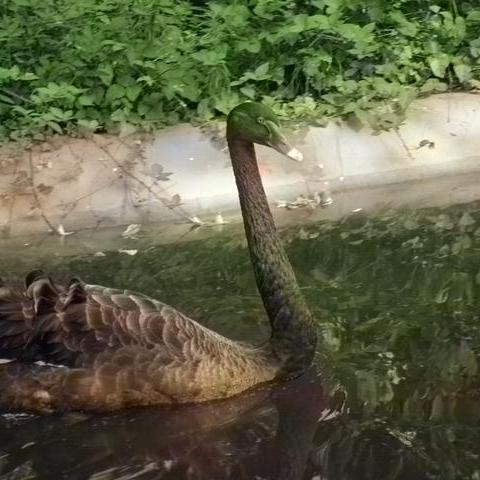}&

\includegraphics[width=0.119\linewidth]{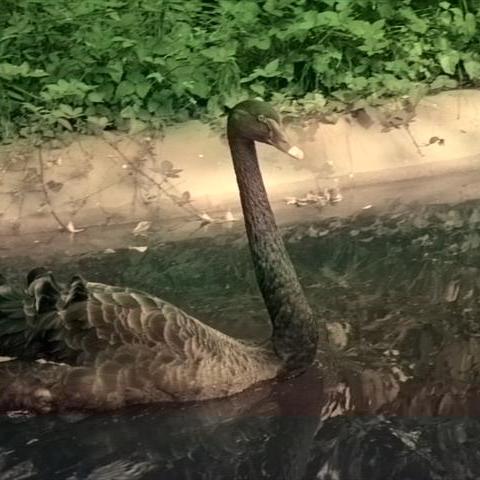}&
\includegraphics[width=0.119\linewidth]{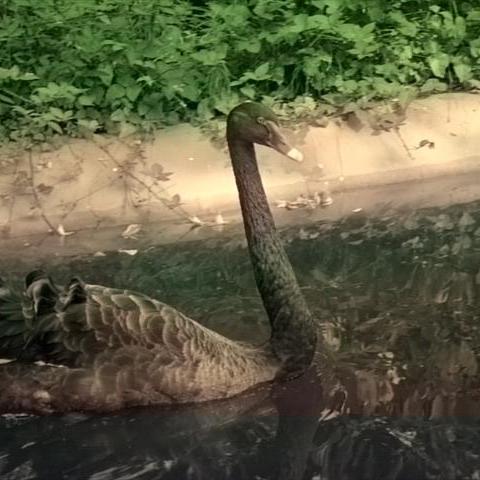}&

\includegraphics[width=0.119\linewidth]{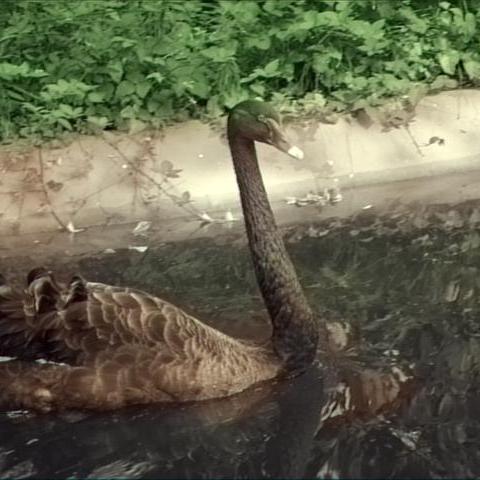}&
\includegraphics[width=0.119\linewidth]{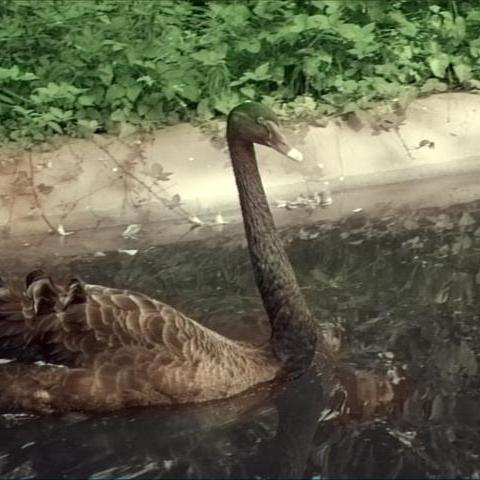}&

\includegraphics[width=0.119\linewidth]{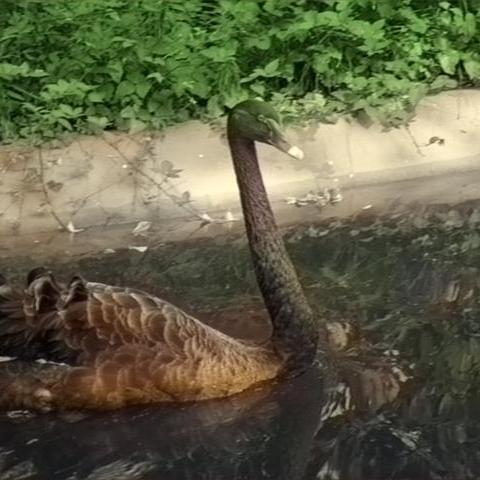}&
\includegraphics[width=0.119\linewidth]{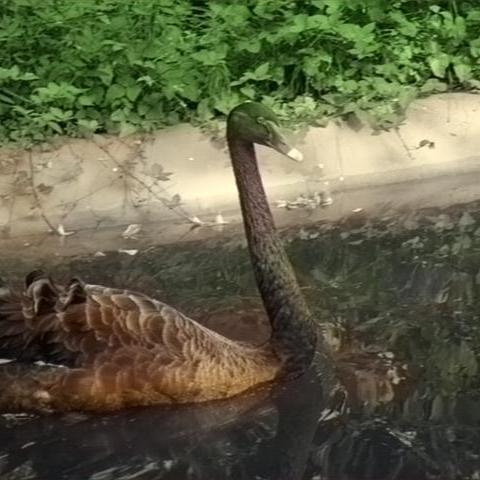}\\
$P_{18}$ & $P_{19}$ &
\small {FCN $O_{18}$} &\small{FCN $O_{19}$}  &
\small{ResUnet $O_{18}$} & \small{ResUnet $O_{19}$}&
\small{U-net $O_{18}$}  & \small{U-net $O_{19}$}  \\


\end{tabular}
\\
\caption{Analysis of CNN architectures for blind temporal consistency via DVP. Deep video prior is applied to various CNN architectures, and the temporal consistency of results by different CNNs is consistently better than processed frames~\cite{zhang2016colorful}.}
\label{fig:AblationStudy_2}
\end{figure*}

\begin{table}[t]
\small
\centering
\caption{Quantitative results of different network architectures for blind temporal consistency via DVP. }
\label{table:DifferentArchs}
\renewcommand{\arraystretch}{1.2}
\begin{tabular}{l|ccc}
\hline
Network & $E_{warp} \downarrow $& $F_{data} \uparrow$  \\
\hline
{Processed~\cite{zhang2016colorful}} & 0.2300 & -\\
\hline
U-net    & 0.1747 & 28.56\\
ResUnet & 0.1746 & 28.67 \\
FCN     & 0.1751  &28.64\\

\hline

\end{tabular}
\vspace{1mm}
\end{table}

\subsection{Ablation study} An ablation study is conducted to analyze the importance of IRT. We implement another version that removes IRT for colorization~\cite{zhang2016colorful} on the DAVIS dataset~\cite{Perazzi2016}. 
We find that $F_{data}$ for the basic architecture is high: 29.61 for the basic architecture and 28.20 for IRT. However, the perceptual performance with IRT is more consistent, as shown in Fig.~\ref{fig:AblationStudy}. This phenomenon is not surprising since it follows our motivation in Section~\ref{sec:IRT}. 

Another controlled experiment is conducted to verify our framework with two more architectures: FCN~\cite{long2015fully} and ResUnet~\cite{zchen2019seeing}. As shown in Fig.~\ref{fig:AblationStudy_2}, deep video prior also can be applied to these two networks. Although there is a minor difference among the results of the three architectures, the temporal consistency of each architecture is satisfactory compared with processed frames. Table~\ref{table:DifferentArchs} presents the quantitative results with different network architectures. All the evaluated networks can be used to improve temporal consistency, and the performance among the three evaluated networks is similar.

\begin{figure*}
\centering
\begin{tabular}{@{}c@{\hspace{1mm}}c@{\hspace{1mm}}c@{\hspace{1mm}}c@{\hspace{1mm}}c@{}}
\rotatebox{90}{\small \hspace{5mm} Reference }&
\includegraphics[width=0.240\linewidth]{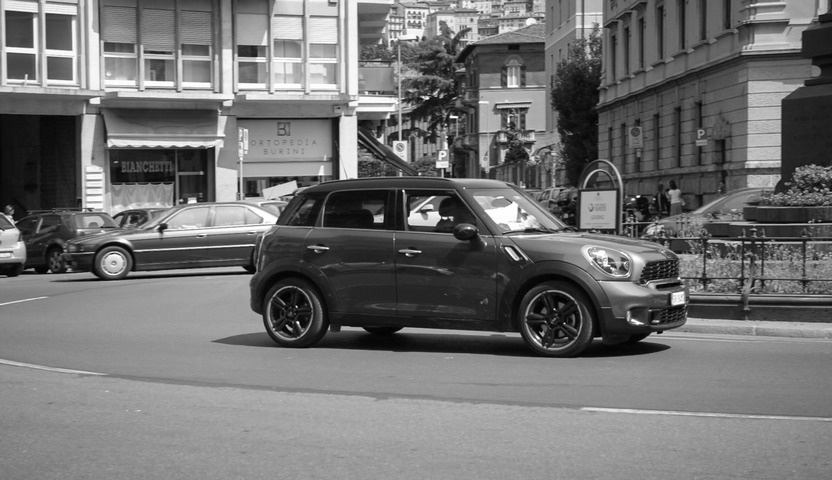}&
\includegraphics[width=0.240\linewidth]{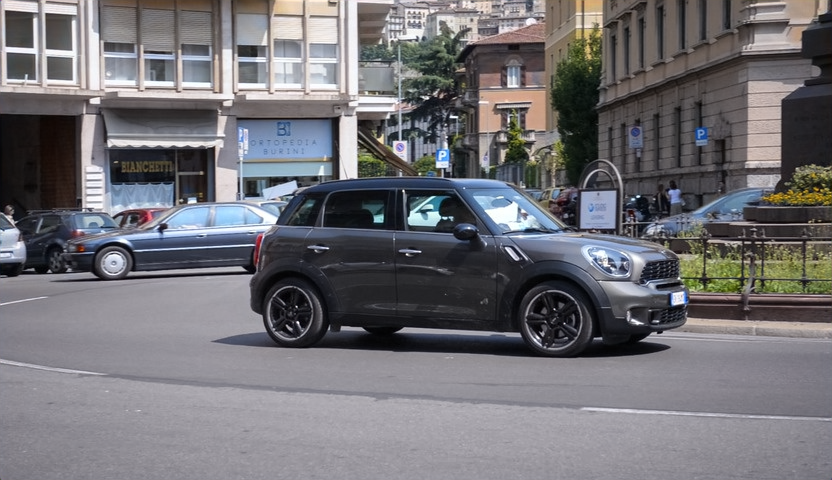}&
\includegraphics[width=0.240\linewidth]{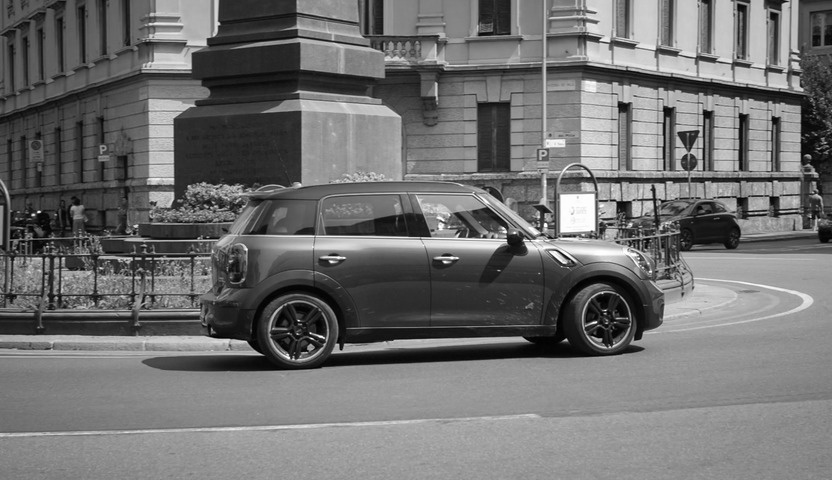}&
\includegraphics[width=0.240\linewidth]{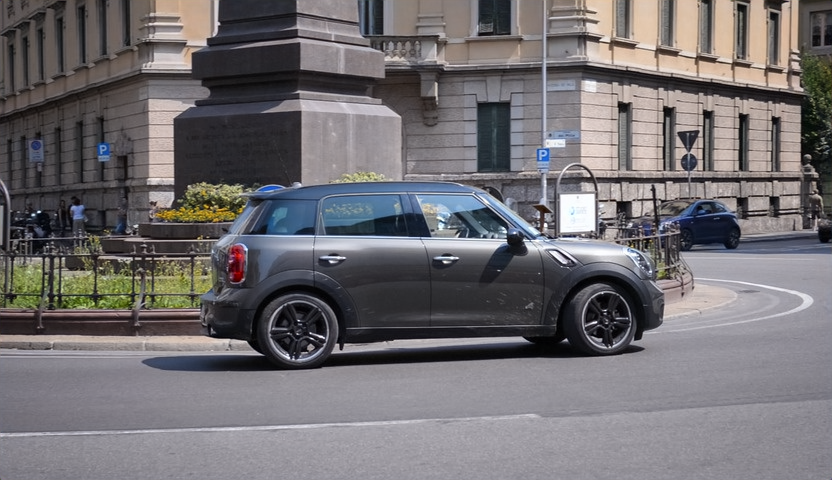}
\\
&  \multicolumn{2}{c}{$t=0$}  &  \multicolumn{2}{c}{$t=37$}  \\

\end{tabular}

\begin{tabular}{@{}c@{\hspace{1mm}}c@{\hspace{2mm}}c@{\hspace{2mm}}c@{}}

\rotatebox{90}{\small \hspace{10mm} $t=40$ }&
\includegraphics[width=0.318\linewidth]{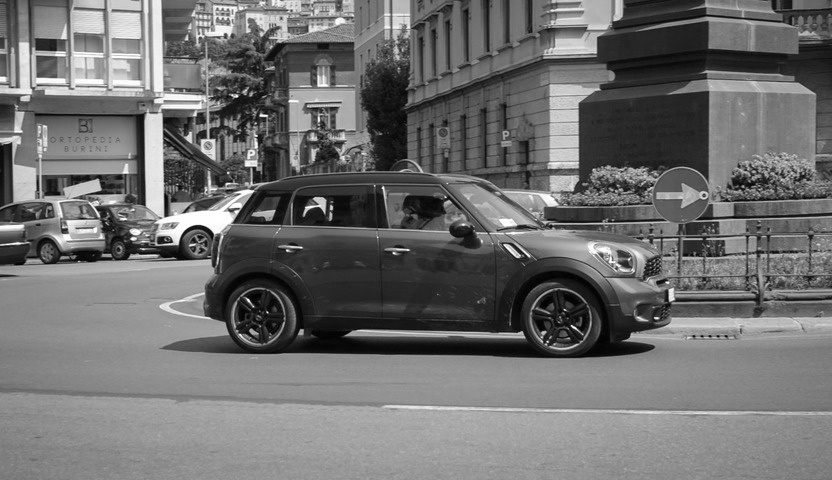}&
\includegraphics[width=0.318\linewidth]{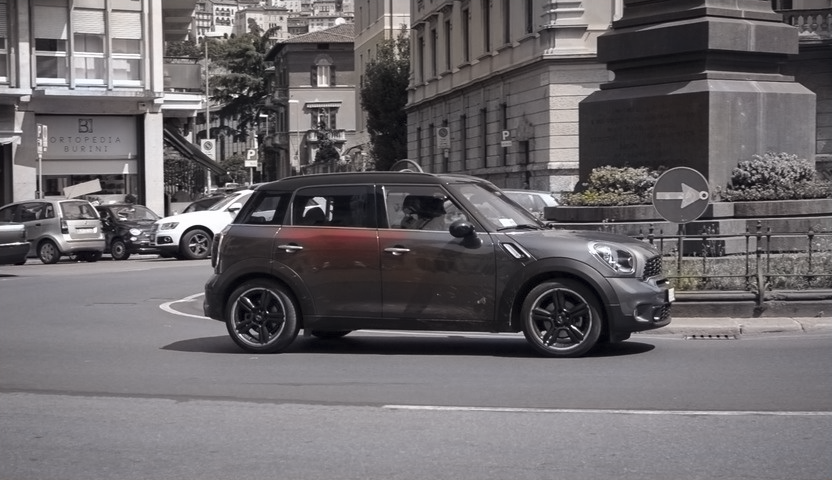}&
\includegraphics[width=0.318\linewidth]{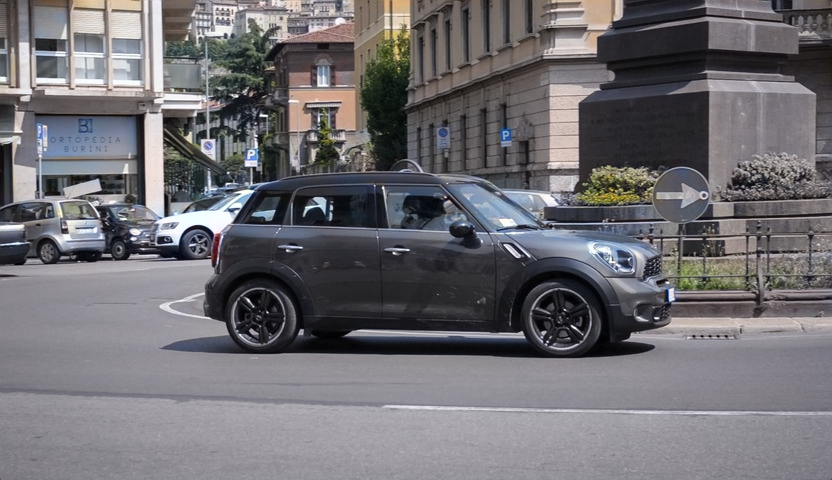}
\\

\rotatebox{90}{\small \hspace{10mm} $t=20$ }&
\includegraphics[width=0.318\linewidth]{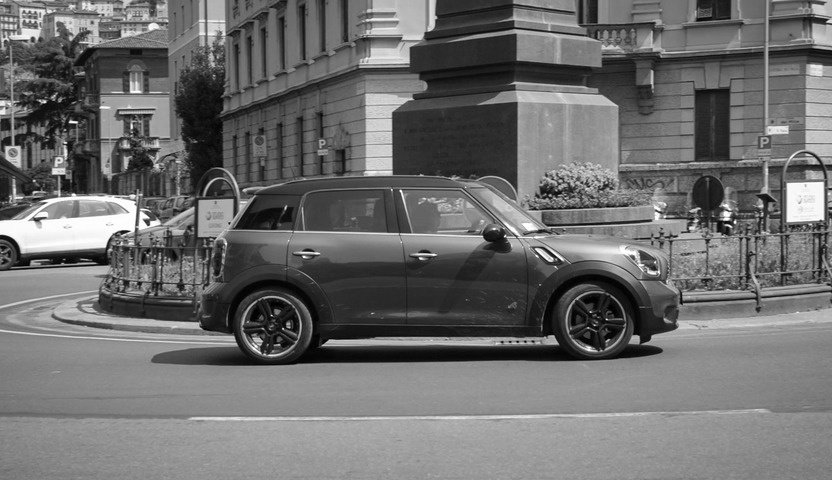}&
\includegraphics[width=0.318\linewidth]{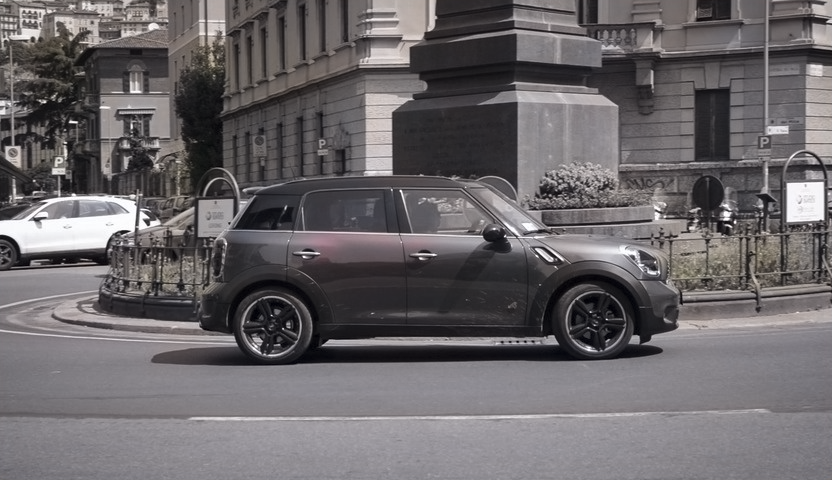}&
\includegraphics[width=0.318\linewidth]{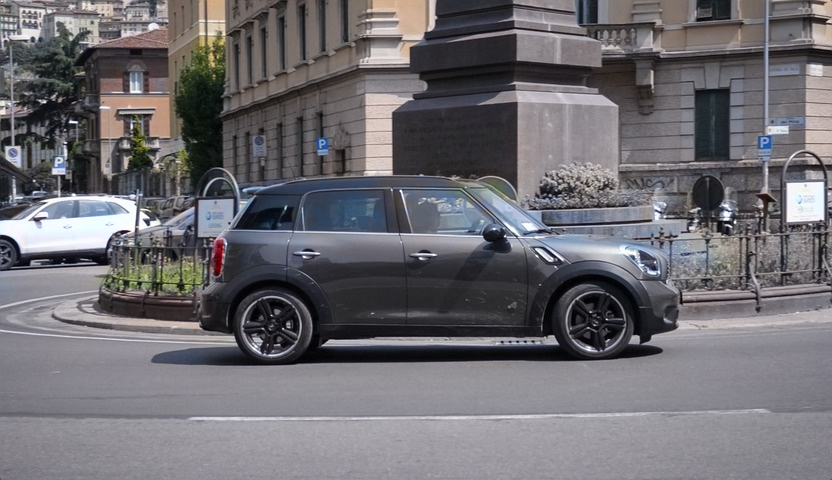}\\
& Input & DeepRemaster~\cite{DBLP:journals/tog/IizukaS19} & Ours \\

\end{tabular}

\vspace{1mm}
\caption{The perceptual comparison of video propagation via DVP and DeepRemaster~\cite{DBLP:journals/tog/IizukaS19} on the DAVIS dataset~\cite{Perazzi2016}. Our method can propagate the color more accurately.}

\label{fig:Color_Perceptual}
\end{figure*}

\begin{figure*}
\centering
\begin{tabular}{@{}c@{\hspace{1mm}}c@{\hspace{1mm}}c@{\hspace{1mm}}c@{\hspace{1mm}}c@{}}
\rotatebox{90}{\small \hspace{8mm} Input }&
\includegraphics[width=0.235\linewidth]{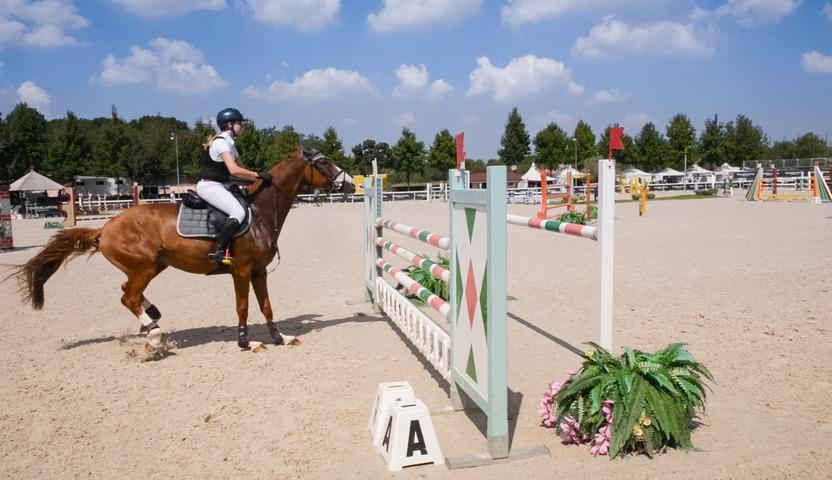}&
\includegraphics[width=0.235\linewidth]{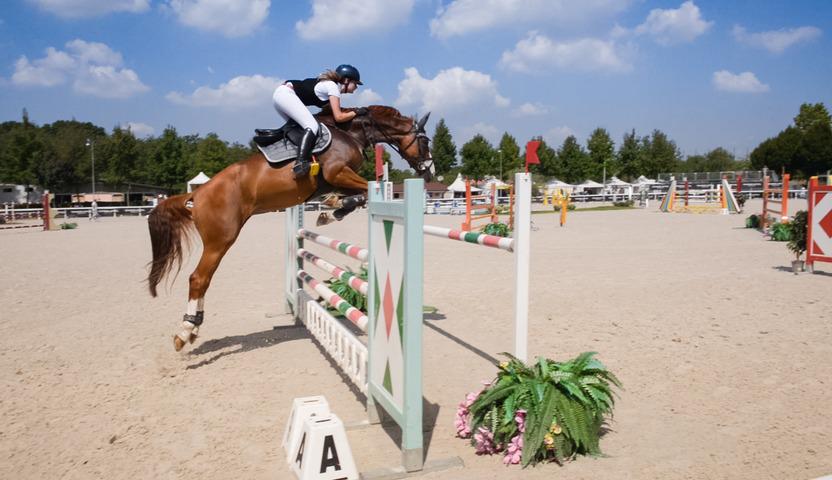}&
\includegraphics[width=0.235\linewidth]{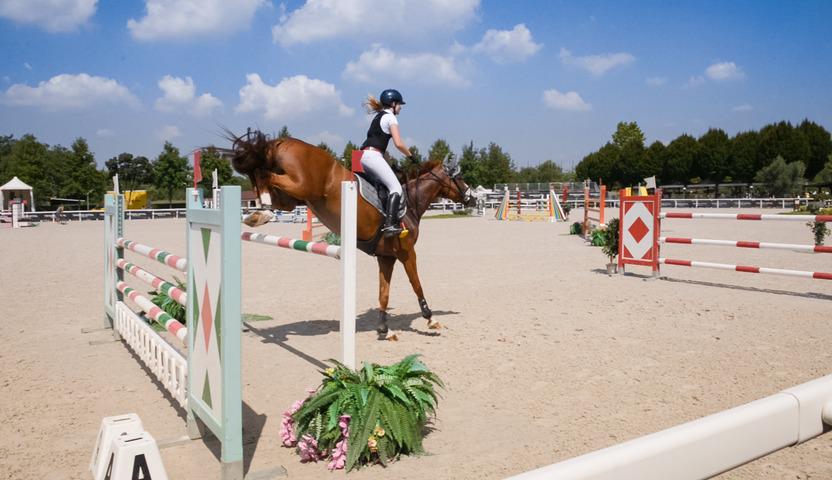}&
\includegraphics[width=0.235\linewidth]{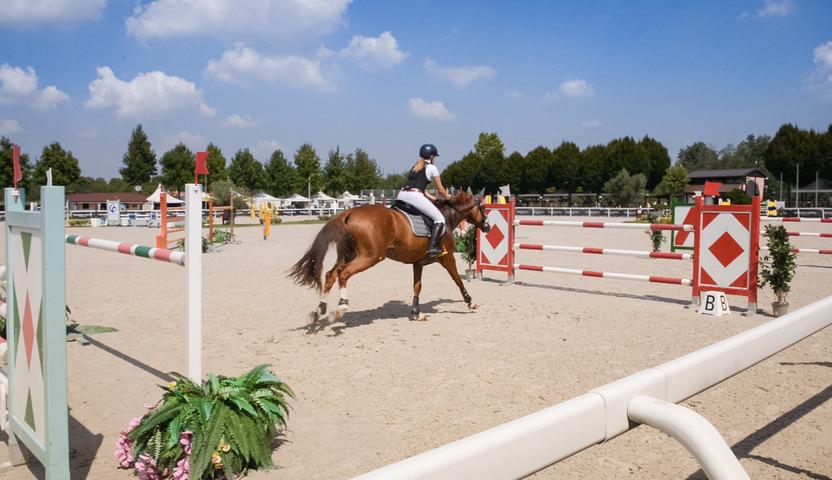}\\
\rotatebox{90}{\small \hspace{8mm} Ours }&
\includegraphics[width=0.235\linewidth]{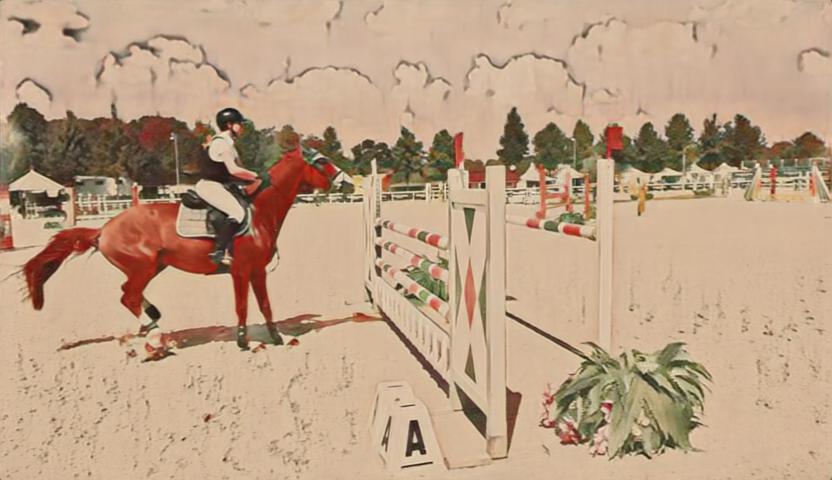}&
\includegraphics[width=0.235\linewidth]{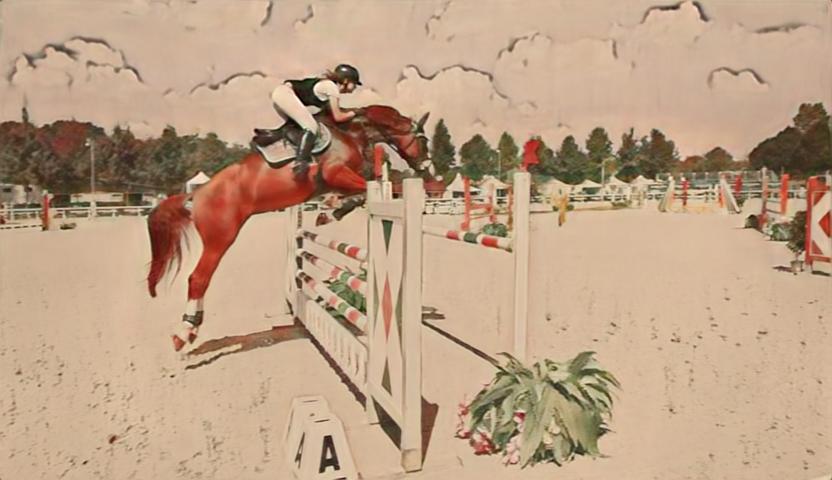}&
\includegraphics[width=0.235\linewidth]{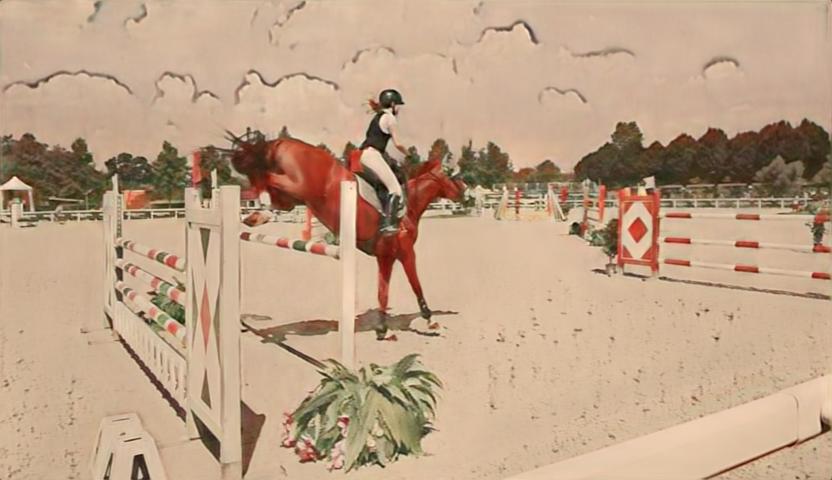}&
\includegraphics[width=0.235\linewidth]{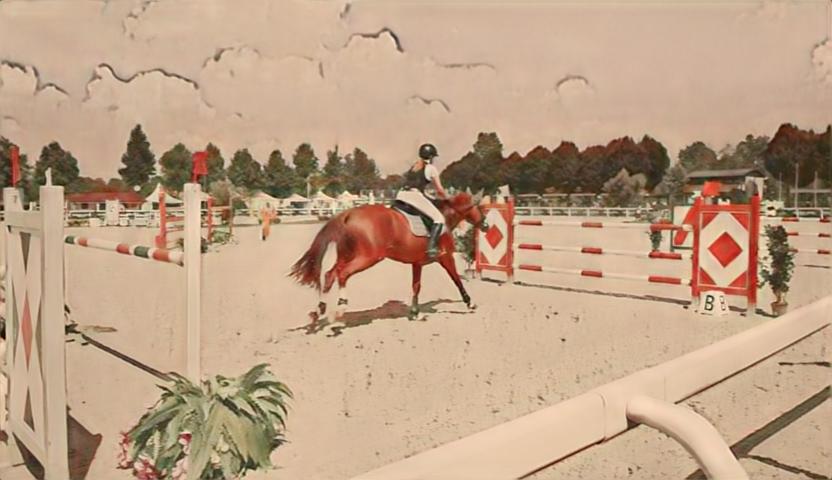}\\
&Reference & $t=10$  & $t=25$ & $t=40$\\
\end{tabular}
\vspace{1mm}

\caption{Results of video propagation via DVP. Our method is able to propagate the style information using a single reference image even when the motion is large.}
\label{fig:Style_MoreResults}
\end{figure*}

\begin{table}
\centering
\caption{Quantitative comparisons of video propagation via DVP and DeepRemaster~\cite{iizuka2016let}. The experiments are conducted on the DAVIS dataset~\cite{Perazzi2016} using various number of reference frames. }
\label{table:Color_Metrics}
\renewcommand{\arraystretch}{1.2}
\begin{tabular}{ccc}
\hline
\# Ref.&\multicolumn{2}{c}{PSNR} \\ 
\hline

     &DeepRemaster~\cite{DBLP:journals/tog/IizukaS19}  & Ours 
     \\ 
 1  & 24.92  & \textbf{27.29} \\
 2  & 25.27  &\textbf{28.40}  \\
 3 & 25.52 &\textbf{29.30}\\ 
 4 & 25.60 &\textbf{29.67}\\ 
 5 & 25.67 &\textbf{29.87}\\ 
\hline
\end{tabular}

\vspace{1mm}
\end{table}

\subsection{Applications of Video Propagation}
In this section, we apply deep video prior to video color propagation and video style propagation.
\subsubsection{Video Color Propagation}

\textbf{Experimental setup.}
We compare with the state-of-the-art video color propagation method DeepRemaster~\cite{DBLP:journals/tog/IizukaS19} since they adopt the same setting with our methods: different from previous video propagation methods~\cite{jampani:cvpr:2017,meyer2018deep}, both DeepRemaster and our method can receive an arbitrary number of reference images. 

We conduct the experiments on the DAVIS dataset~\cite{Perazzi2016}. For the test set of the DAVIS dataset, each video has 34 to 104 frames. For each video, we use one, two, or three color reference frames as reference (i.e., $R=1,2,3,4,5$). We convert the color frames to grayscale images as input frames. We use Peak Signal-to-Noise Ratio (PSNR) as an evaluation metric. 


\noindent \textbf{Color propagation results.} Table ~\ref{table:Color_Metrics} shows the quantitative results of our methods and DeepRemaster~\cite{DBLP:journals/tog/IizukaS19}. Our approach outperforms DeepRemaster when using a different number of reference images. Besides, we notice that the increased reference images can benefit the performance of both propagation methods. However, our method obtains more gain from the increased reference frames. 

Fig.~\ref{fig:Color_Perceptual} shows the perceptual comparison of our approach and DeepRemaster~\cite{DBLP:journals/tog/IizukaS19} on the DAVIS dataset~\cite{Perazzi2016}. We notice that the result of DeepRemaster~\cite{DBLP:journals/tog/IizukaS19} cannot propagate the color accurately in Fig.~\ref{fig:Color_Perceptual}. Their results are not as colorful as reference images. As a comparison, our method can propagate the color accurately, even with drastic background changes.

In practice, we find that DeepRemaster~\cite{DBLP:journals/tog/IizukaS19} is much more memory-hungry than ours: their model costs more memory than our approach, which make it hard to apply their model on high-resolution images. It is because their model uses 3D-CNN, and we adopt a u-net that is trained on a single frame.

\subsubsection{Video Segmentation Propagation}
Semi-supervised video object segmentation~\cite{marki2016bilateral} is to propagate a foreground mask in the first frame to all video frames. We also adopt u-net~\cite{ronneberger2015u} as our architecture and the cross-entropy loss as the data loss. 

\begin{figure*}[h]
\centering
\begin{tabular}{@{}c@{\hspace{1mm}}c@{\hspace{1mm}}c@{\hspace{1mm}}c@{\hspace{1mm}}c@{}}
\rotatebox{90}{\small \hspace{7mm} Input }&
\includegraphics[width=0.237\linewidth]{Figure/Experiments/segmentation/hoursejump/input/00000.jpg}&
\includegraphics[width=0.237\linewidth]{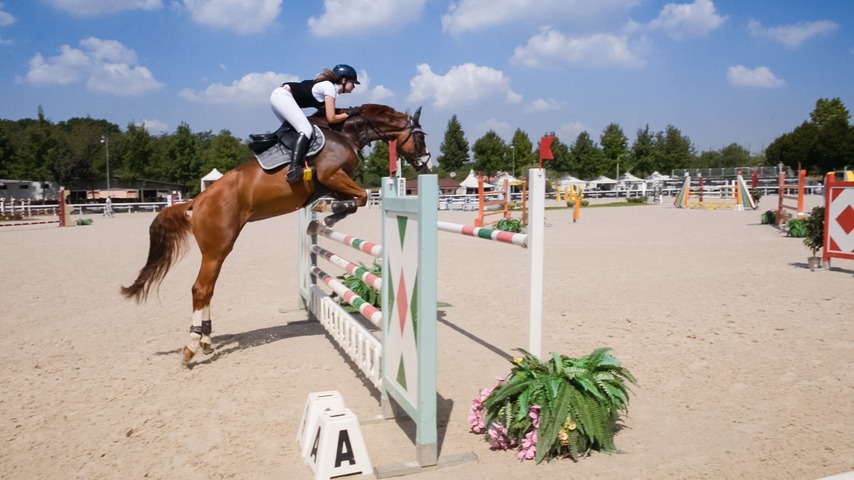}&
\includegraphics[width=0.237\linewidth]{Figure/Experiments/segmentation/hoursejump/input/00029.jpg}&
\includegraphics[width=0.237\linewidth]{Figure/Experiments/segmentation/hoursejump/input/00049.jpg}\\


\rotatebox{90}{\small \hspace{7mm} \scriptsize{STM~\cite{khoreva2019lucid}} }&
\includegraphics[width=0.237\linewidth]{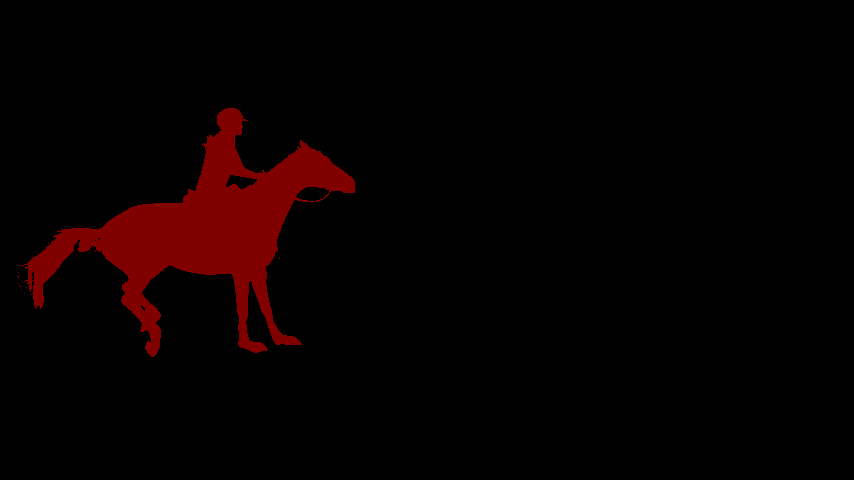}&
\includegraphics[width=0.237\linewidth]{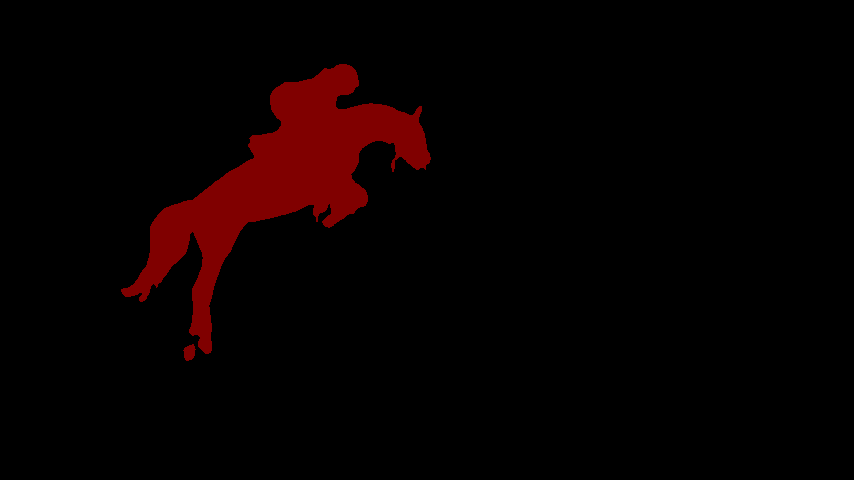}&
\includegraphics[width=0.237\linewidth]{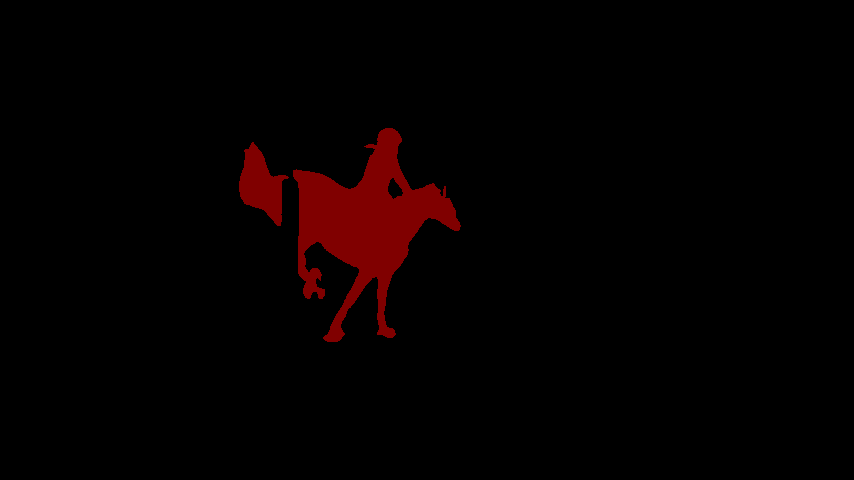}&
\includegraphics[width=0.237\linewidth]{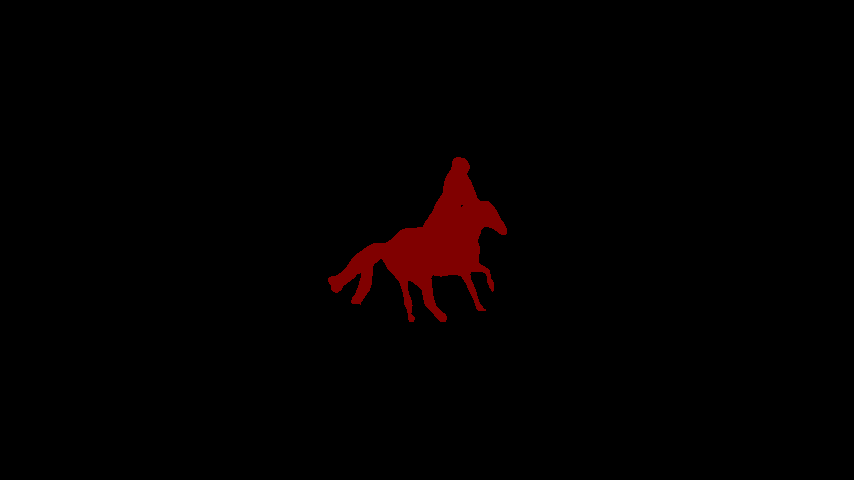}\\

\rotatebox{90}{\small \hspace{7mm} \scriptsize{VPN~\cite{jampani:cvpr:2017}} }&
\includegraphics[width=0.237\linewidth]{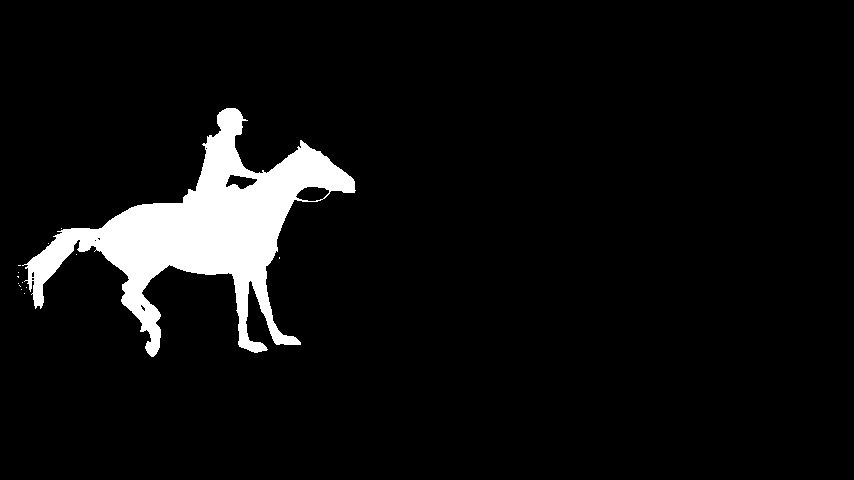}&
\includegraphics[width=0.237\linewidth]{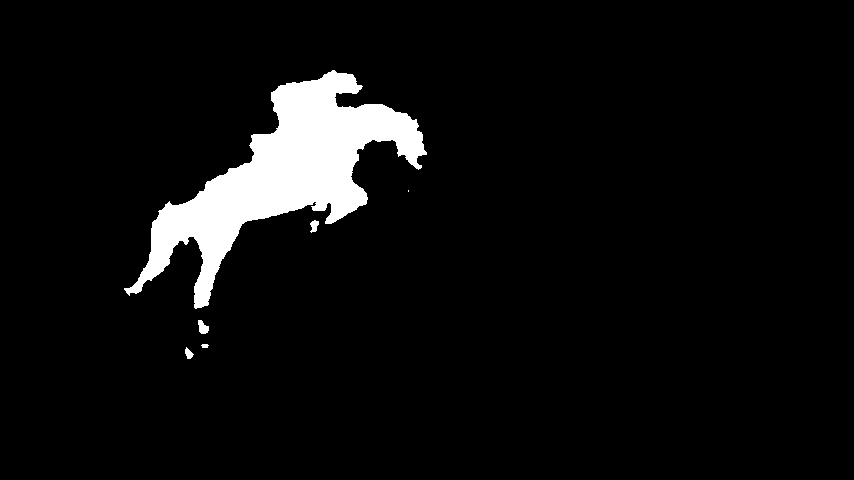}&
\includegraphics[width=0.237\linewidth]{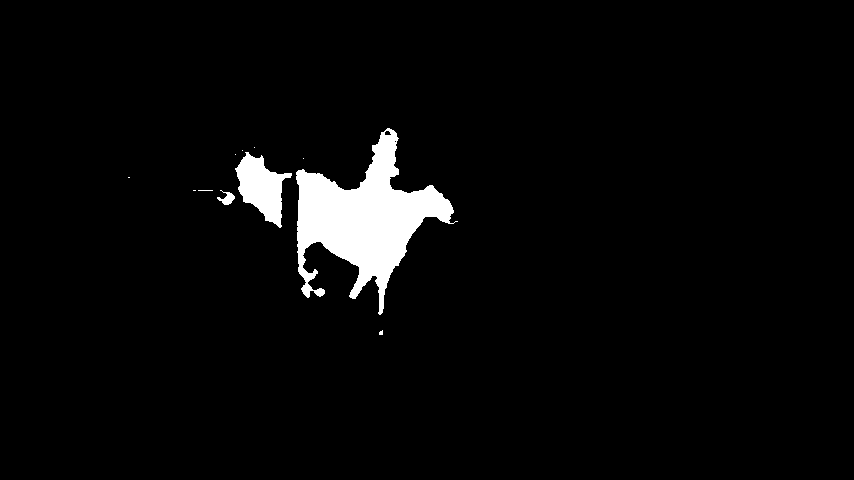}&
\includegraphics[width=0.237\linewidth]{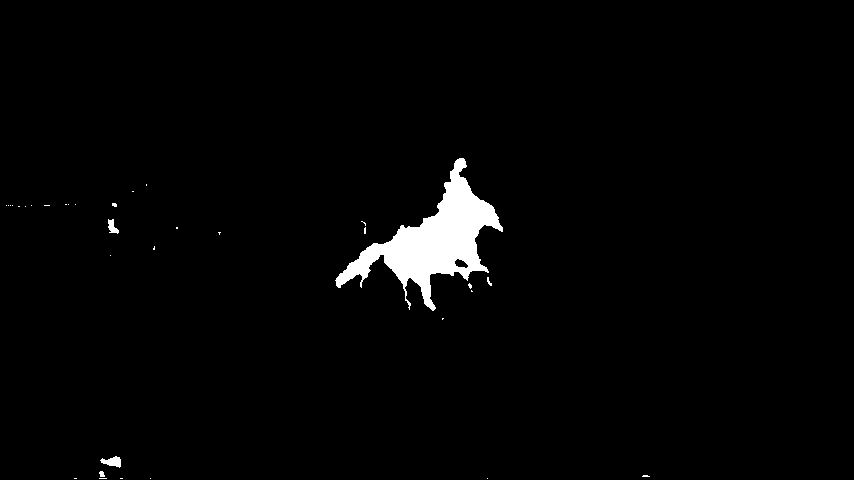}\\

\rotatebox{90}{\small \hspace{7mm} Ours}&
\includegraphics[width=0.237\linewidth]{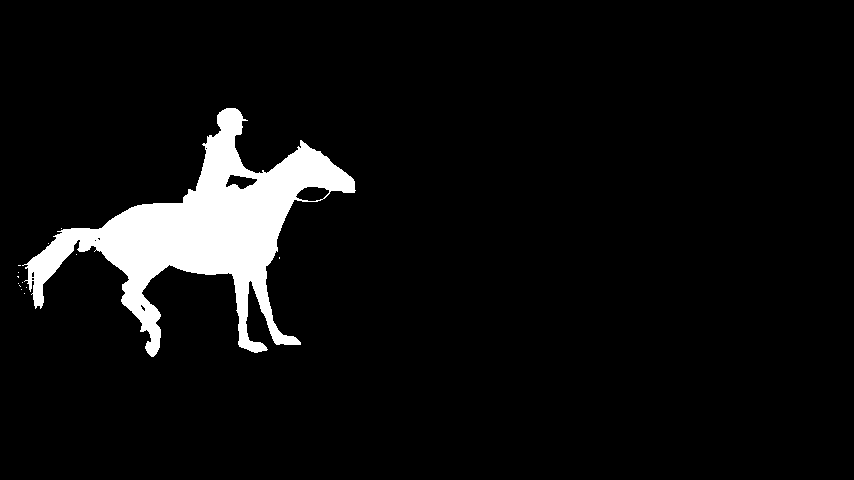}&
\includegraphics[width=0.237\linewidth]{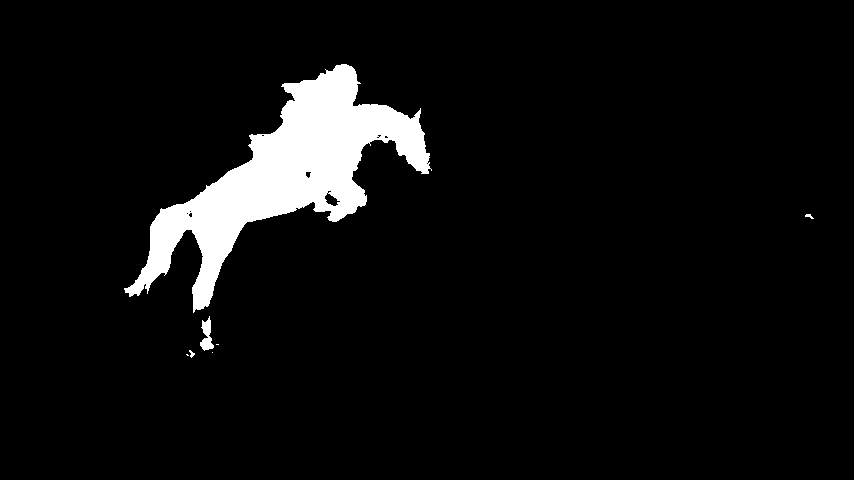}&
\includegraphics[width=0.237\linewidth]{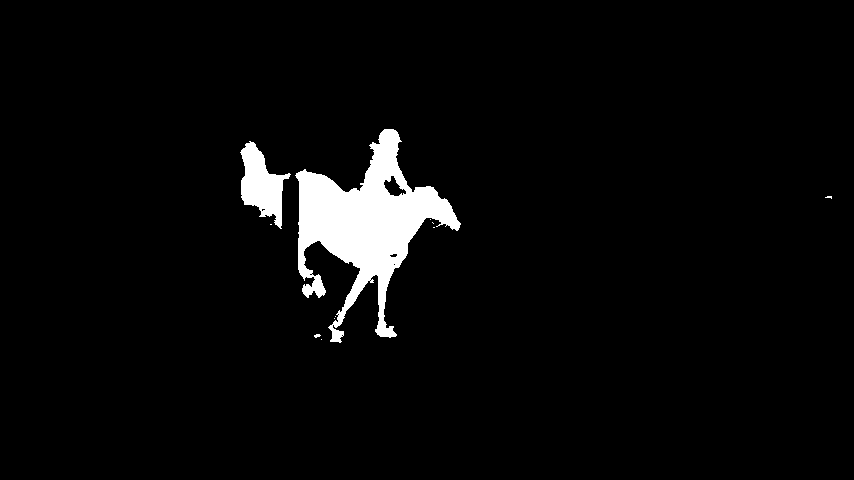}&
\includegraphics[width=0.237\linewidth]{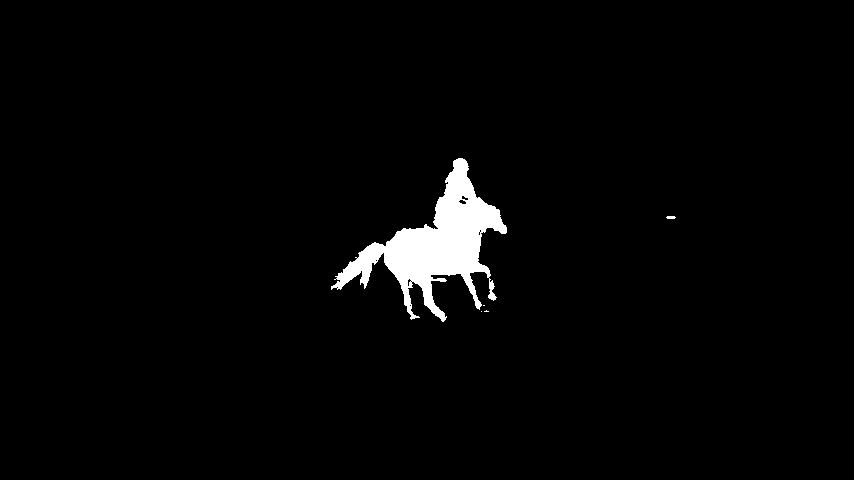}\\

& Reference &   Frame 10 & Frame 30 & Frame 50\\

\end{tabular}
\vspace{1mm}
\caption{Qualitative comparisons to state-of-the-art semi-supervised video object segmentation methods. }
\label{fig:Seg_Perceptual}
\end{figure*}

We present the quantitative results on the DAVIS2016 benchmark~\cite{Perazzi2016} in Table~\ref{table:Seg_Metrics}. Note that our approach is completely internal: we do not use any pretrained module (e.g., pretrained flow estimation networks) compared with other baselines. Compared with models that do not use pretrained models, our approach performs the best. However, our performance is lower than state-of-the-art models with pretrained modules. VPN~\cite{jampani:cvpr:2017} is another general video propagation framework that can be used to propagate segmentation and color information. Fig.~\ref{fig:Seg_Perceptual} shows an example for semi-supervised video segmentation, in which our approach can propagate the segmentation quite well.

We conduct an ablation experiment to study the importance of progressive propagation with pseudo labels (PPPL): we directly remove the memory queue and only use the first frame for the training. Table~\ref{table:PPPL} shows the quantitative results. Removing the PPPL strategy leads to apparent performance degradation in both metrics.

\begin{table*}[t]
\small
\centering
\renewcommand{\arraystretch}{1.2}
\caption{Quantitative results for video object segmentation. Compared with VPN~\cite{jampani:cvpr:2017}, our results are better in both PSNR and SSIM. Note that our model is not trained on large-scale dataset.} 
\begin{tabular}{lcccccccc}
\hline
     & STM & Lucid & OSVOS  &  VPN & BVS &FCP & JMP &Ours \\ 
     & ~\cite{oh2019video_STM}  &~\cite{khoreva2019lucid}  & ~\cite{caelles2017osvos} &~\cite{jampani:cvpr:2017} &   ~\cite{BVS} &~\cite{FCP}&~\cite{JMP}& \\ \hline
{Pretrained module} & {Networks}   & {Optical Flow} & {Networks}  & {Networks}  & {No}  &{No}  &{No}   & {No} \\
 J Mean &\textbf{88.7} &83.9 & 79.8 & 70.2 & 60.0&58.4&60.7&70.5\\ 
 F Mean &\textbf{90.1} & 82.0& 80.6& 65.5  & 58.8  &49.2&58.6&70.7\\ 

\hline

\end{tabular}
\vspace{1mm}
\label{table:Seg_Metrics}
\end{table*}


\begin{table}
\centering
\caption{Ablation experiment for progressive propagation with pseudo labels (PPPL). PPPL can effectively improve the performance of video segmentation propagation.}
\label{table:PPPL}
\renewcommand{\arraystretch}{1.2}
\begin{tabular}{lcc}
\hline
Model &J Mean & F Mean \\ 
\hline
Without PPPL  & 68.3 & 67.8  \\
With PPPL & \textbf{70.5} & \textbf{70.7} \\ 

\hline
\end{tabular}

\vspace{1mm}
\end{table}

\begin{table}
\centering
\caption{Quantitative comparisons for the effectiveness of different augmentation techniques. }
\label{table:Augmentation}
\renewcommand{\arraystretch}{1.2}
\begin{tabular}{lcc}
\hline
Augmentation &J Mean & F Mean \\ 
\hline
 No augmentation  & 60.9 & 60.5 \\
 Crop  & 63.1 & 61.4  \\
 Crop+Flip & 65.4 & 63.8 \\ 
Crop+Flip+Copy-paste & \textbf{70.5} & \textbf{70.7} \\ 

\hline
\end{tabular}

\vspace{1mm}
\end{table}

\subsubsection{Video Style Propagation}

We also apply our method to video style transfer~\cite{gatys2016image} to demonstrate the effectiveness of our video propagation framework. Similarly, we implement this approach by simply training on the single reference frame using a perceptual loss and u-net~\cite{ronneberger2015u}. The experiments are also conducted on the DAVIS dataset~\cite{Perazzi2016}. The stylized images we used are generated by CycleGAN~\cite{CycleGAN2017}. Note that our framework is designed for the situation that only reference images are available. In practice, unlike a pretrained CycleGAN~\cite{CycleGAN2017}, the style generator might be unavailable (e.g., a paint).

Fig.~\ref{fig:Style_MoreResults} shows the perceptual results for our video style propagation. Our method can also propagate the style information even in situations of occlusions, change of background, and moving objects.

\subsection{Analysis}
\textbf{Failure case.} We notice performance degradation for deep video prior when the number of frames is very large (e.g., 3000 frames). We conduct an experiment to study the impact of different window sizes. Specifically, we use a 2-min video from the KITTI dataset~\cite{Geiger2012CVPR} for the task of video colorization. We set the window size to 3 seconds, 10 seconds, 30 seconds, and 120 seconds respectively. Accordingly, we split the video into 40 clips, 12 clips, four clips, and one clip. As shown in Table~\ref{table:WindowSize}, we can observe noticeable performance degradation when the video is long. Also, training the network with a small window size can obtain satisfying performance.

\noindent \textbf{Data augmentation for DVP.} Data augmentation techniques improve the performance of semi-supervised video object segmentation effectively. Specifically, we adopt three data augmentation techniques: crop, flip, copy-paste. The copy-paste means that we copy the foreground and paste it to a random location at the image. As shown in Table~\ref{table:Augmentation}, all augmentation techniques improve the quantitative results.

\begin{table}[h]
\small
\centering
\caption{Comparisons of training on various number of clips on a 120-second video (1200 frames/10 frame per second). Note that the motion of this video is also large.}
\label{table:WindowSize}
\renewcommand{\arraystretch}{1.2}
\begin{tabular}{ll|cc}
\hline
\# Clip & Seconds per clip &  PSNR  \\
\hline
1  & 120 seconds & 31.46  \\
4  & 30 seconds & 31.71  \\
12 & 10 seconds & 32.60    \\
40 & 3 seconds & 32.49  \\
\hline

\end{tabular}
\vspace{1mm}
\end{table}

\section{Discussion}
We have presented a simple and general approach to improving temporal consistency for videos processed by image operators. Utilizing deep video prior that the outputs of a CNN for corresponding patches in video frames should be consistent, we achieve temporal consistency by training a CNN from scratch on a single video. Our approach is considerably simpler than previous work and produces satisfying results with better temporal consistency. Our iteratively reweighted training strategy also solves the challenging multimodal inconsistency well. We believe that the simplicity and effectiveness of the presented approach can transfer image processing algorithms to their video processing counterpart. Consequently, we can benefit from the latest image processing algorithms by applying them to videos directly. 

Practical acceleration strategies are proposed to shorten the processing time. Specifically, we accelerate the training process by providing a pretrained network and a coarse-to-fine training strategy. Although we still need to train an individual model for each video, the process can be accelerated by 5 times through our strategy. 

We further extend the idea of DVP to video propagation. Since the outputs of a CNN for corresponding patches in video frames should be consistent and input frames are quite similar, training on reference frames can help us propagate the information without finding explicit correspondences. Experimental results demonstrate the effectiveness of our approach on propagating color information and style information.

For future work, we believe the idea of DVP can be further expanded to other types of data, such as 3D data and multi-view images. DVP does not rely on the ordering of video frames and should be naturally applicable to maintaining multi-view consistency among multiple images. For 3D volumetric data, a 3D CNN may also exhibit a similar property to DVP.

\ifCLASSOPTIONcaptionsoff
  \newpage
\fi

\vspace{-6mm}

\begin{IEEEbiography}[{\includegraphics[width=1in,height=1.25in,clip,keepaspectratio]{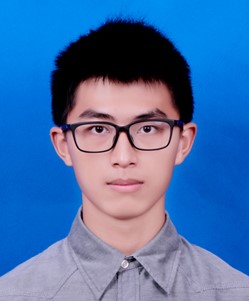}}]{Chenyang Lei}
 received the B.E. degree from Zhejiang University in 2018. He is currently pursuing the Ph.D. degree with the Hong Kong University of Science and Technology. He feels interested in the computation
photography, low-level computer vision and video processing.
\end{IEEEbiography}

\vspace{-6mm}

\begin{IEEEbiography}[{\includegraphics[width=1in,height=1.25in,clip,keepaspectratio]{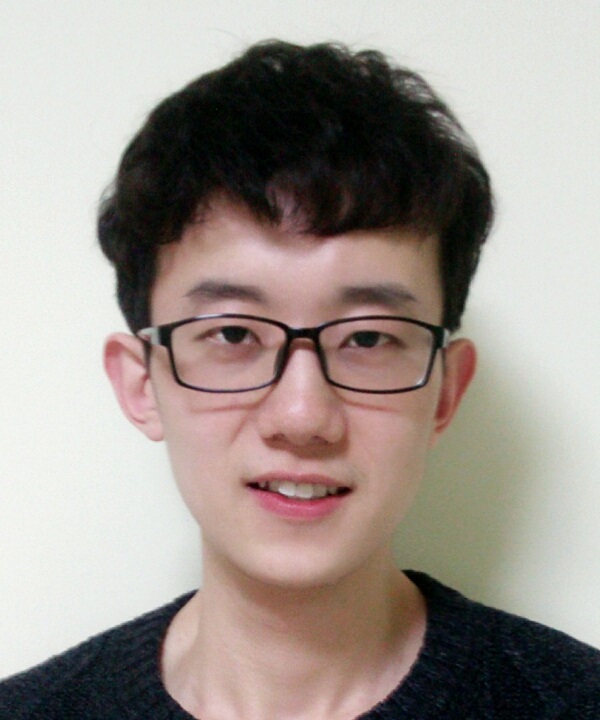}}]{Yazhou Xing} 
 received the B.E. degree from Wuhan University in 2018. He is currently pursuing the Ph.D. degree with the Hong Kong University of Science and Technology. His research interests include computation
photography, low-level computer vision and video processing.
\end{IEEEbiography}

\vspace{-6mm}

\begin{IEEEbiography}[{\includegraphics[width=1in,height=1.25in,clip,keepaspectratio]{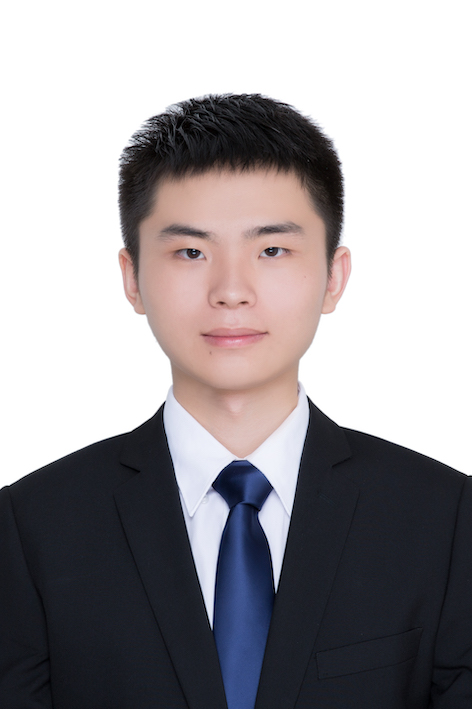}}]{Hao Ouyang}
 received the B.E. degree from the Chinese University of Hong Kong in 2018. He is currently pursuing the Ph.D. degree in the Hong Kong University of Science and Technology. His research interest includes low-level computer vision and steganography.
\end{IEEEbiography}

\vspace{-6mm}

\begin{IEEEbiography}[{\includegraphics[width=1in,height=1.25in,clip,keepaspectratio]{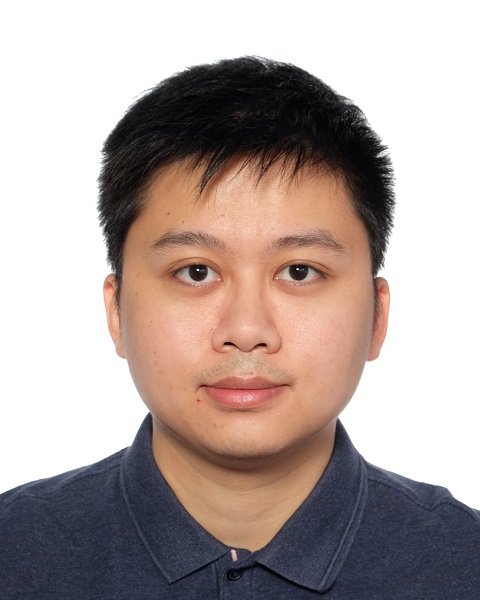}}]{Qifeng Chen} is an assistant professor of the Department of Computer Science and Engineering and the Department of Electronic and Computer Engineering at HKUST. He received his Ph.D. in computer science from Stanford University in 2017, and a bachelor's degree in computer science and mathematics from HKUST in 2012. He is named one of 35 Innovators under 35 in China in 2018 by MIT Technology Review. He won the Google Faculty Research Award 2018. He is the HKUST ACM programming faculty coach and won the 2nd place worldwide at the ACM-ICPC World Finals in 2011. He is a member of IEEE.
\end{IEEEbiography}





\bibliographystyle{plain}
\bibliography{reference}

\end{document}